\documentclass[letterpaper,11pt]{article}
\usepackage[utf8]{inputenc}
\usepackage[T1]{fontenc}

\usepackage[english]{babel}
\usepackage{authblk}
\usepackage[margin=1in]{geometry}

\usepackage{amsfonts}       %
\usepackage{nicefrac}       %
\usepackage{microtype}      %
\usepackage{amsmath,bbm,bm,mathtools} 
\usepackage{float}
\usepackage{placeins}
\usepackage{amssymb}
\usepackage{amsthm}
\usepackage{algorithm}
\usepackage{booktabs}       %
\usepackage{graphicx}
\usepackage{multirow}
\usepackage{subcaption}
\usepackage[colorlinks=true, linkcolor=black, citecolor=black, filecolor=black, urlcolor=blue]{hyperref}
\usepackage{xcolor}
\usepackage{comment}
\usepackage[shortlabels]{enumitem}
\usepackage[normalem]{ulem}
\usepackage{makecell}
\usepackage{algpseudocode}
\usepackage[capitalize,nameinlink]{cleveref}
\newcommand{\eg}{e.g.,}
\newcommand{\ie}{i.e.,}

\usepackage{natbib}
\setcitestyle{authoryear,open={(},close={)}}

\hypersetup{
     colorlinks,
     linkcolor={red!50!black},
     citecolor={blue!50!black},
     urlcolor={blue!80!black}
 }

\usepackage{tcolorbox}
\tcbuselibrary{listings, breakable}

\makeatletter
\renewcommand\texttt[1]{%
  \begingroup\ttfamily\hyphenchar\font=`\-\relax #1\endgroup}
\makeatother

\newcommand{\promptbox}[1]{%
  \smallskip
  \begin{tcolorbox}[
    breakable,
    colback=gray!8,
    colframe=gray!50,
    boxrule=0.4pt,
    arc=2pt,
    left=4pt, right=4pt, top=3pt, bottom=3pt,
    before skip=4pt, after skip=4pt
  ]
  {\small\ttfamily\setlength{\parskip}{2pt} #1}
  \end{tcolorbox}%
  \smallskip
}

\definecolor{DarkGreen}{rgb}{0.1,0.5,0.1}
\definecolor{DarkRed}{rgb}{0.5,0.1,0.1}
\definecolor{DarkBlue}{rgb}{0.1,0.1,0.5}
\definecolor{Gray}{rgb}{0.2,0.2,0.2}

\newcommand{\ours}{TTH}
\newcommand{\cirange}[2]{[#1,\,#2]}

\makeatletter
\newcommand\footnoteref[1]{\protected@xdef\@thefnmark{\ref{#1}}\@footnotemark}
\makeatother

\title{Test-Time Hinting for Black-Box Vision-Language Models}

\author{\large{Kaihua Hou$^{1,2}$, Abhijith Varma Mudunuri$^{1}$, Jiaxing Qiu$^{3}$, Roxana Daneshjou$^{4}$,\\[0.6em] Thomas Hartvigsen$^{3}$, and Ahmed Alaa$^{1,2}$}}

\affil{$^1$\,UC Berkeley ~ $^2$\,UCSF ~ $^3$\,University of Virginia ~ $^4$\,Stanford University}

\date{}%
\begin{document}
\maketitle
\begin{abstract}
Test-time scaling (TTS) methods have proven highly effective for LLMs. Existing TTS methods for vision-language models (VLMs), however, largely require open-weight model access or spend the additional compute on repeated sampling and post-hoc critique. We show that the errors of frontier VLMs are predictable from the input alone and largely shared across models, so test-time scaling can act on a likely error before the model answers. We therefore propose {\it Test-Time Hinting}, a method that improves VLM performance with a single call to the target VLM and requires only black-box API access, which makes it broadly applicable to frontier closed-weight models. A lightweight hint generator, trained on hints verified to repair real errors, writes a short ``hint'' that is prepended to the prompt and points the VLM to the relevant visual evidence without revealing the answer. Test-Time Hinting significantly improves the accuracy of multiple closed-weight VLMs on multiple-choice natural-image VQA benchmarks, and these gains generalize, without retraining, to unseen benchmarks and to unseen black-box and open-weight VLMs. Controls trace the gains to hints grounded in the visual evidence of each input and to training on verified repairs. Our results suggest that the predictability of VLM errors is a practical lever for test-time scaling, with one small model serving many frontier VLMs.
\end{abstract}
{\bf Code:} \href{https://github.com/AlaaLab/test-time-hinting}{https://github.com/AlaaLab/test-time-hinting}

\section{Introduction}
\label{sec:intro}
Test-time scaling (TTS) offers an alternative to the conventional strategy of scaling model size or training data. It instead directs additional compute resources to improve the accuracy of a frozen model at inference time. A wide range of TTS methods, from chain-of-thought prompting \citep{wei2022cot} and self-consistency \citep{wang2023selfconsistency} to process reward models \citep{lightman2023letverify} and optimal compute allocation \citep{snell2024scalingttc}, have shown that performance continues to grow (often significantly) with inference-time compute. However, TTS research has concentrated primarily on LLMs. In this paper, we develop TTS methods tailored to vision-language models (VLMs).

Existing TTS approaches for VLMs have two limitations that restrict their practical application. First, the most effective TTS strategies, such as reward-model-guided search, iterative sampling, and process supervision, require open access to model weights or internal representations, which excludes frontier closed-weight VLMs accessible only through APIs \citep{zhang2023multimodalcot,xu2024llavacot,avogaro2026sparc,kaya2026efficient,jeddi2026avis}. Second, even methods that operate through the model interface rely on \emph{repeated} sampling or multi-pass inference through the same target model, which increases the per-query cost at inference time \citep{madaan2023selfrefine,wang2023selfconsistency}.

\begin{figure}[t]
  \centering
  \includegraphics[width=\textwidth]{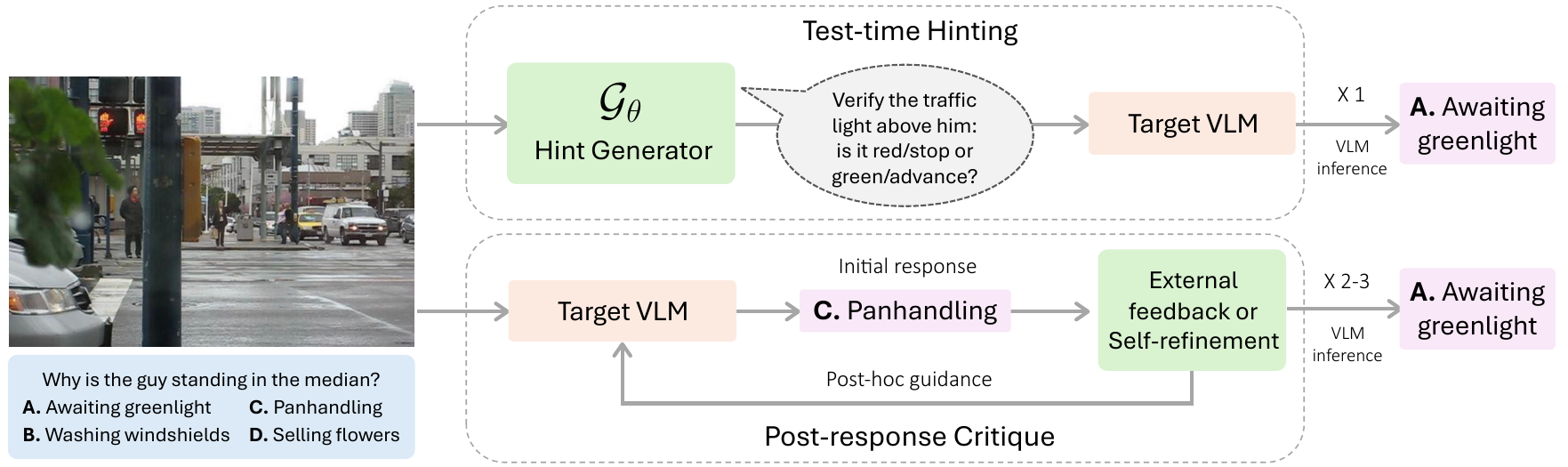}
  \caption{{\small \textbf{Comparison between Test-Time Hinting and post-response critique.} \ours{} generates a hint at inference time \emph{before} the target VLM answers, enabling single-pass inference with closed-weight models.}}
  \label{fig:overview}
  \vspace{-.1in}
\end{figure}

In this paper, we develop a TTS method for VLMs that requires only black-box API access and a single call to the target model, and is applicable to natural-image understanding tasks. The method applies to any VLM, open-weight or closed-weight, but is especially valuable in deployment settings where a frontier model is used in a domain-specific application and further improvement using proprietary data or evaluations is needed. Our method is motivated by an empirical finding that VLM errors cluster around a small set of recurring failure modes that depend largely on the input, and that are correlated across distinct VLMs. These regularities imply that the structure of VLM errors is learnable and can be anticipated from the image and text inputs alone, before the model responds. Test-time scaling can therefore act on a likely error before the target answers, in a single call.

We call our method \textbf{Test-Time Hinting (TTH)}. TTH trains a (lightweight) hint generator to produce, for a given image--text pair, a short natural-language hint that is prepended to the target VLM's prompt before it responds. The hint points the model to the visual evidence the question depends on, steering it away from a likely failure. To construct training data, we search for hints that, when prepended to the prompt, cause the target model to correct its previously incorrect response, so the generator learns what to say to prevent an error and not merely that an error is likely. Because failure modes recur systematically, it can anticipate them from the input alone. At inference time, the hint generator processes a new image--text pair, produces a hint, and passes it along with the original input to the black-box VLM in a single call (\Cref{fig:overview}).

\ours{} improves three frontier VLMs by 3.8 to 7.7 points on A-OKVQA and, without retraining, by 3.3 to 11.5 points on three unseen benchmarks. For two of the three targets it is significantly more accurate than chain-of-thought prompting, self-refinement, and an external judge on all four benchmarks. The same hint generator also improves VLMs that it was never trained on. Static and optimized task prompts, a small model's own answer, and a rationale written by the same small model all fall well short of these gains.

Small models have been trained to guide frozen black-box LLMs on text tasks with keywords or rationales \citep{li2023dsp,lee2024lmguidedcot,cheng2024black}. \ours{} starts instead from the errors of the target, which gives a different perspective on test-time scaling for VLMs. Because these errors are predictable from the input and largely shared across models, a single small generator can act on them before the model answers, for many targets and in a single call. Specifically, our contributions are:

\vspace{-.1in}
\begin{enumerate}
\item We show that the errors of frontier VLMs are shared across model families and predictable from the image and question alone, which makes them preventable before the model answers.
\item We propose \ours{}, a preemptive single-call TTS method for black-box VLMs, in which a lightweight hint generator prepends input-conditioned, failure-mode-anticipatory guidance to the target VLM's prompt before it answers. It is trained by agentic-search distillation and reinforcement learning on the targets' responses.
\item We show that \ours{} significantly improves the accuracy of multiple frontier VLMs and transfers zero-shot, with no retraining, to three unseen benchmarks and six unseen black-box and open-weight VLMs. Controls and four baselines built from the same data trace the gains to hints grounded in the visual evidence of each input and to training on verified repairs.
\end{enumerate}

\section{Related Work}
\label{sec:related}

{\bf Test-time scaling and post-hoc critique.} Test-time scaling methods treat inference compute as a resource to optimize rather than a fixed cost. For LLMs, gains have been achieved through chain-of-thought prompting \citep{wei2022cot}, self-consistency \citep{wang2023selfconsistency}, process reward models \citep{lightman2023letverify}, and best-of-$N$ sampling with learned verifiers \citep{snell2024scalingttc}. Analogous efforts for VLMs include multi-stage chain-of-thought reasoning \citep{zhang2023multimodalcot,xu2024llavacot,rose2023visualcot,lu2022scienceqa} and reward-model-guided search \citep{wang2025visualprm}. More recent methods sample the target several times \citep{avogaro2026sparc,kaya2026efficient,jeddi2026avis}. As evaluated, they rely on open weights or on internals such as token probabilities, gradients, or attention. A second family critiques and revises a response after it is given, with the critique coming from the target model itself \citep{madaan2023selfrefine,shinn2023reflexion}, from multi-agent debate \citep{du2023debate,liang2023debate}, from a separate LLM acting as a judge \citep{zheng2023llmasjudge}, or from critique models trained for the task \citep{luo2024criticv,xie2025ctrl}. Across all post-response methods, at least two inference passes through the target are required, with external-judge approaches adding a third call. \ours{} requires only black-box prompt access and replaces these additional calls with a single pass through a compact hint generator.

\noindent
{\bf Prompt optimization and adaptation.} Another line of work improves a frozen model by optimizing its prompt. APE \citep{zhou2023ape}, OPRO \citep{yang2024opro}, DSPy \citep{khattab2023dspy}, and GEPA \citep{agrawal2026gepa} search offline for instructions or demonstrations and apply them to all inputs of a task, which helps when the instruction is the bottleneck, and LEAP \citep{zhang2024leap} derives benchmark-level principles from mistakes on few-shot examples. In natural-image VQA, however, the target already understands the task, and its likely error depends on the specific image and question. Consistent with this, neither a failure-mode checklist nor GEPA reliably helps in our ablation (\Cref{sec:exp-ablation}). Other methods tune prompts using gradients, repeated queries, feedback, or the target's hidden states \citep{shu2022tpt,b2tpt2025,aozpt2026,vpae2026,zhang2022tempera}. \ours{} instead amortizes this search into a generator trained offline, which at test time needs no gradients, feedback, or per-query search, and produces an input-specific natural-language hint before a single target call.

\noindent
{\bf Learned guidance for black-box models.} Closest to \ours{} are methods that train a small model to guide a frozen black-box LLM on each input. DSP \citep{li2023dsp} generates instance-specific stimuli such as keywords, LM-Guided CoT \citep{lee2024lmguidedcot} writes a rationale from which the large model answers, and BPO \citep{cheng2024black} rewrites prompts using human preference data. \ours{} differs in where its guidance comes from. It starts from the errors of the target VLM. Its hints are verified with the target in the loop to repair them, and its reward is defined relative to the target's base answer and penalizes turning a correct answer into a wrong one. Its hints point to visual evidence and leave the reasoning to the target, whereas LM-Guided CoT supplies it. Prophet \citep{yu2025prophet} feeds candidate answers and confidence scores from a small VQA model to an LLM, and RICES \citep{yang2022pica,alayrac2022flamingo} retrieves similar training examples as demonstrations. We compare \ours{} with LM-Guided CoT, Prophet, RICES, and GEPA, each built from the same training data (\Cref{sec:exp-ablation}). An extended discussion of related work is in \Cref{sec:supp:related}.

\section{VLM Failures Are Shared and Predictable}
\label{sec:analysis}
One might expect that generating input-specific guidance toward the correct answer should be as hard as answering the question itself. We show that this is not the case: frontier VLMs exhibit a small set of recurring ``failure modes'' that are largely shared across models and predictable from the input alone, a structure that a simpler VLM can learn to anticipate for new multimodal inputs.

To establish this, we evaluate three proprietary models from different families: GPT-5 Nano (OpenAI), Gemini 2.5 Flash Lite (Google), and Claude 4.5 Haiku (Anthropic), on the A-OKVQA visual question answering benchmark (17,056 training questions) \citep{schwenk2022aokvqa}. Each model is queried three times per question with API-default temperature and top-$p$, and a question is classified as incorrect for a given model only if all three trials produce an incorrect answer. We ask two questions: do failure modes overlap across models, and is the failure mode predictable from the input alone?

\noindent
{\bf Cross-Model Failure Mode Overlap.} We define a taxonomy of 12 failure modes in VLMs: recognition, attribute binding, counting, spatial relation, optical character recognition, charts and tables, math/quantitative, knowledge, logic/negation, hallucination, instruction format, and ``other'' (full definitions in \Cref{sec:supp:taxonomy}). We use this taxonomy to categorize all error instances from various VLMs in the A-OKVQA dataset. Each incorrect response is labeled by an independent GPT-5 model with access to the image, question, ground-truth answer with rationale, and the model's incorrect response along with its reasoning.

\begin{figure}[ht]
  \centering
  \includegraphics[width=0.85\textwidth]{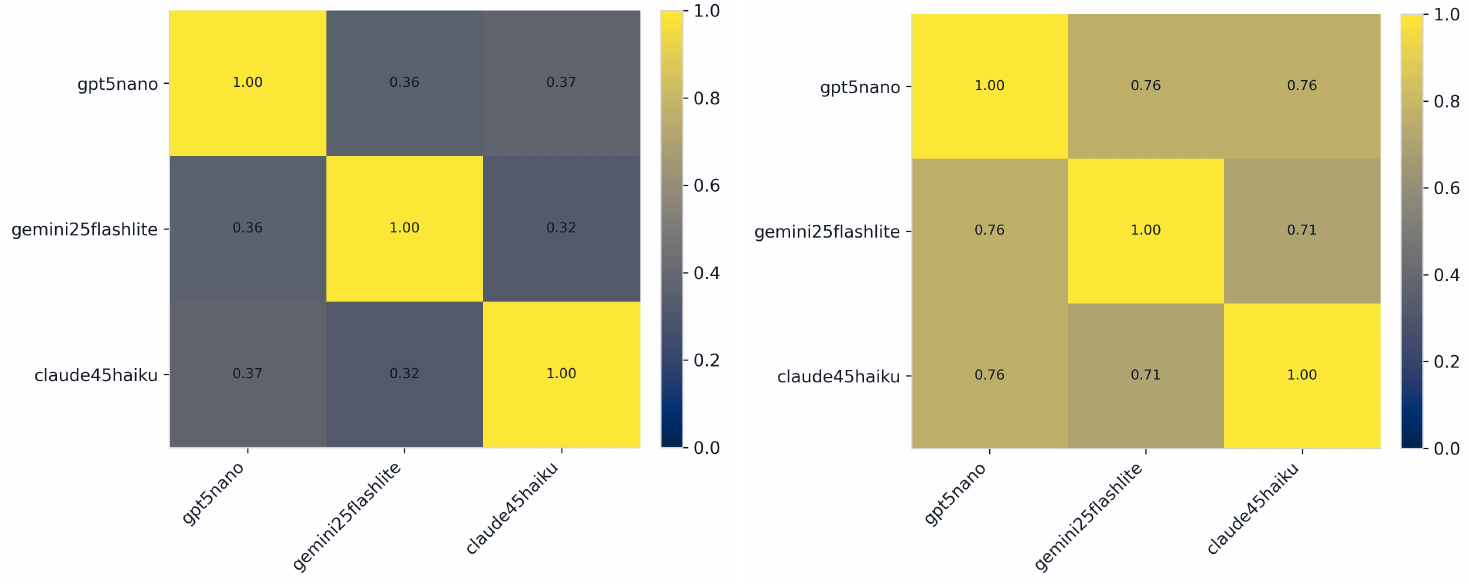}
  \vspace{-.1in}
  \caption{{\footnotesize \textbf{Cross-model error structure on A-OKVQA (training split).}
  \emph{Left:} Pairwise Jaccard overlap of base-incorrect question sets across the three frontier VLMs. \emph{Right:} Failure-mode agreement rate among shared errors, \ie{} the fraction of questions both models answer incorrectly that are assigned the same failure-mode label by the GPT-5 annotator.}}
  \label{fig:supp:overlap}
\end{figure}

Conditioned on shared errors among pairs of VLMs (i.e., questions both models answer incorrectly, with a pairwise Jaccard overlap of 0.32--0.37 between their incorrect question sets), we find that failure mode agreement is as high as 71--76\%: when a pair of VLMs both fail, they tend to fail for the same reason (\Cref{fig:supp:overlap}). This observation suggests that a hint targeting the anticipated failure mode for a given input could steer multiple models away from this common error.

\noindent
{\bf Failure Modes are Predictable from Inputs Alone.} In addition to retrospectively characterizing the VLM failure modes based on ground-truth answers and rationales, we also test whether these failure modes can be predicted from the inputs alone. To do so, we extract the final hidden state of Qwen3-VL-4B-Instruct (2560-dim) from its last transformer layer for each image--question pair, supplemented with 22 scalar uncertainty features derived from its own output distribution (e.g., self-reported confidence, log-probability statistics, entropy, and calibration signals). No output of the target VLMs is used as input. A hidden encoder projects the 2560-dim representation through two fully connected layers ($2560 \to 256 \to 64$), while a scalar encoder maps the 22 uncertainty features to a 32-dim embedding, yielding a 96-dim concatenated representation passed to a task-specific classification head.

A binary classifier ($96 \to 64 \to 1$) trained with this representation achieves AUROC of 0.79--0.81 for predicting whether each target VLM will answer a given question incorrectly. Additionally, a shared-backbone variant in which a single encoder supplies VLM-specific binary heads simultaneously yields very similar performance (\Cref{tab:supp:risk}). A 12-class classifier using the same architecture (with inverse-frequency class weighting to handle label imbalance) achieves top-2 accuracy of 73--78\%, meaning the true failure mode can be narrowed to two candidates with high accuracy (\Cref{tab:supp:failmode}). Together, these results suggest that a lightweight model can learn to predict not only \textit{when} a VLM will fail but \textit{how}, and that this capacity generalizes across architecturally distinct target VLMs. These two properties shape our method. Predictable errors let a small model act before the target answers, and shared errors let one generator serve many targets.

\noindent
Further details of the experiments above are provided in \Cref{sec:supp:overlap}.

\begin{table}[t]
\small
\centering
\setlength{\tabcolsep}{3.8pt}
\renewcommand{\arraystretch}{1.12}
\begin{tabular}{l c c c c c c c c}
\toprule
& \multicolumn{2}{c}{AUROC ($\uparrow$)} &
\multicolumn{2}{c}{Max-F1 ($\uparrow$)} &
\multicolumn{2}{c}{Precision ($\uparrow$)} &
\multicolumn{2}{c}{Recall ($\uparrow$)} \\
\cmidrule(lr){2-3}\cmidrule(lr){4-5}\cmidrule(lr){6-7}\cmidrule(lr){8-9}
Target Model &
\makecell[c]{Indiv.\\MLP} & \makecell[c]{Shared\\MLP} &
\makecell[c]{Indiv.\\MLP} & \makecell[c]{Shared\\MLP} &
\makecell[c]{Indiv.\\MLP} & \makecell[c]{Shared\\MLP} &
\makecell[c]{Indiv.\\MLP} & \makecell[c]{Shared\\MLP} \\
\midrule
Claude 4.5 Haiku      & \textbf{0.79} & 0.78 & 0.54 & \textbf{0.54} & 0.41 & \textbf{0.43} & \textbf{0.76} & 0.73 \\
Gemini 2.5 Flash Lite & \textbf{0.81} & 0.80 & \textbf{0.46} & 0.44 & \textbf{0.36} & 0.33 & 0.61 & \textbf{0.68} \\
GPT-5 Nano            & 0.79 & \textbf{0.80} & 0.36 & \textbf{0.36} & 0.25 & \textbf{0.28} & \textbf{0.67} & 0.52 \\
\bottomrule
\end{tabular}
\caption{\small \textbf{Binary risk prediction.} \emph{Indiv.~MLP}: one predictor per target model. \emph{Shared MLP}: single predictor for all target models.}
\label{tab:supp:risk}
\end{table}

\begin{table}[t]
\centering
\small
\setlength{\tabcolsep}{4.2pt}
\renewcommand{\arraystretch}{1.12}
\begin{tabular}{l c c c c c}
\toprule
Target Model & Top-1 ($\uparrow$) & Top-2 ($\uparrow$) & Macro-F1 ($\uparrow$) & Wt-F1 ($\uparrow$) & BalAcc ($\uparrow$) \\
\midrule
Claude 4.5 Haiku      & 0.54 & 0.73 & 0.40 & 0.52 & 0.484 \\
Gemini 2.5 Flash Lite & 0.50 & 0.78 & 0.30 & 0.47 & 0.434 \\
GPT-5 Nano            & 0.66 & 0.78 & 0.47 & 0.66 & 0.311 \\
\bottomrule
\end{tabular}%
\caption{\small \textbf{Failure-mode prediction} (12-class classification).}
\label{tab:supp:failmode}
\end{table}

\section{Test-Time Hinting for Black-Box VLMs}
\label{sec:method}

Motivated by the observations in the previous section, we propose a method that moves from predicting an error to acting on it. Its hints are built from observed failures of each target VLM, and a generator is trained to pre-empt such failures from the input alone. We first establish notation before describing the method.

\noindent
{\bf Notation.} We use the following notation throughout the paper:
\begin{itemize}[leftmargin=*,topsep=2pt,itemsep=2pt]
    \item $x$, $q$: input image and question/prompt (the latter includes answer options for multiple-choice).
    \item $\hat{a}$, $\hat{r}$: a VLM's answer and the corresponding reasoning trace.
    \item $a^{*}$, $r^{*}$: the ground-truth answer and the corresponding rationale.
    \item $\mathcal{M}$: a \emph{target VLM} whose errors we aim to fix using our method. We write $\hat{a} = \mathcal{M}(x, q)$ for its baseline response on input $(x, q)$.
    \item $m$: a natural-language \emph{model identifier} for $\mathcal{M}$ (\eg{} ``\texttt{GPT-5 Nano}''), passed as an input token to the hint generator when multiple targets are considered.
    \item $h = \{\text{\texttt{"hint"}}: [s_1, \dots, s_k]\}$ with $k \in \{1, 2, 3\}$: a \emph{hint} formatted as a short JSON list of natural-language items prepended to $\mathcal{M}$'s prompt. At least one item must be \emph{contrastive}, prompting the model to discriminate between two plausible alternatives without revealing the answer. This contrastive form is central to the design, since it forces hints to surface perceptual or procedural cues rather than to encode the answer itself. $h$ must not reveal $a^{*}$.
    \item $\mathcal{G}_\theta$: the \emph{hint generator}, a small VLM with parameters $\theta$ that produces $h$ from $(x, q, m)$.
\end{itemize}

\begin{figure}[t]
  \centering
  \includegraphics[width=\textwidth]{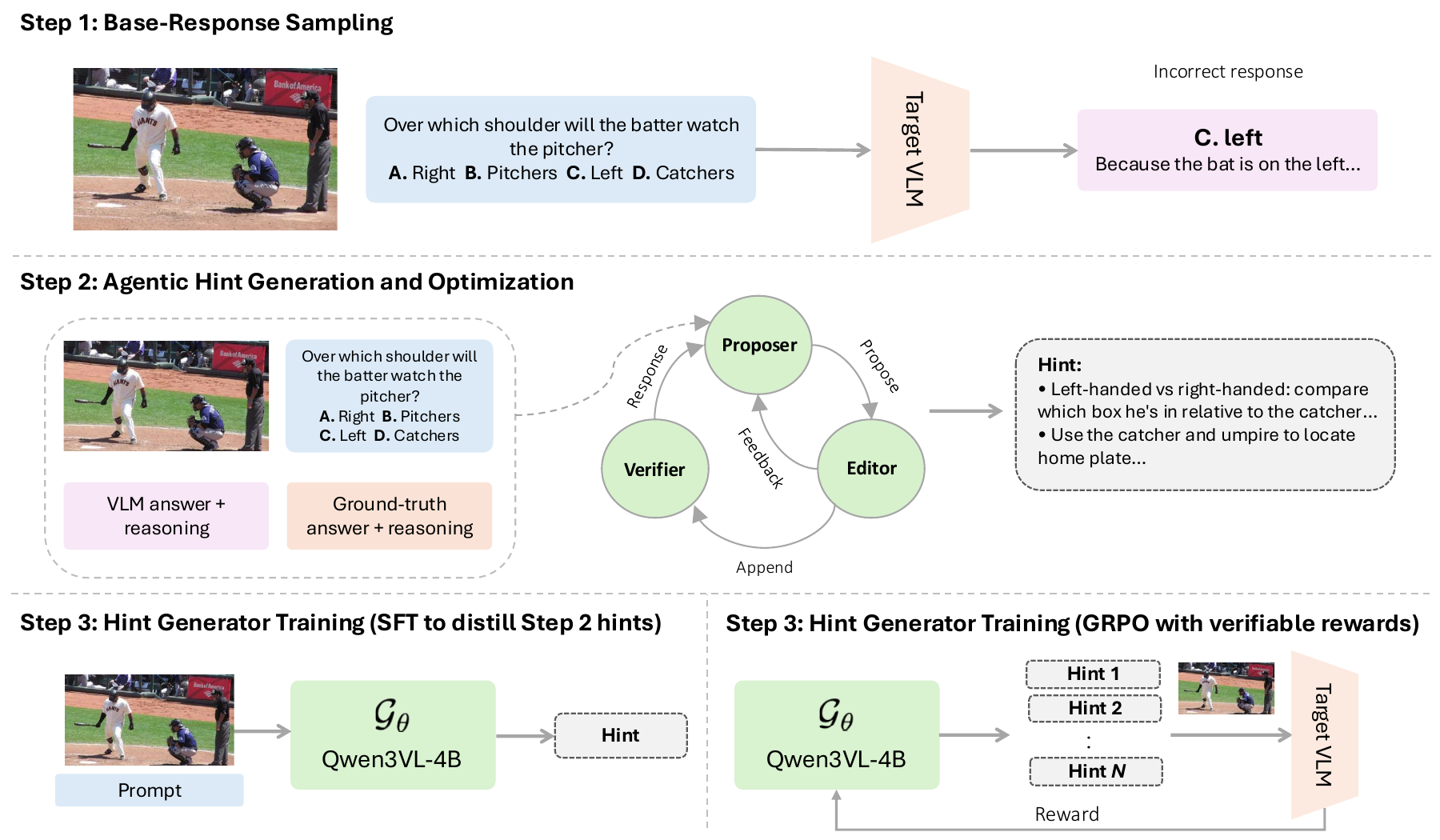}
  \caption{{\small \textbf{Development pipeline of \ours{}.}}}
  \label{fig:pipeline}
  \vspace{-.1in}
\end{figure}

\noindent
Our Test-Time Hinting approach (TTH) involves three steps (\Cref{fig:pipeline}): (1) sampling base responses from each target VLM, (2) eliciting per-question hints through an agentic loop, and (3) distilling these hints into $\mathcal{G}_\theta$ via supervised fine-tuning followed by a reinforcement-learning (RL) refinement stage.

\subsection{Step 1: Base-Response Sampling}
\label{sec:phase1}
We run each target VLM on every training question with three independent passes at API-default temperature and top-p. Each question-model pair is classified as \emph{base-correct} (correct on all three trials), \emph{base-incorrect} (incorrect on all three trials with a valid parse), or \emph{mixed} (otherwise). Only base-correct and base-incorrect pairs are then used to train the hint generator in the following steps.

\subsection{Step 2: Agentic Hint Generation and Optimization}
\label{sec:phase2}
Given an input $(x, q)$ for which the VLM $\mathcal{M}$ produces an incorrect answer $\hat{a} =\mathcal{M}(x,q) \neq a^*$, we construct a hint $h$ such that prepending it to the same image-question pair yields the correct answer, i.e., $\mathcal{M}(x,[h;q])=a^*$. To generate such hints, we design an iterative agentic pipeline comprising three roles, a Proposer, an Editor, and a Verifier, detailed below (pseudocode provided in \Cref{sec:supp:agentic}).

Our agentic process handles two hint types: \emph{repair} hints for base-incorrect questions, where it seeks a hint that flips the target to the correct answer, and \emph{reinforcement} hints for base-correct questions, where it seeks a hint that retains the correct answer while keeping attention on the relevant visual evidence. A \textbf{proposer} $\mathcal{P}$ (GPT-5) with full privileged context $(x, q, a^{*}, r^{*}, \hat{a}, \hat{r})$ proposes a candidate hint. It is instructed to target the failure mode seen in the target's response, using the visual context and without revealing the answer. An \textbf{editor} $\mathcal{E}$ (GPT-5) modifies and revises the hint to enforce certain criteria (no direct leakage of $a^*$, at least one contrastive item, no certainty language, avoiding extraneous reasoning steps). The target VLM $\mathcal{M}$ then acts as the \textbf{verifier} by checking that the answer $\hat{a}$ corresponding to $(x, [h; q])$ matches $a^*$. The loop runs for up to $R_{\max}{=}3$ rounds. For repair questions it terminates on the first correct re-answer and otherwise retains the last editor-approved hint as an \emph{unsuccessful repair} target, which is excluded from downstream phases. For reinforcement hints, the system likewise terminates on the first correct re-answer, and trials in which every round disrupts the originally correct answer are discarded entirely.

\subsection{Step 3: Hint Generator Training}
\label{sec:phase3}
In Step 3, we train a lightweight model $\mathcal{G}_\theta$ to generate the hints $h$ elicited in Step 2 conditioned on the triplet $(x, q, m)$ (i.e., the only information available at inference time). We initialize $\mathcal{G}_\theta$ from Qwen3-VL-4B-Instruct \citep{bai2025qwen3vl} and attach LoRA adapters \citep{hu2022lora} to the language-model backbone, the vision-language merger, and the last two blocks of the vision encoder. The full set of modules and training hyperparameters for both stages, including LoRA rank, learning rates, and optimization schedule, is provided in \Cref{sec:supp:training_hparams}. Training proceeds in two stages: supervised fine-tuning using the hints generated in Step~2, followed by a reinforcement learning step that directly optimizes the actual downstream behavior of the hint on the target VLMs (i.e., whether it flips incorrect answers). We show in \Cref{sec:exp-ablation} how both stages empirically contribute to $\mathcal{G}_\theta$.

\noindent
{\bf Stage~1: Supervised fine-tuning on agentic hints.} Each training instance is a tuple $(x, q, m, h^{*})$, where $m$ is the natural-language target identifier and $h^{*}$ is a hint produced in Step~2. We retain only Step~2 hint generations that lead to correct answers: the hint either flipped a base error to a correct answer (a successful repair) or preserved an already-correct answer (a successful reinforcement). Trials in which the verifier never reached a correct answer are discarded. Their unsuccessful-repair hints would supervise the generator on guidance whose efficacy was never verified end-to-end, and Stage~2 picks up these harder cases by directly observing downstream behavior. The remaining pool of hints is balanced by target identifier via median-count resampling so that no single target dominates updates, and we minimize next-token cross-entropy on the hint JSON.

\noindent
{\bf Stage~2: Reinforcement learning with verifiable rewards.} Stage~1 supervises the generator to \emph{imitate} hints that worked in Step~2, but it does not directly optimize the generator's own samples for downstream effect on the target VLM. Stage~2 closes this gap. For each training prompt $(x, q, m)$ the policy samples a small group of candidate hints. Each candidate is sent to the target VLM named by $m$, and we compare that target's hinted answer to its base (unhinted) answer, classifying the outcome as one of: \emph{repair} (a previous error becomes correct), \emph{no-op} (a correct answer is preserved), \emph{harm} (a correct answer becomes wrong), or \emph{unrepaired} (a previous error remains wrong). The outcome is mapped to a scalar reward ($+1$, $0$, $-1$, and $-0.5$, respectively) for Group Relative Policy Optimization (GRPO) \citep{shao2024deepseekmath,guo2025deepseek}, KL-anchored to the Stage-1-trained reference. Details of the reward design choices are provided in \Cref{sec:supp:reward_table}.

\section{Experiments}
\label{sec:experiments}

\subsection{Experimental Setup}
\label{sec:exp-setup}

{\bf Datasets.} Our primary development and evaluation benchmark is \textbf{A-OKVQA} \citep{schwenk2022aokvqa}, a knowledge-based VQA dataset with ground-truth rationales. Because its official test labels are not public, we use the validation split of 1{,}145 questions as our held-out \emph{test set}. All model development is performed on the 17{,}056-question training split, with 855 questions reserved as an internal \emph{validation set} for hyperparameter selection and early stopping. The test set is used only for final model evaluation. To test whether $\mathcal{G}_\theta$ learns general failure-anticipation strategies rather than dataset-specific patterns, we additionally evaluate zero-shot transfer (no retraining or domain adaptation) on three additional natural-image benchmarks: \textbf{VCR} \citep{zellers2019vcr} (5{,}000 unique image-question pairs sampled from the validation set), \textbf{Visual7W} \citep{zhu2016visual7w} (5{,}000 sampled test pairs), and \textbf{RealWorldQA} \citep{xai2024realworldqa,xai_realworldqa_2024} (628 four-way MCQ / yes-no / left-right / color questions retained from the original 765 for consistent MCQ-style evaluation). All four benchmarks are multiple-choice, so correctness is exact and repair, harm, and answer leakage can be measured directly. Together they span knowledge-based VQA, visual commonsense reasoning, grounded perception questions, and real-world spatial understanding (\Cref{sec:supp:eval}).

\noindent
{\bf Target models.} Our primary target models are three proprietary frontier VLMs from different families: Claude 4.5 Haiku (Anthropic), Gemini 2.5 Flash Lite (Google), and GPT-5 Nano (OpenAI). The hint generator is trained on these three targets and additionally evaluated on five unseen closed-weight targets and one open-weight target without retraining (\Cref{sec:exp-transfer}). For \emph{same-family transfer}, we apply hints generated for each training model to a larger model from the same family: GPT-5 Nano hints are reused for GPT-5.2, Gemini 2.5 Flash Lite hints for Gemini 2.5 Flash, and Claude 4.5 Haiku hints for Claude 4.5 Opus. For the other unseen targets, $m$ is the model's own name. All target models are evaluated using their default API hyperparameters.

\noindent
{\bf Baselines.} We compare \ours{} (1~target call) with:
(i)~\textbf{Base} (1~call): direct prompting with the question and image;
(ii)~\textbf{CoT} \citep{wei2022cot,kojima2022zeroshot} (1~call): the base prompt augmented with ``Think step-by-step. Then output ONLY the JSON object'';
(iii)~\textbf{Self-refine} \citep{madaan2023selfrefine} (2~calls): the target answers, then is shown its own response and asked to critique and optionally correct it; (iv)~\textbf{External judge} \citep{du2023debate,zheng2023llmasjudge} (2~target calls + 1~judge call): a more capable model from the same family critiques the target's answer, and the target re-answers accordingly. Judges are GPT-5-Mini for GPT-5 Nano, Gemini 2.5 Flash for Gemini 2.5 Flash Lite, and Claude 4.5 Sonnet for Claude 4.5 Haiku (judges of the unseen targets are listed in \Cref{sec:supp:eval}).

\noindent
{\bf Metrics.} We report {\it three} metrics. \emph{Overall accuracy} is the fraction of test questions answered correctly. The \emph{repair rate}, $\Pr[\hat{a}_h\!=\!a^{*} \mid \hat{a}\!\neq\!a^{*}]$, restricts to the questions the base model answered incorrectly and reports what fraction each method corrects. The \emph{harm rate}, $\Pr[\hat{a}_h\!\neq\!a^{*} \mid \hat{a}\!=\!a^{*}]$, restricts to base-correct questions and reports the fraction that flip to incorrect after applying each method. Repair rate and harm rate are the two complementary axes along which a hinting method's deployment value is most directly measured: a useful method must both correct previous errors \emph{and} avoid disrupting answers the target already had right. The two rates determine the change in overall accuracy, $\Delta\mathrm{Acc}=(1-B)R-BH$ for base accuracy $B$, repair rate $R$, and harm rate $H$, so overall accuracy is our primary criterion and the two rates are its diagnostic components. We assess statistical significance with a joint paired nonparametric bootstrap over test questions (2{,}000 replicates), where a difference is significant if its 95\% confidence interval excludes zero. \Cref{sec:supp:eval} reports 95\% confidence intervals for all results. API responses vary between calls at fixed settings, which is a known property of hosted models. We therefore also report a second unhinted Base sample as a reference (\Cref{tab:ablation}).

\noindent
{\bf Code and prompts.} Code and the prompts for all agents, baselines, and target models are available at \url{https://github.com/AlaaLab/test-time-hinting}.

\subsection{Main Results Across Benchmarks}
\label{sec:exp-main}

\Cref{tab:main} reports overall accuracy, repair rate, and harm rate for the three primary target models. On the A-OKVQA test set, \ours{} improves accuracy by 3.8 to 7.7 points over Base and achieves the highest repair rate on \emph{every} target, far above the 6 to 16\% that a second unhinted sample repairs (\Cref{tab:ablation}). The same checkpoint is then applied without retraining to VCR, Visual7W, and RealWorldQA. The benchmarks differ from A-OKVQA in question style and answer distribution, and VCR and RealWorldQA also in image domain (movie stills and mostly driving scenes), providing a stringent test of out-of-distribution generalization. \ours{} gains 3.3 to 11.5 points on them, confirming that the gains are not specific to A-OKVQA, and the largest lift appears on RealWorldQA, the benchmark furthest in domain from A-OKVQA.

Across the 12 target and benchmark cells, \ours{} is significantly more accurate than Base in every cell, attains the highest accuracy in 9, and is significantly more accurate than every baseline in 8, namely for Claude 4.5 Haiku and Gemini 2.5 Flash Lite on all four benchmarks. The external judge exceeds \ours{} only for GPT-5 Nano, where the judge is GPT-5-Mini. Since self-refine provides no comparable advantage on the same target, the gap reflects the capability asymmetry between GPT-5 Nano and the much stronger judge, not the preemptive vs.\ post-hoc distinction. The pattern reverses for GPT-5.2 (\Cref{tab:unseen}), whose judge (GPT-5.2-Pro) is closer in capability: \ours{} outperforms all baselines including the external judge. Where the external judge wins repair rate but \ours{} still wins overall accuracy, the judge harms more originally correct answers: aggressive repair coverage and overall accuracy are not the same objective. Qualitative examples of repairs are shown in \Cref{sec:supp:qualitative}.

\begin{table}[t]
\caption{{\small \textbf{Results on four benchmarks.} $\mathcal{G}_\theta$ is trained on A-OKVQA only and applied to the other three benchmarks without retraining. \textbf{Bold}: best per row and metric; \uline{underline}: second best; $^{\dagger}$: significantly more accurate than Base and every baseline.}}
\label{tab:main}
\centering
\small
\setlength{\tabcolsep}{4.0pt}
\renewcommand{\arraystretch}{1.10}
\resizebox{\textwidth}{!}{%
\begin{tabular}{l l c c c c c c c c c c c c c}
\toprule
& & \multicolumn{5}{c}{Overall Accuracy (\%, $\uparrow$)} &
\multicolumn{4}{c}{Repair Rate (\%, $\uparrow$)} &
\multicolumn{4}{c}{Harm Rate (\%, $\downarrow$)} \\
\cmidrule(lr){3-7}\cmidrule(lr){8-11}\cmidrule(lr){12-15}
Dataset & Target Model &
\makecell[c]{Base} &
\makecell[c]{CoT} &
\makecell[c]{Self\\refine} &
\makecell[c]{External\\judge} &
\makecell[c]{\ours{}\\(Ours)} &
\makecell[c]{CoT} &
\makecell[c]{Self\\refine} &
\makecell[c]{External\\judge} &
\makecell[c]{\ours{}\\(Ours)} &
\makecell[c]{CoT} &
\makecell[c]{Self\\refine} &
\makecell[c]{External\\judge} &
\makecell[c]{\ours{}\\(Ours)} \\
\midrule
\multirow{3}{*}{A-OKVQA} & Claude 4.5 Haiku & 78.25 & 80.26 & 77.90 & \uline{80.87} & \textbf{84.28}\rlap{$^{\dagger}$} & 26.10 & 30.52 & \uline{44.58} & \textbf{49.40} & \textbf{4.69} & 8.93 & 9.04 & \uline{6.03} \\
 & Gemini 2.5 Flash Lite & 82.88 & 84.10 & \uline{87.69} & 86.81 & \textbf{90.57}\rlap{$^{\dagger}$} & 21.43 & 35.71 & \uline{43.37} & \textbf{61.73} & \uline{2.95} & \textbf{1.58} & 4.21 & 3.48 \\
 & GPT-5 Nano & 84.37 & 82.79 & 85.24 & \textbf{89.43} & \uline{88.12} & 18.44 & 17.88 & \uline{44.69} & \textbf{45.81} & 5.28 & \textbf{2.28} & \textbf{2.28} & \uline{4.04} \\
\midrule
\multirow{3}{*}{VCR} & Claude 4.5 Haiku & 61.64 & 62.06 & 62.20 & \uline{62.60} & \textbf{64.98}\rlap{$^{\dagger}$} & 15.12 & 11.47 & \uline{21.38} & \textbf{34.41} & \uline{8.73} & \textbf{6.23} & 11.75 & 16.00 \\
& Gemini 2.5 Flash Lite & 61.68 & 63.02 & \uline{65.40} & 63.66 & \textbf{67.04}\rlap{$^{\dagger}$} & 14.56 & 14.56 & \uline{19.26} & \textbf{23.17} & 6.87 & \textbf{3.02} & 8.75 & \uline{5.71} \\
& GPT-5 Nano & 63.00 & 64.04 & 63.20 & \uline{68.64} & \textbf{68.68} & 18.81 & 11.89 & \uline{21.62} & \textbf{24.22} & 9.40 & 6.67 & \textbf{3.75} & \uline{5.21} \\
\midrule
\multirow{3}{*}{Visual7W} & Claude 4.5 Haiku & 71.60 & 74.68 & 73.86 & \uline{78.20} & \textbf{81.26}\rlap{$^{\dagger}$} & 21.13 & 25.35 & \textbf{47.89} & \uline{37.04} & \uline{4.08} & 6.90 & 9.78 & \textbf{1.20} \\
& Gemini 2.5 Flash Lite & 75.44 & 78.48 & 81.84 & \uline{82.66} & \textbf{85.22}\rlap{$^{\dagger}$} & 20.36 & 31.68 & \textbf{45.52} & \uline{45.28} & 2.60 & \uline{1.83} & 5.25 & \textbf{1.78} \\
& GPT-5 Nano & 80.20 & 81.86 & 83.82 & \textbf{85.08} & \uline{84.44} & 25.15 & 26.26 & \textbf{41.41} & \uline{38.79} & 4.14 & \textbf{1.97} & \uline{4.14} & 4.29 \\
\midrule
\multirow{3}{*}{RealWorldQA} & Claude 4.5 Haiku & 62.74 & 60.67 & \uline{63.85} & 60.83 & \textbf{69.11}\rlap{$^{\dagger}$} & 10.26 & 27.35 & \textbf{49.15} & \uline{31.20} & \uline{9.39} & 14.47 & 32.23 & \textbf{8.38} \\
& Gemini 2.5 Flash Lite & 65.13 & 66.72 & 68.15 & \uline{71.34} & \textbf{76.59}\rlap{$^{\dagger}$} & 13.24 & 14.16 & \uline{34.25} & \textbf{34.70} & 4.65 & \uline{2.93} & 8.80 & \textbf{0.98} \\
& GPT-5 Nano & 70.06 & 71.50 & 71.97 & \textbf{80.25} & \uline{75.64} & \uline{27.13} & 26.60 & \textbf{49.47} & 26.60 & 9.55 & 8.64 & \uline{6.59} & \textbf{3.41} \\
\bottomrule
\end{tabular}%
}
\end{table}

\subsection{Zero-Shot Transfer to Unseen Target Models}
\label{sec:exp-transfer}

A core design goal of \ours{} is that hints optimized for a set of training models should generalize to unseen targets. \Cref{tab:unseen} applies the same checkpoint, without any retraining or adaptation, to five closed-weight VLMs that were never used in training. Three are larger models of the training families and two come from providers unseen in training. \ours{} achieves both the best overall accuracy and the best repair rate on every one of the five unseen targets, repairing base errors at 1.4 to 3.0 times the rate of the next-best baseline, and its gain over Base is significant for Grok 4 Fast Reasoning. These targets already answer 84 to 92\% of the A-OKVQA questions correctly, which leaves few errors to repair for any method. Across all target--benchmark pairs, gains shrink with base accuracy because fewer errors remain, not because correction weakens (\Cref{sec:supp:headroom}). The small margins on these targets therefore reflect a saturated benchmark rather than a limit of the method.

\begin{table}[t]
\caption{{\small \textbf{Zero-shot transfer to unseen closed-weight target models.} No target is used in training. $^{\ast}$ marks providers unseen in training. On RealWorldQA the benchmark is unseen as well. Markers as in \Cref{tab:main}.}}
\label{tab:unseen}
\centering
\small
\setlength{\tabcolsep}{4.0pt}
\renewcommand{\arraystretch}{1.10}
\resizebox{\textwidth}{!}{%
\begin{tabular}{l l c c c c c c c c c c c c c}
\toprule
& & \multicolumn{5}{c}{Overall Accuracy (\%, $\uparrow$)} &
\multicolumn{4}{c}{Repair Rate (\%, $\uparrow$)} &
\multicolumn{4}{c}{Harm Rate (\%, $\downarrow$)} \\
\cmidrule(lr){3-7}\cmidrule(lr){8-11}\cmidrule(lr){12-15}
Dataset & Target Model &
\makecell[c]{Base} &
\makecell[c]{CoT} &
\makecell[c]{Self\\refine} &
\makecell[c]{External\\judge} &
\makecell[c]{\ours{}\\(Ours)} &
\makecell[c]{CoT} &
\makecell[c]{Self\\refine} &
\makecell[c]{External\\judge} &
\makecell[c]{\ours{}\\(Ours)} &
\makecell[c]{CoT} &
\makecell[c]{Self\\refine} &
\makecell[c]{External\\judge} &
\makecell[c]{\ours{}\\(Ours)} \\
\midrule
\multirow{5}{*}{A-OKVQA} & Claude 4.5 Opus & 91.00 & 90.92 & 91.09 & \uline{91.27} & \textbf{91.62} & \uline{13.59} & 5.83 & 12.62 & \textbf{37.86} & 1.44 & \textbf{0.48} & \uline{0.96} & 3.07 \\
 & Gemini 2.5 Flash & \uline{91.88} & 91.53 & \uline{91.88} & 91.44 & \textbf{92.05} & \uline{12.90} & 7.53 & 6.45 & \textbf{27.96} & 1.52 & \textbf{0.67} & \uline{1.05} & 2.28 \\
 & GPT-5.2 & 90.22 & 89.17 & \uline{90.83} & 90.48 & \textbf{91.09} & 10.71 & \uline{20.54} & 2.68 & \textbf{30.36} & 2.32 & \uline{1.55} & \textbf{0.00} & 2.32 \\
 & Grok 4 Fast Reasoning$^{\ast}$ & 85.33 & 84.89 & \uline{86.90} & 86.38 & \textbf{87.60} & 7.14 & 11.31 & \uline{14.29} & \textbf{43.45} & 1.74 & \textbf{0.10} & \uline{1.23} & 4.81 \\
 & Nova Pro$^{\ast}$ & 83.76 & 83.58 & \uline{84.37} & 83.32 & \textbf{84.98} & 13.44 & 22.04 & \uline{30.11} & \textbf{43.55} & \textbf{2.82} & \uline{3.55} & 6.36 & 6.99 \\
\midrule
\multirow{3}{*}{RealWorldQA} & Claude 4.5 Opus & 76.91 & 76.11 & 77.07 & \textbf{78.18} & \uline{78.03} & 8.28 & 3.45 & \uline{14.48} & \textbf{33.79} & 3.52 & \textbf{0.83} & \uline{2.69} & 8.70 \\
 & Gemini 2.5 Flash & 76.75 & \uline{77.71} & 77.55 & \uline{77.71} & \textbf{81.85}\rlap{$^{\dagger}$} & \uline{15.75} & 4.79 & 4.11 & \textbf{33.56} & 3.53 & \uline{0.41} & \textbf{0.00} & 3.53 \\
 & GPT-5.2 & 77.23 & 77.23 & \uline{78.18} & 77.07 & \textbf{83.44}\rlap{$^{\dagger}$} & 16.08 & \uline{20.28} & 0.70 & \textbf{49.65} & \uline{4.74} & \uline{4.74} & \textbf{0.41} & 6.60 \\
\bottomrule
\end{tabular}%
}
\end{table}

\noindent
{\bf The same targets on a less saturated benchmark.} We therefore move the three same-family targets to RealWorldQA, where they reach only about 77\% at Base. This is our hardest transfer setting, because the questions are harder and neither the benchmark nor the targets were seen in training. With more errors left to repair, \ours{} is on par with the strongest baseline for Claude 4.5 Opus and significantly more accurate than Base and every baseline for Gemini 2.5 Flash and GPT-5.2, with gains of 5.1 and 6.2 points over Base.

\noindent
{\bf Open-weight target.} \ours{} needs only black-box access, so the same generator also applies to open-weight models, where it can be compared with test-time methods that use model internals. As a check, we apply $\mathcal{G}_\theta$ without retraining to the open-weight Inkling-Small \citep{tml2026inkling}, whose name is passed as the target identifier and never appears in training. Besides the four baselines we run VISER \citep{izadi2025viser} and TTAug \citep{kaya2026efficient}, which averages the next-token probabilities of 16 augmented inputs at every decoding step. \ours{} is significantly more accurate than Base and every other method on all four benchmarks (\Cref{tab:openweight}). It gains 6.4 to 15.5 points over Base with a single target call, whereas TTAug gains 1.4 to 5.9 points with 16 forward passes.

\begin{table}[t]
\caption{{\small \textbf{Open-weight target (Inkling-Small),} without retraining $\mathcal{G}_\theta$. Accuracy (\%). Passes are target forward passes per query. $^{\dagger}$ marks accuracy significantly above Base and every other method.}}
\label{tab:openweight}
\centering
\small
\setlength{\tabcolsep}{6pt}
\renewcommand{\arraystretch}{1.10}
\begin{tabular}{l c cccc}
\toprule
Method & Passes & A-OKVQA & VCR & Visual7W & RealWorldQA \\
\midrule
Base & 1 & 70.39 & 50.22 & 67.38 & 67.04 \\
CoT \citep{wei2022cot} & 1 & 69.34 & 51.28 & 68.46 & 65.45 \\
Self-refine \citep{madaan2023selfrefine} & 2 & 75.11 & 51.44 & 71.26 & 67.99 \\
External judge \citep{zheng2023llmasjudge} & 2+1 & 75.72 & 53.86 & 74.96 & 69.11 \\
VISER \citep{izadi2025viser} & 1 & 67.34 & 51.40 & 68.92 & 66.72 \\
TTAug \citep{kaya2026efficient} & 16 & 76.33 & 54.30 & 72.56 & 68.47 \\
\midrule
\ours{} (Ours) & 1 & \textbf{85.85}\rlap{$^{\dagger}$} & \textbf{57.74}\rlap{$^{\dagger}$} & \textbf{79.18}\rlap{$^{\dagger}$} & \textbf{73.41}\rlap{$^{\dagger}$} \\
\bottomrule
\end{tabular}
\end{table}

\subsection{Cost of an Accuracy Point}
\label{sec:exp-cost}

Another way to raise the accuracy of a target is to switch to a stronger model of the same family. \Cref{tab:costpoint} compares this with adding \ours{} to the cheaper model. Costs are API costs in USD per 1{,}000 questions. The cost of \ours{} includes the hint in the target's prompt and the pass of $\mathcal{G}_\theta$, priced at the pay-per-token rate of a hosted Qwen3-VL-4B endpoint (0.04 USD per million input tokens and 0.07 per million output tokens), which adds about 0.03 on A-OKVQA and 0.07 on RealWorldQA. The cost of an accuracy point is the extra cost over the cheaper model divided by the accuracy gained over it, with the accuracies of \Cref{tab:main,tab:unseen}.

\begin{table}[t]
\caption{{\small \textbf{Cost of an accuracy point with \ours{} and by switching to a stronger model of the same family.} Costs are in USD per 1{,}000 questions. The last column gives how many times cheaper an accuracy point is with \ours{} than by switching, computed from the listed costs.}}
\label{tab:costpoint}
\centering
\footnotesize
\setlength{\tabcolsep}{5pt}
\renewcommand{\arraystretch}{1.05}
\resizebox{\textwidth}{!}{%
\begin{tabular}{lll rrr rrr}
\toprule
& & & \multicolumn{3}{c}{Cost} & \multicolumn{3}{c}{Cost per accuracy point} \\
\cmidrule(lr){4-6}\cmidrule(lr){7-9}
Benchmark & Cheaper model & Stronger model & Cheaper & \makecell{Cheaper\\with \ours{}} & Stronger & \makecell{Adding\\\ours{}} & \makecell{Switching\\model} & \makecell{Times cheaper\\with \ours{}} \\
\midrule
\multirow{3}{*}{A-OKVQA} & Claude 4.5 Haiku & Claude 4.5 Opus & 0.90 & 0.95 & 9.09 & 0.008 & 0.642 & 77$\times$ \\
& GPT-5 Nano & GPT-5.2 & 0.08 & 0.12 & 3.32 & 0.011 & 0.554 & 52$\times$ \\
& Gemini 2.5 Flash Lite & Gemini 2.5 Flash & 0.21 & 0.26 & 1.18 & 0.007 & 0.108 & 17$\times$ \\
\midrule
\multirow{3}{*}{RealWorldQA} & Claude 4.5 Haiku & Claude 4.5 Opus & 2.35 & 2.82 & 15.86 & 0.074 & 0.953 & 13$\times$ \\
& GPT-5 Nano & GPT-5.2 & 0.17 & 0.26 & 6.55 & 0.016 & 0.890 & 55$\times$ \\
& Gemini 2.5 Flash Lite & Gemini 2.5 Flash & 0.47 & 0.55 & 1.38 & 0.007 & 0.078 & 11$\times$ \\
\bottomrule
\end{tabular}%
}
\end{table}

On both benchmarks and in all three families, an accuracy point costs more than 10 times less with \ours{} than by switching to the stronger model. \ours{} is also complementary to a stronger model, since it adds accuracy to cheaper and stronger targets alike (\Cref{tab:unseen}).

\subsection{Ablations and Analysis of the Sources of Gain}
\label{sec:exp-ablation}

To isolate where the gains in \ours{} come from, we test four alternative explanations on the A-OKVQA test set (\Cref{tab:ablation}) and audit the hints for answer leakage. Setups are given in \Cref{sec:supp:controls}.
\begin{itemize}[leftmargin=*,topsep=2pt,itemsep=1pt]
  \item \textbf{Static prompts.} One prompt is shared by all inputs, either a checklist of all twelve failure modes or a task prompt optimized with GEPA \citep{agrawal2026gepa} on our training data. These test whether a better prompt for the whole task is enough.
  \item \textbf{Visual-grounding controls.} A fixed instruction to ground the answer in visual evidence, the training-free visual prompt VISER \citep{izadi2025viser}, and a hint that $\mathcal{G}_\theta$ writes from a different image than the one the target sees test whether the gains come from the visual evidence in each hint or from a general push to look more carefully.
  \item \textbf{Answer heuristics.} Prophet \citep{yu2025prophet} replaces the hint with the answer and confidence of a Qwen3-VL-4B answerer LoRA-tuned on the same training split and budget as $\mathcal{G}_\theta$, and RICES \citep{yang2022pica,alayrac2022flamingo} prepends the 8 nearest training questions as demonstrations. These test whether a small model's answer or retrieved examples can take the place of the hint.
  \item \textbf{Generator training.} The off-the-shelf Qwen3-VL-4B-Instruct, prompted to write a contrastive hint without any training, and the Stage~1 generator without GRPO isolate what each training stage adds. LM-Guided CoT \citep{lee2024lmguidedcot} trains the same model with the same recipe and budget to write a rationale from the ground-truth rationales of A-OKVQA, which replaces the hint, and tests whether what the model is trained to write matters.
\end{itemize}

\noindent
{\bf A static prompt is not enough.} The failure-mode checklist does not improve accuracy reliably, and GEPA is significantly less accurate than \ours{} on all three targets. Useful anticipation must be fine-grained, pointing to the question or image evidence where the target is likely to err, which a prompt shared by all inputs cannot supply.

\noindent
{\bf The hints depend on the visual evidence.} A general push to look more carefully does not reproduce the gains. The grounding instruction and VISER change accuracy by between $-3.3$ and $+2.6$ points, and \ours{} is significantly more accurate than both on every target. The content of a hint is also tied to its own image. When $\mathcal{G}_\theta$ writes the hint from a different image, the gain disappears and accuracy falls to or below Base on all three targets. What the generator learns is therefore to surface the visual evidence that matters for the specific image and question.

\noindent
{\bf Answer heuristics cannot replace the hint.} Asked directly, $\mathcal{G}_\theta$ answers 77.55\% of the A-OKVQA test questions correctly, yet the targets it hints reach 84.28 to 90.57\%. Prophet and RICES, which draw on the same training data, are both significantly worse than \ours{} on all three targets, and \ours{} is also significantly more accurate than a confidence-based cascade of the LoRA-tuned answerer and the target in five of six target and benchmark pairs (\Cref{sec:supp:cascade}). The gains of \ours{} are therefore not explained simply by a 4B model having been trained on this dataset, and are more plausibly due to the hints themselves.

\noindent
{\bf Answer-leakage audit.} Because the Step~2 proposer sees the ground-truth answer, we audit whether test-time hints leak it, at three levels (\Cref{sec:supp:leakage}). \emph{(i) The hint text does not reveal the answer.} A manual inspection of 100 random A-OKVQA test hints (with their questions and options) found none that revealed or entailed the answer, and text-only probes recover it no better from the hint and the options than from the options alone. \emph{(ii) The hint does not work without the image.} A target that reads the question, the options, and the hint but not the image stays 24 to 44 points below \ours{}. \emph{(iii) The hint does not hand over the generator's own answer.} On the questions that $\mathcal{G}_\theta$ itself answers incorrectly, the hinted target chooses that wrong answer no more often than without the hint and is generally correct more often than Base. \ours{} therefore works differently from Prophet, and a wrong answer of the generator is not passed on to the target.

\noindent
{\bf Training on verified repairs matters.} The untrained hint writer gains at most 1.3 points of accuracy, whereas \ours{} is more accurate on all three targets and roughly doubles its repair rate at comparable harm, suggesting that the gains come from the training pipeline rather than from merely adding a hint. LM-Guided CoT is significantly less accurate than \ours{} on all three targets, although a rationale is free to state the answer. A hint trained on verified repairs helps more than a rationale from the same model and data. SFT accounts for most of the gain over Base, and GRPO then lowers the harm rate on every target, significantly for Claude 4.5 Haiku, and gives the highest accuracy on all three. This could be because imitating Step~2's repair-successful hints (i.e., the bulk of the SFT signal) teaches the generator \emph{what to do when there is something to fix}, whereas \emph{when not to interfere} with an already-correct answer is hard to learn from an SFT pool that only contains hints that happened to be safe. Stage~2's reward provides this discrimination directly.

\begin{table}[t]
\caption{{\small \textbf{Ablations and controls on the A-OKVQA test set}, grouped by the explanation they test. \textbf{Bold}: best per metric per target; \uline{underline}: second best. The second Base sample is the resampling noise floor and is not ranked.}}
\label{tab:ablation}
\centering
\small
\setlength{\tabcolsep}{4.0pt}
\renewcommand{\arraystretch}{1.10}
\resizebox{\textwidth}{!}{%
\begin{tabular}{l l ccc ccc ccc}
\toprule
& & \multicolumn{3}{c}{Claude 4.5 Haiku} & \multicolumn{3}{c}{Gemini 2.5 Flash Lite} & \multicolumn{3}{c}{GPT-5 Nano} \\
\cmidrule(lr){3-5}\cmidrule(lr){6-8}\cmidrule(lr){9-11}
Tests & Condition & Acc.\,($\uparrow$) & Repair\,($\uparrow$) & Harm\,($\downarrow$) & Acc.\,($\uparrow$) & Repair\,($\uparrow$) & Harm\,($\downarrow$) & Acc.\,($\uparrow$) & Repair\,($\uparrow$) & Harm\,($\downarrow$) \\
\midrule
\multirow{2}{*}{Reference} & Base & 78.25 & -- & -- & 82.88 & -- & -- & 84.37 & -- & -- \\
 & Base, second sample & 77.21 & 6.02 & 3.01 & 82.53 & 8.16 & 2.11 & 84.98 & 15.64 & 2.17 \\
\midrule
\multirow{2}{*}{\makecell[l]{Static\\prompts}} & Failure-mode checklist & 75.63 & 17.27 & 8.15 & 85.15 & 32.65 & 4.00 & 84.63 & 28.49 & 4.97 \\
 & GEPA \citep{agrawal2026gepa} & 75.37 & 14.06 & 7.59 & 84.19 & 33.16 & 5.27 & 85.68 & 29.05 & \uline{3.83} \\
\midrule
\multirow{3}{*}{\makecell[l]{Visual\\grounding}} & Grounding instruction & 76.24 & 18.88 & 7.81 & 85.50 & 32.14 & \uline{3.48} & 84.63 & 31.84 & 5.59 \\
 & VISER \citep{izadi2025viser} & 75.20 & 14.06 & 7.81 & 79.56 & 22.96 & 8.75 & 83.76 & 29.05 & 6.11 \\
 & Mismatched-image hint & 74.85 & 20.88 & 10.16 & 81.40 & 12.24 & 4.32 & 84.02 & 29.61 & 5.90 \\
\midrule
\multirow{2}{*}{\makecell[l]{Answer\\heuristics}} & Prophet \citep{yu2025prophet} & 75.02 & 24.90 & 11.05 & 86.11 & 41.84 & 4.74 & 85.07 & 27.37 & 4.24 \\
 & RICES \citep{yang2022pica,alayrac2022flamingo} ($k{=}8$) & 77.55 & 15.66 & \textbf{5.25} & 84.89 & 21.94 & \textbf{2.11} & 84.89 & 17.88 & \textbf{2.69} \\
\midrule
\multirow{4}{*}{\makecell[l]{Generator\\training}} & Untrained hint writer & 79.48 & 25.70 & \uline{5.58} & 84.19 & 29.08 & 4.43 & 84.19 & 26.82 & 5.18 \\
 & LM-Guided CoT \citep{lee2024lmguidedcot} & 76.33 & 19.28 & 7.81 & 85.33 & 32.14 & 3.69 & 85.76 & 39.66 & 5.69 \\
 & \ours{} (SFT only) & \uline{82.88} & \textbf{50.20} & 8.04 & \uline{89.78} & \uline{60.20} & 4.11 & \uline{87.07} & \uline{40.78} & 4.35 \\
 & \ours{} (SFT $+$ GRPO) & \textbf{84.28} & \uline{49.40} & 6.03 & \textbf{90.57} & \textbf{61.73} & \uline{3.48} & \textbf{88.12} & \textbf{45.81} & 4.04 \\
\bottomrule
\end{tabular}%
}
\end{table}

\section{Discussion}
\label{sec:discussion}

\noindent
{\bf Deployment use case and cost.}
\ours{} is designed for settings where a downstream consumer of frontier VLMs observes systematic in-domain failure patterns but cannot modify the model, retrain it, or afford the latency and cost of multi-pass correction at scale. It is most useful where frontier models are not yet saturated, such as in domain-specific applications. The total inference overhead is one small hint generator call plus a single target call, compared with two target calls for self-refine and three (two target plus one judge) for the external-judge baseline. The one-time training cost is about \$260 in API calls and 11 GPU hours (\Cref{sec:supp:cost}), and the same generator is then used on unseen benchmarks and target models without retraining (\Cref{sec:exp-main,sec:exp-transfer}). An accuracy point gained with \ours{} costs over 10 times less than one gained by switching to a stronger model of the same family (\Cref{sec:exp-cost}).

\noindent
{\bf \ours{} in compound AI systems.}
In the increasingly common compound AI systems and agentic workflows \citep{zaharia2024compound,khattab2023dspy}, self-refinement \citep{madaan2023selfrefine,shinn2023reflexion} and external judges or critics \citep{zheng2023llmasjudge,xie2025ctrl} are routinely used to check and improve the outputs of large models, at the cost of additional calls to large models or longer reasoning. \ours{} instead acts on a likely error before the target responds. Because the hint generator is much smaller and faster than the target and no revision rounds are needed, \ours{} avoids the sequential target and judge calls of post-response critique. Its plain-text hints can also precede any black-box call in a workflow, alone or alongside reflection or judge modules.

\noindent
{\bf Future directions.}
Two natural extensions follow. The error predictors of \Cref{sec:analysis} suggest gating hint generation by predicted failure risk, so that the hint pipeline only fires on inputs the target is likely to get wrong without help, further lowering the average inference cost. A second direction is adapting the framework beyond multiple choice to open-ended generative VQA, which requires replacing exact-match correctness in the reward with a task-appropriate outcome signal and extending the no-leakage check to free-form answers. Both extensions preserve the central design of \ours{}, namely a small generator that produces input-conditioned hints in a single call, and we leave them to future work.

\section{Conclusion}
\label{sec:conclusion}

We introduced Test-Time Hinting, a preemptive single-call test-time scaling method for black-box VLMs. The method exploits an empirical regularity, namely that failure modes are recurring, cross-model, and predictable from the input alone, to train a compact generator that prepends failure-mode-aware guidance to the target's prompt. A two-stage recipe combining distillation from agentic search with reinforcement learning against the actual downstream behavior of each candidate hint on its target VLM yields gains on overall accuracy across our primary targets, and these gains transfer zero-shot to unseen targets and unseen benchmarks. Controls trace them to hints grounded in the visual evidence of each input and to training on verified repairs. We believe this perspective reaches beyond our method. Test-time scaling methods can capitalize on the fact that certain errors of VLMs are predictable from the input and can be acted upon, and this generalizes to many frontier models with a single small model when it is optimized accordingly.

\section{Acknowledgments}
This work was supported by NIH Award 1OT2OD038079-01.

\bibliography{main_v3}

\begin{thebibliography}{51}
\providecommand{\natexlab}[1]{#1}
\providecommand{\url}[1]{\texttt{#1}}
\expandafter\ifx\csname urlstyle\endcsname\relax
  \providecommand{\doi}[1]{doi: #1}\else
  \providecommand{\doi}{doi: \begingroup \urlstyle{rm}\Url}\fi

\bibitem[Agrawal et~al.(2026)Agrawal, Tan, Soylu, Ziems, Khare, Opsahl-Ong,
  Singhvi, Shandilya, Ryan, Jiang, Potts, Sen, Dimakis, Stoica, Klein, Zaharia,
  and Khattab]{agrawal2026gepa}
Lakshya~A. Agrawal, Shangyin Tan, Dilara Soylu, Noah Ziems, Rishi Khare, Krista
  Opsahl-Ong, Arnav Singhvi, Herumb Shandilya, Michael~J. Ryan, Meng Jiang,
  Christopher Potts, Koushik Sen, Alexandros~G. Dimakis, Ion Stoica, Dan Klein,
  Matei Zaharia, and Omar Khattab.
\newblock {GEPA}: Reflective prompt evolution can outperform reinforcement
  learning.
\newblock In \emph{International Conference on Learning Representations}, 2026.

\bibitem[Alayrac et~al.(2022)Alayrac, Donahue, Luc, Miech, Barr, Hasson, Lenc,
  Mensch, Millican, Reynolds, et~al.]{alayrac2022flamingo}
Jean-Baptiste Alayrac, Jeff Donahue, Pauline Luc, Antoine Miech, Iain Barr,
  Yana Hasson, Karel Lenc, Arthur Mensch, Katherine Millican, Malcolm Reynolds,
  et~al.
\newblock Flamingo: a visual language model for few-shot learning.
\newblock In \emph{Advances in Neural Information Processing Systems},
  volume~35, 2022.

\bibitem[Avogaro et~al.(2026)Avogaro, Debnath, Mi, Frick, Wang, He, Hua,
  Schindler, and Rigotti]{avogaro2026sparc}
Niccolo Avogaro, Nayanika Debnath, Li~Mi, Thomas Frick, Junling Wang, Zexue He,
  Hang Hua, Konrad Schindler, and Mattia Rigotti.
\newblock {SPARC}: Separating perception and reasoning circuits for test-time
  scaling of {VLMs}.
\newblock In \emph{International Conference on Machine Learning}, 2026.

\bibitem[Bai et~al.(2025)Bai, Cai, Chen, Chen, Chen, Cheng, Deng, Ding, Gao,
  Ge, et~al.]{bai2025qwen3vl}
Shuai Bai, Yuxuan Cai, Ruizhe Chen, Keqin Chen, Xionghui Chen, Zesen Cheng,
  Lianghao Deng, Wei Ding, Chang Gao, Chunjiang Ge, et~al.
\newblock {Qwen3-VL} technical report.
\newblock \emph{arXiv preprint arXiv:2511.21631}, 2025.

\bibitem[Chen et~al.(2024)Chen, Zaharia, and Zou]{chen2023frugalgpt}
Lingjiao Chen, Matei Zaharia, and James Zou.
\newblock {FrugalGPT}: How to use large language models while reducing cost and
  improving performance.
\newblock \emph{Transactions on Machine Learning Research}, 2024.

\bibitem[Cheng et~al.(2024)Cheng, Liu, Zheng, Ke, Wang, Dong, Tang, and
  Huang]{cheng2024black}
Jiale Cheng, Xiao Liu, Kehan Zheng, Pei Ke, Hongning Wang, Yuxiao Dong, Jie
  Tang, and Minlie Huang.
\newblock Black-box prompt optimization: Aligning large language models without
  model training.
\newblock In \emph{Proceedings of the 62nd Annual Meeting of the Association
  for Computational Linguistics (Volume 1: Long Papers)}, pages 3201--3219,
  2024.

\bibitem[Du et~al.(2024)Du, Li, Torralba, Tenenbaum, and
  Mordatch]{du2023debate}
Yilun Du, Shuang Li, Antonio Torralba, Joshua~B. Tenenbaum, and Igor Mordatch.
\newblock Improving factuality and reasoning in language models through
  multiagent debate.
\newblock In \emph{Proceedings of the 41st International Conference on Machine
  Learning}, volume 235 of \emph{Proceedings of Machine Learning Research},
  pages 11733--11763. PMLR, 2024.

\bibitem[Fang et~al.(2026)Fang, You, Shi, Wang, Wang, Zhang, Chang, and
  Gu]{aozpt2026}
Jinjie Fang, Runwen You, Wanli Shi, Wenkang Wang, Ganyu Wang, Haozhen Zhang,
  Yi~Chang, and Bin Gu.
\newblock Online black-box prompt optimization with regret guarantees under
  noisy feedback.
\newblock In \emph{International Conference on Learning Representations}, 2026.

\bibitem[Guo et~al.(2025)Guo, Yang, Zhang, Song, Wang, Zhu, Xu, Zhang, Ma, Bi,
  et~al.]{guo2025deepseek}
Daya Guo, Dejian Yang, Haowei Zhang, Junxiao Song, Peiyi Wang, Qihao Zhu,
  Runxin Xu, Ruoyu Zhang, Shirong Ma, Xiao Bi, et~al.
\newblock {DeepSeek-R1} incentivizes reasoning in {LLMs} through reinforcement
  learning.
\newblock \emph{Nature}, 645\penalty0 (8081):\penalty0 633--638, 2025.

\bibitem[Hu et~al.(2022)Hu, Shen, Wallis, Allen-Zhu, Li, Wang, Wang, and
  Chen]{hu2022lora}
Edward~J. Hu, Yelong Shen, Phillip Wallis, Zeyuan Allen-Zhu, Yuanzhi Li, Shean
  Wang, Lu~Wang, and Weizhu Chen.
\newblock {LoRA}: Low-rank adaptation of large language models.
\newblock In \emph{International Conference on Learning Representations}, 2022.

\bibitem[Izadi et~al.(2025)Izadi, Banayeeanzade, Askari, Rahimiakbar, Vahedi,
  Hasani, and Soleymani~Baghshah]{izadi2025viser}
Amirmohammad Izadi, Mohammad~Ali Banayeeanzade, Fatemeh Askari, Ali
  Rahimiakbar, Mohammad~Mahdi Vahedi, Hosein Hasani, and Mahdieh
  Soleymani~Baghshah.
\newblock Visual structures help visual reasoning: Addressing the binding
  problem in {LVLMs}.
\newblock In \emph{Advances in Neural Information Processing Systems},
  volume~38, 2025.

\bibitem[Jeddi et~al.(2026)Jeddi, Le, Kazerouni, Karaimer, Nguyen, Mohomed,
  Brudno, Levinshtein, Derpanis, Taati, and Grzeszczuk]{jeddi2026avis}
Ahmadreza Jeddi, Minh~Ngoc Le, Amirhossein Kazerouni, Hakki~Can Karaimer, Hue
  Nguyen, Iqbal Mohomed, Michael Brudno, Alex Levinshtein, Konstantinos~G.
  Derpanis, Babak Taati, and Radek Grzeszczuk.
\newblock {AVIS}: Adaptive test-time scaling for vision-language models.
\newblock \emph{arXiv preprint arXiv:2606.11576}, 2026.

\bibitem[Kaya et~al.(2026)Kaya, Elliott, and Papadopoulos]{kaya2026efficient}
Mehmet~Onurcan Kaya, Desmond Elliott, and Dim~P. Papadopoulos.
\newblock Efficient test-time scaling for small vision-language models.
\newblock In \emph{International Conference on Learning Representations}, 2026.

\bibitem[Khattab et~al.(2024)Khattab, Singhvi, Maheshwari, Zhang, Santhanam,
  Vardhamanan, Haq, Sharma, Joshi, Moazam, Miller, Zaharia, and
  Potts]{khattab2023dspy}
Omar Khattab, Arnav Singhvi, Paridhi Maheshwari, Zhiyuan Zhang, Keshav
  Santhanam, Sri Vardhamanan, Saiful Haq, Ashutosh Sharma, Thomas~T. Joshi,
  Hanna Moazam, Heather Miller, Matei Zaharia, and Christopher Potts.
\newblock {DSPy}: Compiling declarative language model calls into
  self-improving pipelines.
\newblock In \emph{International Conference on Learning Representations}, pages
  54928--54958, 2024.

\bibitem[Kojima et~al.(2022)Kojima, Gu, Reid, Matsuo, and
  Iwasawa]{kojima2022zeroshot}
Takeshi Kojima, Shixiang~Shane Gu, Machel Reid, Yutaka Matsuo, and Yusuke
  Iwasawa.
\newblock Large language models are zero-shot reasoners.
\newblock In \emph{Advances in Neural Information Processing Systems},
  volume~35, pages 22199--22213, 2022.

\bibitem[Lee et~al.(2024)Lee, Yang, Tran, Hu, Barut, and
  Chang]{lee2024lmguidedcot}
Jooyoung Lee, Fan Yang, Thanh Tran, Qian Hu, Emre Barut, and Kai-Wei Chang.
\newblock Can small language models help large language models reason better?:
  {LM}-guided chain-of-thought.
\newblock In \emph{Proceedings of the 2024 Joint International Conference on
  Computational Linguistics, Language Resources and Evaluation (LREC-COLING
  2024)}, pages 2835--2843, 2024.

\bibitem[Li et~al.(2023)Li, Peng, He, Galley, Gao, and Yan]{li2023dsp}
Zekun Li, Baolin Peng, Pengcheng He, Michel Galley, Jianfeng Gao, and Xifeng
  Yan.
\newblock Guiding large language models via directional stimulus prompting.
\newblock In \emph{Advances in Neural Information Processing Systems},
  volume~36, 2023.

\bibitem[Liang et~al.(2024)Liang, He, Jiao, Wang, Wang, Wang, Yang, Shi, and
  Tu]{liang2023debate}
Tian Liang, Zhiwei He, Wenxiang Jiao, Xing Wang, Yan Wang, Rui Wang, Yujiu
  Yang, Shuming Shi, and Zhaopeng Tu.
\newblock Encouraging divergent thinking in large language models through
  multi-agent debate.
\newblock In \emph{Proceedings of the 2024 Conference on Empirical Methods in
  Natural Language Processing}, pages 17889--17904. Association for
  Computational Linguistics, 2024.

\bibitem[Lightman et~al.(2024)Lightman, Kosaraju, Burda, Edwards, Baker, Lee,
  Leike, Schulman, Sutskever, and Cobbe]{lightman2023letverify}
Hunter Lightman, Vineet Kosaraju, Yura Burda, Harri Edwards, Bowen Baker, Teddy
  Lee, Jan Leike, John Schulman, Ilya Sutskever, and Karl Cobbe.
\newblock Let's verify step by step.
\newblock In \emph{International Conference on Learning Representations}, 2024.

\bibitem[Lu et~al.(2022)Lu, Mishra, Xia, Qiu, Chang, Zhu, Tafjord, Clark, and
  Kalyan]{lu2022scienceqa}
Pan Lu, Swaroop Mishra, Tony Xia, Liang Qiu, Kai-Wei Chang, Song-Chun Zhu,
  Oyvind Tafjord, Peter Clark, and Ashwin Kalyan.
\newblock Learn to explain: Multimodal reasoning via thought chains for science
  question answering.
\newblock In \emph{Advances in Neural Information Processing Systems},
  volume~35, pages 2507--2521, 2022.

\bibitem[Madaan et~al.(2023)Madaan, Tandon, Gupta, Hallinan, Gao, Wiegreffe,
  Alon, Dziri, Prabhumoye, Yang, Gupta, Majumder, Hermann, Welleck,
  Yazdanbakhsh, and Clark]{madaan2023selfrefine}
Aman Madaan, Niket Tandon, Prakhar Gupta, Skyler Hallinan, Luyu Gao, Sarah
  Wiegreffe, Uri Alon, Nouha Dziri, Shrimai Prabhumoye, Yiming Yang, Shashank
  Gupta, Bodhisattwa~Prasad Majumder, Katherine Hermann, Sean Welleck, Amir
  Yazdanbakhsh, and Peter Clark.
\newblock {Self-Refine}: Iterative refinement with self-feedback.
\newblock In \emph{Advances in Neural Information Processing Systems},
  volume~36, pages 46534--46594, 2023.

\bibitem[Meng et~al.(2025)Meng, Cui, Dai, and Gong]{b2tpt2025}
Fan'an Meng, Chaoran Cui, Hongjun Dai, and Shuai Gong.
\newblock Black-box test-time prompt tuning for vision-language models.
\newblock In \emph{Proceedings of the AAAI Conference on Artificial
  Intelligence}, volume~39, pages 6099--6107, 2025.

\bibitem[Radford et~al.(2021)Radford, Kim, Hallacy, Ramesh, Goh, Agarwal,
  Sastry, Askell, Mishkin, Clark, Krueger, and Sutskever]{radford2021clip}
Alec Radford, Jong~Wook Kim, Chris Hallacy, Aditya Ramesh, Gabriel Goh,
  Sandhini Agarwal, Girish Sastry, Amanda Askell, Pamela Mishkin, Jack Clark,
  Gretchen Krueger, and Ilya Sutskever.
\newblock Learning transferable visual models from natural language
  supervision.
\newblock In Marina Meila and Tong Zhang, editors, \emph{Proceedings of the
  38th International Conference on Machine Learning}, volume 139 of
  \emph{Proceedings of Machine Learning Research}, pages 8748--8763. PMLR,
  18--24 Jul 2021.
\newblock URL \url{https://proceedings.mlr.press/v139/radford21a.html}.

\bibitem[Schwenk et~al.(2022)Schwenk, Khandelwal, Clark, Marino, and
  Mottaghi]{schwenk2022aokvqa}
Dustin Schwenk, Apoorv Khandelwal, Christopher Clark, Kenneth Marino, and
  Roozbeh Mottaghi.
\newblock {A-OKVQA}: A benchmark for visual question answering using world
  knowledge.
\newblock In \emph{European Conference on Computer Vision (ECCV)}, volume 13668
  of \emph{Lecture Notes in Computer Science}, pages 146--162. Springer, 2022.

\bibitem[Shao et~al.(2024{\natexlab{a}})Shao, Qian, Xiao, Song, Zong, Wang,
  Liu, and Li]{rose2023visualcot}
Hao Shao, Shengju Qian, Han Xiao, Guanglu Song, Zhuofan Zong, Letian Wang,
  Yu~Liu, and Hongsheng Li.
\newblock Visual {CoT}: Advancing multi-modal language models with a
  comprehensive dataset and benchmark for chain-of-thought reasoning.
\newblock In \emph{Advances in Neural Information Processing Systems (Datasets
  and Benchmarks Track)}, volume~37, pages 8612--8642, 2024{\natexlab{a}}.

\bibitem[Shao et~al.(2024{\natexlab{b}})Shao, Wang, Zhu, Xu, Song, Bi, Zhang,
  Zhang, Li, Wu, and Guo]{shao2024deepseekmath}
Zhihong Shao, Peiyi Wang, Qihao Zhu, Runxin Xu, Junxiao Song, Xiao Bi, Haowei
  Zhang, Mingchuan Zhang, Y.~K. Li, Y.~Wu, and Daya Guo.
\newblock {DeepSeekMath}: Pushing the limits of mathematical reasoning in open
  language models.
\newblock \emph{arXiv preprint arXiv:2402.03300}, 2024{\natexlab{b}}.

\bibitem[Shinn et~al.(2023)Shinn, Cassano, Gopinath, Narasimhan, and
  Yao]{shinn2023reflexion}
Noah Shinn, Federico Cassano, Ashwin Gopinath, Karthik Narasimhan, and Shunyu
  Yao.
\newblock Reflexion: Language agents with verbal reinforcement learning.
\newblock In \emph{Advances in Neural Information Processing Systems},
  volume~36, pages 8634--8652, 2023.

\bibitem[Shu et~al.(2022)Shu, Nie, Huang, Yu, Goldstein, Anandkumar, and
  Xiao]{shu2022tpt}
Manli Shu, Weili Nie, De-An Huang, Zhiding Yu, Tom Goldstein, Anima Anandkumar,
  and Chaowei Xiao.
\newblock Test-time prompt tuning for zero-shot generalization in
  vision-language models.
\newblock In \emph{Advances in Neural Information Processing Systems},
  volume~35, pages 14274--14289, 2022.

\bibitem[Snell et~al.(2025)Snell, Lee, Xu, and Kumar]{snell2024scalingttc}
Charlie Snell, Jaehoon Lee, Kelvin Xu, and Aviral Kumar.
\newblock Scaling {LLM} test-time compute optimally can be more effective than
  scaling parameters for reasoning.
\newblock In \emph{International Conference on Learning Representations}, 2025.

\bibitem[{Thinking Machines Lab}(2026)]{tml2026inkling}
{Thinking Machines Lab}.
\newblock Inkling: Our open-weights model.
\newblock \url{https://thinkingmachines.ai/news/introducing-inkling/}, 2026.

\bibitem[Wang et~al.(2026{\natexlab{a}})Wang, Fan, Zhang, Jing, Di, Song, Liu,
  and Cong]{vpae2026}
Junze Wang, Lei Fan, Dezheng Zhang, Weipeng Jing, Donglin Di, Yang Song, Sidong
  Liu, and Cong Cong.
\newblock Visual prompt-agnostic evolution.
\newblock In \emph{International Conference on Learning Representations},
  2026{\natexlab{a}}.

\bibitem[Wang et~al.(2026{\natexlab{b}})Wang, Gao, Chen, Chen, Zhu, Zhao, Liu,
  Cao, Ye, Zhu, Lu, Duan, Qiao, Dai, and Wang]{wang2025visualprm}
Weiyun Wang, Zhangwei Gao, Lianjie Chen, Zhe Chen, Jinguo Zhu, Xiangyu Zhao,
  Yangzhou Liu, Yue Cao, Shenglong Ye, Xizhou Zhu, Lewei Lu, Haodong Duan,
  Yu~Qiao, Jifeng Dai, and Wenhai Wang.
\newblock {VisualPRM400K}: An effective dataset for training multimodal process
  reward models.
\newblock In \emph{International Conference on Learning Representations},
  2026{\natexlab{b}}.

\bibitem[Wang et~al.(2023)Wang, Wei, Schuurmans, Le, Chi, Narang, Chowdhery,
  and Zhou]{wang2023selfconsistency}
Xuezhi Wang, Jason Wei, Dale Schuurmans, Quoc Le, Ed~Chi, Sharan Narang,
  Aakanksha Chowdhery, and Denny Zhou.
\newblock Self-consistency improves chain of thought reasoning in language
  models.
\newblock In \emph{International Conference on Learning Representations}, 2023.

\bibitem[Wei et~al.(2022)Wei, Wang, Schuurmans, Bosma, Ichter, Xia, Chi, Le,
  and Zhou]{wei2022cot}
Jason Wei, Xuezhi Wang, Dale Schuurmans, Maarten Bosma, Brian Ichter, Fei Xia,
  Ed~Chi, Quoc Le, and Denny Zhou.
\newblock Chain-of-thought prompting elicits reasoning in large language
  models.
\newblock In \emph{Advances in Neural Information Processing Systems},
  volume~35, pages 24824--24837, 2022.

\bibitem[{xAI}(2024{\natexlab{a}})]{xai2024realworldqa}
{xAI}.
\newblock {RealWorldQA} (dataset).
\newblock \url{https://huggingface.co/datasets/xai-org/RealworldQA},
  2024{\natexlab{a}}.
\newblock Official xAI release on Hugging Face. Accessed: 2026-03-05.

\bibitem[{xAI}(2024{\natexlab{b}})]{xai_realworldqa_2024}
{xAI}.
\newblock {Grok-1.5 Vision Preview}.
\newblock \url{https://x.ai/news/grok-1.5v}, April 2024{\natexlab{b}}.
\newblock Introduces the RealWorldQA benchmark and provides the official
  dataset download link. Accessed: 2026-03-05.

\bibitem[Xie et~al.(2025)Xie, Chen, Chen, Mao, Xu, and Kong]{xie2025ctrl}
Zhihui Xie, Jie Chen, Liyu Chen, Weichao Mao, Jingjing Xu, and Lingpeng Kong.
\newblock Teaching language models to critique via reinforcement learning.
\newblock In \emph{Proceedings of the 42nd International Conference on Machine
  Learning}, volume 267 of \emph{Proceedings of Machine Learning Research},
  pages 68559--68577. PMLR, 2025.

\bibitem[Xu et~al.(2025)Xu, Jin, Wu, Li, Song, Sun, and Yuan]{xu2024llavacot}
Guowei Xu, Peng Jin, Ziang Wu, Hao Li, Yibing Song, Lichao Sun, and Li~Yuan.
\newblock {LLaVA-CoT}: Let vision language models reason step-by-step.
\newblock In \emph{IEEE/CVF International Conference on Computer Vision
  (ICCV)}, pages 2087--2098, 2025.

\bibitem[Yang et~al.(2024)Yang, Wang, Lu, Liu, Le, Zhou, and
  Chen]{yang2024opro}
Chengrun Yang, Xuezhi Wang, Yifeng Lu, Hanxiao Liu, Quoc~V. Le, Denny Zhou, and
  Xinyun Chen.
\newblock Large language models as optimizers.
\newblock In \emph{International Conference on Learning Representations}, 2024.

\bibitem[Yang et~al.(2022)Yang, Gan, Wang, Hu, Lu, Liu, and Wang]{yang2022pica}
Zhengyuan Yang, Zhe Gan, Jianfeng Wang, Xiaowei Hu, Yumao Lu, Zicheng Liu, and
  Lijuan Wang.
\newblock An empirical study of {GPT-3} for few-shot knowledge-based {VQA}.
\newblock In \emph{Proceedings of the AAAI Conference on Artificial
  Intelligence}, volume~36, 2022.

\bibitem[Yao et~al.(2023)Yao, Yu, Zhao, Shafran, Griffiths, Cao, and
  Narasimhan]{yao2024tot}
Shunyu Yao, Dian Yu, Jeffrey Zhao, Izhak Shafran, Thomas~L. Griffiths, Yuan
  Cao, and Karthik Narasimhan.
\newblock Tree of thoughts: Deliberate problem solving with large language
  models.
\newblock In \emph{Advances in Neural Information Processing Systems},
  volume~36, pages 11809--11822, 2023.

\bibitem[Yu et~al.(2025)Yu, Ouyang, Shao, Wang, and Yu]{yu2025prophet}
Zhou Yu, Xuecheng Ouyang, Zhenwei Shao, Meng Wang, and Jun Yu.
\newblock Prophet: Prompting large language models with complementary answer
  heuristics for knowledge-based visual question answering.
\newblock \emph{IEEE Transactions on Pattern Analysis and Machine
  Intelligence}, 47\penalty0 (8):\penalty0 6797--6808, 2025.

\bibitem[Zaharia et~al.(2024)Zaharia, Khattab, Chen, Davis, Miller, Potts, Zou,
  Carbin, Frankle, Rao, and Ghodsi]{zaharia2024compound}
Matei Zaharia, Omar Khattab, Lingjiao Chen, Jared~Quincy Davis, Heather Miller,
  Chris Potts, James Zou, Michael Carbin, Jonathan Frankle, Naveen Rao, and Ali
  Ghodsi.
\newblock The shift from models to compound {AI} systems.
\newblock \url{https://bair.berkeley.edu/blog/2024/02/18/compound-ai-systems/},
  2024.
\newblock Berkeley Artificial Intelligence Research (BAIR) Blog.

\bibitem[Zellers et~al.(2019)Zellers, Bisk, Farhadi, and Choi]{zellers2019vcr}
Rowan Zellers, Yonatan Bisk, Ali Farhadi, and Yejin Choi.
\newblock From recognition to cognition: {V}isual commonsense reasoning.
\newblock In \emph{IEEE/CVF Conference on Computer Vision and Pattern
  Recognition (CVPR)}, pages 6720--6731, 2019.

\bibitem[Zhang et~al.(2025)Zhang, Lei, Li, Wang, Liu, Yang, Li, Wang, Yang, Wu,
  Ye, Ouyang, and Zhou]{luo2024criticv}
Di~Zhang, Jingdi Lei, Junxian Li, Xunzhi Wang, Yujie Liu, Zonglin Yang, Jiatong
  Li, Weida Wang, Suorong Yang, Jianbo Wu, Peng Ye, Wanli Ouyang, and Dongzhan
  Zhou.
\newblock {Critic-V}: {VLM} critics help catch {VLM} errors in multimodal
  reasoning.
\newblock In \emph{IEEE/CVF Conference on Computer Vision and Pattern
  Recognition (CVPR)}, pages 9050--9061, 2025.

\bibitem[Zhang et~al.(2022)Zhang, Wang, Zhou, Schuurmans, and
  Gonzalez]{zhang2022tempera}
Tianjun Zhang, Xuezhi Wang, Denny Zhou, Dale Schuurmans, and Joseph~E Gonzalez.
\newblock Tempera: Test-time prompting via reinforcement learning.
\newblock \emph{arXiv preprint arXiv:2211.11890}, 2022.

\bibitem[Zhang et~al.(2024{\natexlab{a}})Zhang, Madaan, Gao, Zheng, Mishra,
  Yang, Tandon, and Alon]{zhang2024leap}
Tianjun Zhang, Aman Madaan, Luyu Gao, Steven Zheng, Swaroop Mishra, Yiming
  Yang, Niket Tandon, and Uri Alon.
\newblock In-context principle learning from mistakes.
\newblock In \emph{Proceedings of the 41st International Conference on Machine
  Learning}, volume 235 of \emph{Proceedings of Machine Learning Research},
  pages 59520--59558. PMLR, 2024{\natexlab{a}}.

\bibitem[Zhang et~al.(2024{\natexlab{b}})Zhang, Zhang, Li, Zhao, Karypis, and
  Smola]{zhang2023multimodalcot}
Zhuosheng Zhang, Aston Zhang, Mu~Li, Hai Zhao, George Karypis, and Alex Smola.
\newblock Multimodal chain-of-thought reasoning in language models.
\newblock \emph{Transactions on Machine Learning Research}, 2024{\natexlab{b}}.

\bibitem[Zheng et~al.(2023)Zheng, Chiang, Sheng, Zhuang, Wu, Zhuang, Lin, Li,
  Li, Xing, Zhang, Gonzalez, and Stoica]{zheng2023llmasjudge}
Lianmin Zheng, Wei-Lin Chiang, Ying Sheng, Siyuan Zhuang, Zhanghao Wu, Yonghao
  Zhuang, Zi~Lin, Zhuohan Li, Dacheng Li, Eric~P. Xing, Hao Zhang, Joseph~E.
  Gonzalez, and Ion Stoica.
\newblock Judging {LLM}-as-a-judge with {MT-Bench} and {Chatbot Arena}.
\newblock In \emph{Advances in Neural Information Processing Systems (Datasets
  and Benchmarks Track)}, volume~36, pages 46595--46623, 2023.

\bibitem[Zhou et~al.(2023)Zhou, Muresanu, Han, Paster, Pitis, Chan, and
  Ba]{zhou2023ape}
Yongchao Zhou, Andrei~Ioan Muresanu, Ziwen Han, Keiran Paster, Silviu Pitis,
  Harris Chan, and Jimmy Ba.
\newblock Large language models are human-level prompt engineers.
\newblock In \emph{International Conference on Learning Representations}, 2023.

\bibitem[Zhu et~al.(2016)Zhu, Groth, Bernstein, and Fei-Fei]{zhu2016visual7w}
Yuke Zhu, Oliver Groth, Michael Bernstein, and Li~Fei-Fei.
\newblock {Visual7W}: Grounded question answering in images.
\newblock In \emph{IEEE Conference on Computer Vision and Pattern Recognition
  (CVPR)}, pages 4995--5004, 2016.

\end{thebibliography}
\bibliographystyle{plainnat}

\appendix
\newpage

\section{Extended Related Work}
\label{sec:supp:related}

This appendix expands \Cref{sec:related}. Methods that improve a frozen model at inference time differ in three respects that decide whether they can be used in our setting. The first is the access to the model that they need. The second is the number of times they call the model for one query. The third is where their guidance comes from, whether it is shared by a whole task or written for each input, and what it is learned from. \ours{} is built for the most restrictive answers to the first two, black-box access and a single call. On the third, its guidance is written for each input and starts from the errors of the target, which are predictable from the input and largely shared across models. We discuss prior work along these three questions.

\subsection{What Access to the Model Is Needed?}

Test-time scaling methods treat inference compute as a resource to optimize rather than a fixed cost. For LLMs, gains have been achieved through chain-of-thought prompting \citep{wei2022cot}, self-consistency \citep{wang2023selfconsistency}, process reward models \citep{lightman2023letverify}, and best-of-$N$ sampling with learned verifiers \citep{snell2024scalingttc}. Analogous efforts for VLMs include multi-stage chain-of-thought reasoning \citep{zhang2023multimodalcot,xu2024llavacot} and reward-model-guided search \citep{wang2025visualprm}. The strongest of these results depend on access to the model itself. Test-time scaling for open-weight and for closed-weight VLMs can be seen as two design pathways that allow different levels of intervention. SPARC \citep{avogaro2026sparc}, the test-time augmentation and adaptation methods for small VLMs, TTAug and TTAdapt \citep{kaya2026efficient}, and AVIS \citep{jeddi2026avis} were developed and evaluated on open-weight models. They sample the target several times, through repeated relevance detection before a cropped re-answer, averaged predictions over augmented inputs, or an adaptive self-consistency budget, and, as evaluated, they rely on open weights or internals such as token probabilities, gradients, or attention keys. Because open-weight models expose strictly more information, such methods often cannot be applied to closed-weight models, whereas a method that needs only black-box access applies to both.

Several prompt adaptation methods have the same requirement. Test-time prompt tuning adapts the prompts of CLIP-style VLMs to test inputs via gradients \citep{shu2022tpt}. Visual prompt-agnostic evolution \citep{vpae2026} inserts learnable visual prompt tokens into a frozen ViT and models their layerwise evolution with a shared Koopman operator and Lyapunov regularization, which requires the backbone architecture and internal token sequence that a closed API does not expose. TEMPERA \citep{zhang2022tempera} trains an RL policy that edits each query's instruction, in-context examples, and label words, using the target's hidden states and label probabilities. Closed-weight models expose neither, which is exactly the setting \ours{} is built for.

As practitioners increasingly rely on frontier models whose weights they cannot access, we designed \ours{} for the closed-weight pathway, with one pass of a compact generator and a single call to the target. Its hints can also be prepended to the inputs of open-weight methods. On an open-weight target, \ours{} is significantly more accurate than TTAug on all four benchmarks (\Cref{sec:supp:openweight}).

\subsection{How Many Times Is the Target Called?}

A second group of methods works through the model interface and pays for it in calls to the target. A prominent family of TTS methods improves accuracy through repeated inference, i.e., generating a response, critiquing it, and optionally revising. The critique signal varies across approaches, ranging from the target model itself \citep{madaan2023selfrefine,shinn2023reflexion} to multi-agent debate \citep{du2023debate,liang2023debate} or a separate LLM acting as a judge \citep{zheng2023llmasjudge}, and to dedicated critique models trained for the task \citep{luo2024criticv,xie2025ctrl}. Across all post-response methods, at least two inference passes through the target are required, with external-judge approaches adding a third call, incurring meaningful cost at frontier API pricing. Test-time hinting replaces these additional calls with a single pass through a compact hint generator, thereby achieving comparable or superior accuracy improvement with a single call to the target model. Self-refine and an external judge are baselines in all our experiments.

Black-box prompt adaptation at test time follows the same pattern. B$^2$TPT \citep{b2tpt2025} learns low-dimensional text and vision prompts for CLIP-style classification at test time: it forms pseudo-labels from target outputs and evaluates candidate prompts iteratively with derivative-free evolutionary optimization. AOZPT \citep{aozpt2026} learns prompts sequentially from noisy performance feedback, combining zeroth-order updates with uncertainty scaling to obtain sublinear regret. In contrast, \ours{} learns its generator offline, uses no target output, confidence score, or online correctness signal at test time, and emits an input-specific natural-language hint before the target responds, with one target call per example.

\subsection{Where Does the Guidance Come From?}

{\bf Task-level prompts (APE, OPRO, DSPy, GEPA, and LEAP).} Prompt optimizers and \ours{} both improve behavior through prompts, but they solve different problems under different deployment constraints. APE \citep{zhou2023ape} uses demonstrations to propose and score candidate instructions, then applies one selected task-level instruction to all test inputs. This helps when a shared instruction is the bottleneck. In natural-image VQA, however, the target already knows the task, while its likely error depends on the image and question, and a shared instruction cannot identify what a particular input is likely to make the target mishandle. Our ablation supports this point, since a checklist of all twelve failure modes does not improve accuracy reliably (\Cref{sec:exp-ablation}). OPRO \citep{yang2024opro} searches for a task-level instruction by giving an optimizer LLM a history of prompts and scores. The search can run offline, but its deployed result is again one shared prompt. Using it for instance-specific repair would require evaluating candidates for each test image, which needs a correctness signal that is unavailable before the answer, as well as repeated target calls. DSPy \citep{khattab2023dspy} compiles a user-defined LM program against examples and a metric. It could wrap a VQA API and optimize task-level instructions or demonstrations, but it does not itself identify the current image's likely failure. Matching \ours{} would require the designer to construct multimodal failure-diagnosis and conditional-repair modules. GEPA \citep{agrawal2026gepa} evolves instructions through LLM reflection on execution traces, and its deployed result is likewise one optimized prompt that is applied to every input of a task. A task prompt optimized with GEPA on our training data is significantly less accurate than \ours{} on all three primary targets (\Cref{sec:supp:static}). LEAP \citep{zhang2024leap} mines mistakes on few-shot examples to derive benchmark-level principles, but the resulting summaries are difficult to apply to individual questions when benchmark identity is unknown. Chain-of-thought prompting \citep{wei2022cot} with its visual extensions \citep{rose2023visualcot,lu2022scienceqa} and Tree of Thoughts \citep{yao2024tot} are task-generic and do not condition on the failure mode a given input is likely to trigger. \ours{} instead amortizes this optimization into an input-conditioned generator that learns the mapping directly from image--question inputs and target outcomes, and it deploys with one compact-generator pass and one black-box target call, without prompt history or per-query search.

{\bf Learned input-level guidance (DSP, LM-Guided CoT, BPO, and Prophet).} These methods are the closest prior work. Like \ours{}, DSP, LM-Guided CoT, and BPO each train a small model to steer a frozen black-box LLM in a single call. DSP \citep{li2023dsp} is the closest. A small policy model such as T5, trained with SFT and RL, generates instance-specific stimuli such as keywords that steer a text-only LLM toward a desired output in summarization, dialogue, and reasoning. LM-Guided CoT \citep{lee2024lmguidedcot} trains a small model by distillation and RL to write a rationale for each input, from which a frozen large model predicts the answer. BPO \citep{cheng2024black} trains a prompt rewriter on human preference data so that black-box LLMs give better-aligned responses. Prophet \citep{yu2025prophet} moves toward input-specific prompting by feeding candidate answers and confidence scores from a small VQA model to an LLM, but its heuristics are not anticipatory of the target's likely errors and provide candidate answers regardless of whether the target is likely to fail. RICES \citep{yang2022pica,alayrac2022flamingo} retrieves the training examples most similar to a test input and prepends them as demonstrations.

\ours{} differs in where its guidance comes from. It starts from the errors of the target VLM, which are predictable from the input and largely shared across models (\Cref{sec:analysis}). Its hints are verified with the target in the loop to repair such errors, and its reward is defined relative to the target's base answer and penalizes turning a correct answer into a wrong one, whereas DSP rewards the quality of the LLM's output. It works on images, so a hint must point to visual evidence. Its hints must also contrast plausible alternatives and must not reveal the answer, which leaves the reasoning and the answer to the target, whereas LM-Guided CoT supplies the reasoning itself. One generator is then used unchanged on models and benchmarks that it never saw. We compare \ours{} with LM-Guided CoT, Prophet, RICES, and GEPA, each built from the same training data, and \ours{} is significantly more accurate than each of them on all three primary targets (\Cref{sec:supp:controls}). Importantly, CoT, Prophet, and \ours{} are complementary: hints can be prepended to any prompting strategy without architectural changes.

\FloatBarrier
\section{Cross-Model Failure Structure: Full Results}
\label{sec:supp:overlap}

This appendix expands the motivational analysis summarized in \Cref{sec:analysis}. We provide the failure-mode taxonomy in full and a short rationale for using a single model-conditioned hint generator instead of one specialist generator per target.

\subsection{Failure-Mode Taxonomy}
\label{sec:supp:taxonomy}

The taxonomy $\mathcal{F}$ used throughout the paper consists of twelve failure modes, hand-picked by the authors during a manual review of base-incorrect responses across the three primary target models. The list is intended to cover the empirically frequent ways in which the targets fail on natural-image VQA, and is used both for the cross-model analysis below and for the failure-mode checklist of \Cref{sec:exp-ablation}.

\begin{itemize}[leftmargin=*,topsep=2pt,itemsep=2pt]
  \item \textbf{recognition.} The target misidentifies the primary object, scene, or activity depicted in the image (\eg{} confusing a cat for a dog, a kitchen for a dining room).
  \item \textbf{attribute binding.} The target identifies the right entities but binds attributes (color, material, state, role) to the wrong one.
  \item \textbf{counting.} The target's numeric answer for the count of an object, person, or instance is incorrect.
  \item \textbf{spatial relation.} The target misjudges the spatial layout of entities (left/right, behind/in front, above/below, near/far).
  \item \textbf{OCR.} The target misreads or fails to read text rendered inside the image (signs, captions, packaging).
  \item \textbf{chart/table.} The target misinterprets the data structure of a chart, table, or schematic.
  \item \textbf{math/quantitative.} The target reaches an incorrect quantitative conclusion via arithmetic or comparison error after correct perception.
  \item \textbf{knowledge.} The target's perception is correct but it lacks or misapplies the world knowledge required to answer.
  \item \textbf{logic/negation.} The target mishandles negation, conjunction, exclusion, or other propositional structure in the question.
  \item \textbf{hallucination.} The target asserts the presence of entities, attributes, or events that are not in the image.
  \item \textbf{instruction format.} The target's reasoning is correct but its final answer fails to comply with the requested output format.
  \item \textbf{other.} A residual category for failures that do not fit any of the above.
\end{itemize}

\subsection{Single Model-Conditioned Generator vs.\ Per-Model Specialists}
\label{sec:supp:single_generator}

The cross-model overlap and failure-mode agreement reported above motivate a specific architectural choice for the hint generator: rather than training one specialist generator per target VLM, we train a single generator that takes the target identifier $m$ as an input token. The motivation is twofold. First, a substantial fraction of failures are shared across targets, both at the question level (Jaccard overlap of $0.32$ to $0.37$) and at the failure-mode level (agreement of $71$ to $76\%$ on shared errors), so a single model can learn the common failure-anticipation skill from data pooled across targets. Second, a shared-backbone risk predictor (\Cref{tab:supp:risk}) performs comparably to per-target specialists, providing direct evidence that the predictive structure underlying failure anticipation is largely shared. A single model-conditioned generator also enables zero-shot transfer to unseen target identifiers at inference time, which is the property exercised in \Cref{sec:exp-transfer}.

\FloatBarrier
\section{Method Details}
\label{sec:supp:method}

This appendix gives the pseudocode of the agentic hint search of Step~2 and the training details of Step~3.

\subsection{Agentic Hint Optimization: Pseudocode}
\label{sec:supp:agentic}

\Cref{alg:agentic} formalizes the three-role agentic loop summarized in \Cref{sec:phase2}. In the first proposal round, the editor required revisions for 8.4\% of proposals. Hints that still violated its constraints after the loop (2.6\% of all proposed hints) were discarded before Step~3, so no rejected hint was used to train $\mathcal{G}_\theta$ (\Cref{sec:supp:leakage}).

\begin{algorithm}[ht]
\small
\caption{Agentic Hint Optimization}
\label{alg:agentic}
\begin{algorithmic}[1]
\Require Image $x$, question $q$, ground truth $(a^{*}, r^{*})$, target $\mathcal{M}$ with base response $(\hat{a}, \hat{r})$, proposer $\mathcal{P}$, editor $\mathcal{E}$, max rounds $R_{\max}{=}3$, type $\tau\!\in\!\{\text{\texttt{repair}},\text{\texttt{reinforcement}}\}$
\Ensure Selected hint $h^{*}$ and outcome flag, or \textsc{Discard}
\State $\textit{feedback}\gets\varnothing$;\;$\hat{a}_{\textit{prev}}\gets\varnothing$;\;$\hat{r}_{\textit{prev}}\gets\varnothing$;\;$h_{\textit{last}}\gets\varnothing$
\For{$t = 1, \dots, R_{\max}$}
    \State $h \gets \mathcal{P}(x, q, a^{*}, r^{*}, \hat{a}, \hat{r}, \textit{feedback}, \hat{a}_{\textit{prev}}, \hat{r}_{\textit{prev}})$ \Comment{Propose}
    \State $(\textit{verdict}, h', \textit{feedback}) \gets \mathcal{E}(x, q, a^{*}, r^{*}, \hat{a}, \hat{r}, h)$ \Comment{Edit / revise}
    \If{$\textit{verdict} = \text{\texttt{revise}}$}
        \State $h \gets h'$
    \EndIf
    \State $h_{\textit{last}} \gets h$
    \State $(\hat{a}_h, \hat{r}_h) \gets \mathcal{M}(x, q, h)$ \Comment{Verifier re-answers}
    \If{$\hat{a}_h = a^{*}$}
        \State $h^{*} \gets h$
        \State \Return $(h^{*}, \textsc{Success})$
    \EndIf
    \State $\hat{a}_{\textit{prev}} \gets \hat{a}_h$;\; $\hat{r}_{\textit{prev}} \gets \hat{r}_h$
\EndFor
\If{$\tau = \text{\texttt{repair}}$}
    \State \Return $(h_{\textit{last}}, \textsc{UnsuccessfulRepair})$
\Else
    \State \Return \textsc{Discard}
\EndIf
\end{algorithmic}
\end{algorithm}

\subsection{Hint Generator Training: Hyperparameters and Reward Design}
\label{sec:supp:training_hparams}

This section details both stages of the Step~3 hint-generator training summarized in \Cref{sec:phase3}. \Cref{sec:supp:lora_scope} specifies the LoRA target modules and optimization conventions shared by both stages. \Cref{sec:supp:sft_hparams} and \Cref{sec:supp:grpo_optim} give the stage-specific hyperparameters for supervised fine-tuning and GRPO respectively, and \Cref{sec:supp:reward_table} through \Cref{sec:supp:base_correct_share} detail the reward design and data composition for Stage~2. Training was done on 1 $\times$ H200 GPU, where Stage~1 took approximately 2 hours to finish and Stage~2 took approximately 9 hours.

\subsubsection{Shared LoRA Scope and Optimization Conventions}
\label{sec:supp:lora_scope}

Both stages adapt the same set of LoRA target modules (the language-model backbone, the vision-language merger, and the last two blocks of the vision encoder), totaling 23 LoRA target modules. The exact module patterns are:

\begingroup\raggedright
\begin{itemize}[leftmargin=*,topsep=2pt,itemsep=2pt]
  \item \textbf{Language-model backbone:} \texttt{q\_proj}, \texttt{k\_proj}, \texttt{v\_proj}, \texttt{o\_proj}, \texttt{gate\_proj}, \texttt{up\_proj}, \texttt{down\_proj}.
  \item \textbf{Vision-language merger:} \texttt{model.\allowbreak visual.\allowbreak merger.\allowbreak linear\_fc1}, \texttt{model.\allowbreak visual.\allowbreak merger.\allowbreak linear\_fc2}, and the corresponding \texttt{linear\_fc1}/\texttt{linear\_fc2} pairs in each of the three \texttt{model.\allowbreak visual.\allowbreak deepstack\_\allowbreak merger\_list} blocks (eight modules total).
  \item \textbf{Vision encoder (last two blocks):} \texttt{attn.qkv}, \texttt{attn.proj}, \texttt{mlp.fc1}, \texttt{mlp.fc2}.
\end{itemize}
\endgroup

The LoRA configuration is identical across both stages: rank $r{=}16$, scaling $\alpha{=}32$, dropout $0.05$. Both stages use the AdamW optimizer with bf16 mixed precision and gradient checkpointing enabled.

\subsubsection{Stage 1: Supervised Fine-Tuning Hyperparameters}
\label{sec:supp:sft_hparams}

Stage~1 minimizes next-token cross-entropy on the hint JSON over the Step~2 trials retained as described in \Cref{sec:phase3}. We optimize with AdamW at learning rate $1.6 \times 10^{-4}$, weight decay $0$, gradient clipping at max norm $1.0$, and warmup ratio $0.03$. The LoRA scope and mixed-precision setup follow \Cref{sec:supp:lora_scope}.

\subsubsection{Stage 2: GRPO Optimization}
\label{sec:supp:grpo_optim}

Stage~2 is initialized from the Stage~1 LoRA adapter and KL-anchored (with coefficient $\beta{=}0.04$) to that reference, retaining the LoRA scope and configuration described in \Cref{sec:supp:lora_scope}. We use GRPO \citep{shao2024deepseekmath} with group size $G{=}8$, sampling temperature $0.9$, PPO clip $\varepsilon{=}0.2$, and AdamW learning rate $1{\times}10^{-5}$, training for $300$ optimization steps.

\subsubsection{Reward Truth Table}
\label{sec:supp:reward_table}

\Cref{tab:reward_truth_table} gives the reward of a candidate hint as a function of the target model's correctness with vs.\ without the hint. For a training prompt $(x, q, m)$, each candidate is sent to the target VLM named by $m$, and its outcome on that target is the reward used to update the policy.

\begin{table}[ht]
  \caption{\textbf{Reward for Stage~2 reinforcement learning.}}
  \label{tab:reward_truth_table}
  \centering
  \small
  \begin{tabular}{l l c}
    \toprule
    Outcome name & Description & Reward \\
    \midrule
    Repair      & Base wrong, hinted correct        & $+1$ \\
    No-op       & Base correct, hinted correct      & $\phantom{-}0$ \\
    Harm        & Base correct, hinted wrong        & $-1$ \\
    Unrepaired  & Base wrong, hinted wrong          & $-0.5$ \\
    \bottomrule
  \end{tabular}
\end{table}

\subsubsection{Why the Unrepaired Outcome Is Downweighted}
\label{sec:supp:asymmetric_penalty}

The reward in \Cref{tab:reward_truth_table} is asymmetric: the harm outcome (base correct, hinted wrong) receives $-1$, while the unrepaired outcome (base wrong, hinted wrong) receives only $-0.5$. The natural-looking choice would have been to keep the two penalties symmetric at $-1$, since both correspond to a wrong hinted answer. We did not do this for the following reason.

A symmetric $-1$ penalty creates an over-commit pressure on the policy. On hard inputs where every candidate hint is likely to leave the target wrong, a symmetric scheme treats all candidates in a group as equally bad, and the policy receives a strong undifferentiated negative signal that, propagated through the model, is most easily reduced by increasing the assertiveness of the hint, since assertive hints are the ones that occasionally get a hard input correct. Empirically, this assertiveness-inducing style does not stay localized to hard inputs: the same prompts and decoding patterns spill over into nearby base-correct prompts and produce harm.

Downweighting the unrepaired outcome to $-0.5$ keeps the actual harm signal intact (the policy is still penalized strongly when it disrupts a correct answer) while permitting cautious or no-op-style hints when the input is genuinely ambiguous. We selected the $-0.5$ value via a small sweep on the internal validation set, and harm-rate behavior was consistent across nearby choices.

\subsubsection{Mixing Base-Correct Prompts into the Stage~2 Pool}
\label{sec:supp:base_correct_share}

The Step~1 outputs are heavily skewed, since each target answers most training questions correctly at base. For such a base-correct pair of a question and a target, a hint can only preserve the answer or harm it. Strictly excluding these would deprive the policy of exposure to the no-op condition, which is exactly what the harm metric measures at evaluation. We therefore retain a configurable share of base-correct prompts in the Stage~2 training pool. A small sweep over shares of $\{30, 40, 50, 70\}\%$ identified $50\%$ as the most balanced operating point: enough exposure to teach the policy to leave already-correct answers alone, without diluting the harm-bearing prompts to the point of weakening repair coverage. We use $50\%$ for all reported \ours{} results.

\FloatBarrier
\section{Evaluation Protocol and Results with Confidence Intervals}
\label{sec:supp:eval}

{\bf Benchmark selection.} Our selection of benchmarks follows from four considerations. \emph{(i) General rather than capability-specific.} We want to show that learned hints improve visual understanding in general, without tailoring \ours{} to one capability. General VQA questions mix recognition, attributes, counting, spatial relations, and everyday knowledge, so they test the approach broadly. \emph{(ii) Failures a compact generator can address.} We target questions that frontier VLMs can answer with everyday knowledge and short reasoning but sometimes get wrong because they overlook or misread something in the image. This is where a 4B generator can plausibly help a much larger target, and it separates the effect of anticipatory hints from gaps in knowledge or multi-step reasoning. \emph{(iii) Diversity.} The four benchmarks vary the image source, question style, and answer distribution, from everyday scenes in A-OKVQA to movie stills in VCR and mostly driving scenes in RealWorldQA. Together they span knowledge-based VQA, visual commonsense reasoning, grounded perception questions, and real-world spatial understanding. \emph{(iv) Scale and measurability.} All four are established multiple-choice benchmarks with large test sets, including 5{,}000 questions each for VCR and Visual7W. These sizes support paired significance tests, and exact correctness gives GRPO a clear reward and makes repair, harm, and answer leakage directly measurable.

{\bf Why A-OKVQA for training.} A-OKVQA meets two needs of our training pipeline. First, learning to repair errors requires many questions that the targets get wrong, yet the targets answer most questions correctly. Roughly 71\% of A-OKVQA training questions are answered correctly by all three primary targets. With 17{,}056 training questions, A-OKVQA still leaves enough errors to support SFT and RL, which would be much harder with a smaller dataset. Second, its questions come with ground-truth rationales, short explanations that narrate the visual evidence and knowledge behind each answer. The proposer of Step~2 relies on them to ground its hints in the right evidence, which makes the hints more reliable, and the failure-mode annotator of \Cref{sec:analysis} uses them to label why a target failed.

{\bf Bootstrap and significance.} We quantify uncertainty with a joint paired nonparametric bootstrap over test questions. Each of the 2{,}000 replicates resamples the test questions with replacement, and the identical resampled question indices are shared across all methods and target models. Within each replicate, overall accuracy, repair rate, and harm rate are recomputed, with repair and harm rates computed on the resampled questions that the target answers incorrectly and correctly at base, respectively. Brackets in \Cref{tab:supp:ci_main,tab:supp:ci_unseen,tab:supp:ci_transfer} give 95\% confidence intervals. The intervals for the controls and baselines of \Cref{tab:ablation} are given with their setups in \Cref{sec:supp:controls}. To compare two methods, we compute the difference in their accuracy within each replicate and deem the difference significant if the 95\% confidence interval of this paired difference excludes zero.

{\bf External judges.} The external judge is a more capable model from the same family as the target: GPT-5-Mini for GPT-5 Nano, Gemini 2.5 Flash for Gemini 2.5 Flash Lite, and Claude 4.5 Sonnet for Claude 4.5 Haiku. For the unseen targets, the judges are GPT-5.2-Pro for GPT-5.2, Gemini 3 Flash for Gemini 2.5 Flash, Claude 4.6 Opus for Claude 4.5 Opus, Grok 4.1 Fast Reasoning for Grok 4 Fast Reasoning, and Nova Premier for Nova Pro. For the open-weight Inkling-Small, the judge is Inkling.

{\bf Target identifiers.} For same-family transfer, the hints generated for each training target are reused for the larger model of the same family (\Cref{sec:exp-setup}). For Grok 4 Fast Reasoning, Nova Pro, and Inkling-Small, $m$ is each model's own name, which never appears in training.

{\bf Paired-test conclusions.} On A-OKVQA (\Cref{tab:supp:ci_main}), \ours{} is significantly more accurate than Base and every baseline for Claude 4.5 Haiku and Gemini 2.5 Flash Lite. For GPT-5 Nano, it is significantly more accurate than Base, CoT, and self-refine. On the unseen benchmarks (\Cref{tab:supp:ci_transfer}), \ours{} is significantly more accurate than Base for every target on every benchmark, and than every baseline for Claude 4.5 Haiku and Gemini 2.5 Flash Lite. For GPT-5 Nano, \ours{} and the external judge are statistically indistinguishable on VCR and Visual7W, while on RealWorldQA the external judge, which is the most costly baseline and consults the much more capable GPT-5-Mini, is significantly more accurate. \Cref{tab:supp:ci_unseen} reports the five unseen targets on A-OKVQA. There, the gain of \ours{} over Base is significant for Grok 4 Fast Reasoning, and no baseline is significantly more accurate than \ours{} on any of the five targets. These targets answer 84 to 92\% of the 1{,}145 questions correctly at Base, so few errors remain from which a significant net gain can be obtained (\Cref{sec:supp:headroom}). On RealWorldQA, where the same-family targets are less saturated, the same paired tests give significant gains for two of the three (\Cref{sec:supp:rwqa_unseen}).

\begin{table}[H]
\caption{\textbf{Results on A-OKVQA with 95\% confidence intervals} (\Cref{tab:main}). Markers as in \Cref{tab:main}.}
\label{tab:supp:ci_main}
\centering
\footnotesize
\setlength{\tabcolsep}{4pt}
\renewcommand{\arraystretch}{1.05}
\begin{tabular*}{\textwidth}{@{\extracolsep{\fill}} ll r@{\hspace{3pt}}l r@{\hspace{3pt}}l r@{\hspace{3pt}}l @{}}
\toprule
Target Model & Method
& \multicolumn{2}{c}{\makecell{Overall Accuracy\\(\%, $\uparrow$)}}
& \multicolumn{2}{c}{\makecell{Repair Rate\\(\%, $\uparrow$)}}
& \multicolumn{2}{c}{\makecell{Harm Rate\\(\%, $\downarrow$)}} \\
\midrule
\multirow{5}{*}{\makecell[l]{Claude 4.5\\Haiku}}
& Base & 78.25\phantom{$^{\dagger}$} & \cirange{75.89}{80.61} & \multicolumn{2}{c}{--} & \multicolumn{2}{c}{--} \\
& CoT & 80.26\phantom{$^{\dagger}$} & \cirange{77.90}{82.53} & 26.10 & \cirange{20.71}{31.53} & \textbf{4.69} & \cirange{3.35}{6.11} \\
& Self-refine & 77.90\phantom{$^{\dagger}$} & \cirange{75.45}{80.26} & 30.52 & \cirange{24.81}{36.36} & 8.93 & \cirange{7.13}{10.83} \\
& External judge & \uline{80.87}\phantom{$^{\dagger}$} & \cirange{78.60}{83.14} & \uline{44.58} & \cirange{38.41}{50.75} & 9.04 & \cirange{7.10}{11.11} \\
& \ours{} (Ours) & \textbf{84.28}$^{\dagger}$ & \cirange{82.09}{86.45} & \textbf{49.40} & \cirange{43.09}{55.56} & \uline{6.03} & \cirange{4.52}{7.63} \\
\midrule
\multirow{5}{*}{\makecell[l]{Gemini 2.5\\Flash Lite}}
& Base & 82.88\phantom{$^{\dagger}$} & \cirange{80.70}{85.06} & \multicolumn{2}{c}{--} & \multicolumn{2}{c}{--} \\
& CoT & 84.10\phantom{$^{\dagger}$} & \cirange{81.92}{86.20} & 21.43 & \cirange{15.84}{27.45} & \uline{2.95} & \cirange{1.92}{4.05} \\
& Self-refine & \uline{87.69}\phantom{$^{\dagger}$} & \cirange{85.77}{89.61} & 35.71 & \cirange{28.96}{42.52} & \textbf{1.58} & \cirange{0.83}{2.40} \\
& External judge & 86.81\phantom{$^{\dagger}$} & \cirange{84.80}{88.73} & \uline{43.37} & \cirange{36.37}{50.51} & 4.21 & \cirange{3.01}{5.55} \\
& \ours{} (Ours) & \textbf{90.57}$^{\dagger}$ & \cirange{88.73}{92.31} & \textbf{61.73} & \cirange{54.83}{68.56} & 3.48 & \cirange{2.29}{4.80} \\
\midrule
\multirow{5}{*}{GPT-5 Nano}
& Base & 84.37\phantom{$^{\dagger}$} & \cirange{82.27}{86.47} & \multicolumn{2}{c}{--} & \multicolumn{2}{c}{--} \\
& CoT & 82.79\phantom{$^{\dagger}$} & \cirange{80.52}{84.97} & 18.44 & \cirange{12.91}{24.20} & 5.28 & \cirange{3.92}{6.77} \\
& Self-refine & 85.24\phantom{$^{\dagger}$} & \cirange{83.14}{87.25} & 17.88 & \cirange{12.43}{23.70} & \textbf{2.28} & \cirange{1.36}{3.28} \\
& External judge & \textbf{89.43}\phantom{$^{\dagger}$} & \cirange{87.60}{91.18} & \uline{44.69} & \cirange{37.35}{52.02} & \textbf{2.28} & \cirange{1.42}{3.28} \\
& \ours{} (Ours) & \uline{88.12}\phantom{$^{\dagger}$} & \cirange{86.11}{90.04} & \textbf{45.81} & \cirange{38.46}{53.05} & \uline{4.04} & \cirange{2.73}{5.42} \\
\bottomrule
\end{tabular*}
\end{table}

\begin{table}[H]
\caption{\textbf{Zero-shot transfer to unseen target models on A-OKVQA, with 95\% confidence intervals} (\Cref{tab:unseen}). \textbf{Bold}: best per target and metric; \uline{underline}: second best. $^{\ddagger}$ marks accuracy that is significantly above Base.}
\label{tab:supp:ci_unseen}
\centering
\footnotesize
\setlength{\tabcolsep}{4pt}
\renewcommand{\arraystretch}{1.05}
\begin{tabular*}{\textwidth}{@{\extracolsep{\fill}} ll r@{\hspace{3pt}}l r@{\hspace{3pt}}l r@{\hspace{3pt}}l @{}}
\toprule
Target Model & Method
& \multicolumn{2}{c}{\makecell{Overall Accuracy\\(\%, $\uparrow$)}}
& \multicolumn{2}{c}{\makecell{Repair Rate\\(\%, $\uparrow$)}}
& \multicolumn{2}{c}{\makecell{Harm Rate\\(\%, $\downarrow$)}} \\
\midrule
\multicolumn{8}{@{}l}{\emph{Same family}} \\
\midrule
\multirow{5}{*}{\makecell[l]{Claude 4.5\\Opus}}
& Base & 91.00\phantom{$^{\ddagger}$} & \cirange{89.34}{92.66} & \multicolumn{2}{c}{--} & \multicolumn{2}{c}{--} \\
& CoT & 90.92\phantom{$^{\ddagger}$} & \cirange{89.17}{92.58} & \uline{13.59} & \cirange{7.34}{20.56} & 1.44 & \cirange{0.76}{2.21} \\
& Self-refine & 91.09\phantom{$^{\ddagger}$} & \cirange{89.43}{92.66} & 5.83 & \cirange{1.86}{10.82} & \textbf{0.48} & \cirange{0.10}{0.95} \\
& External judge & \uline{91.27}\phantom{$^{\ddagger}$} & \cirange{89.61}{92.84} & 12.62 & \cirange{6.59}{19.54} & \uline{0.96} & \cirange{0.38}{1.60} \\
& \ours{} (Ours) & \textbf{91.62}\phantom{$^{\ddagger}$} & \cirange{90.05}{93.19} & \textbf{37.86} & \cirange{28.09}{47.37} & 3.07 & \cirange{1.98}{4.15} \\
\midrule
\multirow{5}{*}{\makecell[l]{Gemini 2.5\\Flash}}
& Base & \uline{91.88}\phantom{$^{\ddagger}$} & \cirange{90.22}{93.45} & \multicolumn{2}{c}{--} & \multicolumn{2}{c}{--} \\
& CoT & 91.53\phantom{$^{\ddagger}$} & \cirange{89.87}{93.10} & \uline{12.90} & \cirange{6.45}{20.00} & 1.52 & \cirange{0.85}{2.28} \\
& Self-refine & \uline{91.88}\phantom{$^{\ddagger}$} & \cirange{90.22}{93.45} & 7.53 & \cirange{2.50}{13.34} & \textbf{0.67} & \cirange{0.20}{1.22} \\
& External judge & 91.44\phantom{$^{\ddagger}$} & \cirange{89.78}{93.01} & 6.45 & \cirange{2.04}{11.93} & \uline{1.05} & \cirange{0.48}{1.71} \\
& \ours{} (Ours) & \textbf{92.05}\phantom{$^{\ddagger}$} & \cirange{90.48}{93.62} & \textbf{27.96} & \cirange{19.00}{37.36} & 2.28 & \cirange{1.42}{3.23} \\
\midrule
\multirow{5}{*}{GPT-5.2}
& Base & 90.22\phantom{$^{\ddagger}$} & \cirange{88.47}{91.88} & \multicolumn{2}{c}{--} & \multicolumn{2}{c}{--} \\
& CoT & 89.17\phantom{$^{\ddagger}$} & \cirange{87.34}{90.92} & 10.71 & \cirange{5.26}{16.82} & 2.32 & \cirange{1.44}{3.29} \\
& Self-refine & \uline{90.83}\phantom{$^{\ddagger}$} & \cirange{89.08}{92.49} & \uline{20.54} & \cirange{13.34}{28.34} & \uline{1.55} & \cirange{0.87}{2.34} \\
& External judge & 90.48\phantom{$^{\ddagger}$} & \cirange{88.73}{92.14} & 2.68 & \cirange{0.00}{6.09} & \textbf{0.00} & \cirange{0.00}{0.00} \\
& \ours{} (Ours) & \textbf{91.09}\phantom{$^{\ddagger}$} & \cirange{89.43}{92.66} & \textbf{30.36} & \cirange{21.70}{39.58} & 2.32 & \cirange{1.47}{3.29} \\
\midrule
\multicolumn{8}{@{}l}{\emph{Different family}} \\
\midrule
\multirow{5}{*}{\makecell[l]{Grok 4 Fast\\Reasoning}}
& Base & 85.33\phantom{$^{\ddagger}$} & \cirange{83.41}{87.25} & \multicolumn{2}{c}{--} & \multicolumn{2}{c}{--} \\
& CoT & 84.89\phantom{$^{\ddagger}$} & \cirange{82.97}{86.72} & 7.14 & \cirange{3.50}{11.29} & 1.74 & \cirange{0.95}{2.60} \\
& Self-refine & \uline{86.90}$^{\ddagger}$ & \cirange{84.98}{88.73} & 11.31 & \cirange{6.74}{16.34} & \textbf{0.10} & \cirange{0.00}{0.31} \\
& External judge & 86.38\phantom{$^{\ddagger}$} & \cirange{84.46}{88.22} & \uline{14.29} & \cirange{9.20}{19.77} & \uline{1.23} & \cirange{0.61}{1.96} \\
& \ours{} (Ours) & \textbf{87.60}$^{\ddagger}$ & \cirange{85.68}{89.43} & \textbf{43.45} & \cirange{35.93}{51.13} & 4.81 & \cirange{3.57}{6.15} \\
\midrule
\multirow{5}{*}{Nova Pro}
& Base & 83.76\phantom{$^{\ddagger}$} & \cirange{81.58}{85.86} & \multicolumn{2}{c}{--} & \multicolumn{2}{c}{--} \\
& CoT & 83.58\phantom{$^{\ddagger}$} & \cirange{81.40}{85.68} & 13.44 & \cirange{8.76}{18.50} & \textbf{2.82} & \cirange{1.84}{3.90} \\
& Self-refine & \uline{84.37}\phantom{$^{\ddagger}$} & \cirange{82.23}{86.34} & 22.04 & \cirange{16.18}{28.12} & \uline{3.55} & \cirange{2.42}{4.76} \\
& External judge & 83.32\phantom{$^{\ddagger}$} & \cirange{81.14}{85.50} & \uline{30.11} & \cirange{23.60}{36.81} & 6.36 & \cirange{4.86}{7.95} \\
& \ours{} (Ours) & \textbf{84.98}\phantom{$^{\ddagger}$} & \cirange{82.88}{87.08} & \textbf{43.55} & \cirange{37.00}{50.31} & 6.99 & \cirange{5.42}{8.64} \\
\bottomrule
\end{tabular*}
\end{table}

\begin{table}[H]
\caption{\textbf{Zero-shot transfer across benchmark datasets, with 95\% confidence intervals} (\Cref{tab:main}). Markers as in \Cref{tab:main}.}
\label{tab:supp:ci_transfer}
\centering
\footnotesize
\setlength{\tabcolsep}{4pt}
\renewcommand{\arraystretch}{1.0}
\setlength{\aboverulesep}{0.2ex}\setlength{\belowrulesep}{0.3ex}
\begin{tabular*}{\textwidth}{@{\extracolsep{\fill}} ll r@{\hspace{3pt}}l r@{\hspace{3pt}}l r@{\hspace{3pt}}l @{}}
\toprule
Target Model & Method
& \multicolumn{2}{c}{\makecell{Overall Accuracy\\(\%, $\uparrow$)}}
& \multicolumn{2}{c}{\makecell{Repair Rate\\(\%, $\uparrow$)}}
& \multicolumn{2}{c}{\makecell{Harm Rate\\(\%, $\downarrow$)}} \\
\midrule
\multicolumn{8}{@{}l}{\emph{VCR ($n=5{,}000$)}} \\
\midrule
\multirow{5}{*}{\makecell[l]{Claude 4.5\\Haiku}}
& Base & 61.64\phantom{$^{\dagger}$} & \cirange{60.31}{62.97} & \multicolumn{2}{c}{--} & \multicolumn{2}{c}{--} \\
& CoT & 62.06\phantom{$^{\dagger}$} & \cirange{60.73}{63.39} & 15.12 & \cirange{13.55}{16.74} & \uline{8.73} & \cirange{7.77}{9.77} \\
& Self-refine & 62.20\phantom{$^{\dagger}$} & \cirange{60.81}{63.53} & 11.47 & \cirange{10.10}{12.94} & \textbf{6.23} & \cirange{5.42}{7.12} \\
& External judge & \uline{62.60}\phantom{$^{\dagger}$} & \cirange{61.27}{63.93} & \uline{21.38} & \cirange{19.56}{23.23} & 11.75 & \cirange{10.63}{12.90} \\
& \ours{} (Ours) & \textbf{64.98}$^{\dagger}$ & \cirange{63.67}{66.29} & \textbf{34.41} & \cirange{32.32}{36.63} & 16.00 & \cirange{14.71}{17.30} \\
\midrule
\multirow{5}{*}{\makecell[l]{Gemini 2.5\\Flash Lite}}
& Base & 61.68\phantom{$^{\dagger}$} & \cirange{60.29}{63.01} & \multicolumn{2}{c}{--} & \multicolumn{2}{c}{--} \\
& CoT & 63.02\phantom{$^{\dagger}$} & \cirange{61.69}{64.35} & 14.56 & \cirange{13.05}{16.20} & 6.87 & \cirange{6.05}{7.80} \\
& Self-refine & \uline{65.40}\phantom{$^{\dagger}$} & \cirange{64.07}{66.73} & 14.56 & \cirange{13.05}{16.22} & \textbf{3.02} & \cirange{2.43}{3.63} \\
& External judge & 63.66\phantom{$^{\dagger}$} & \cirange{62.33}{64.99} & \uline{19.26} & \cirange{17.58}{21.08} & 8.75 & \cirange{7.80}{9.80} \\
& \ours{} (Ours) & \textbf{67.04}$^{\dagger}$ & \cirange{65.78}{68.30} & \textbf{23.17} & \cirange{21.30}{25.08} & \uline{5.71} & \cirange{4.90}{6.54} \\
\midrule
\multirow{5}{*}{GPT-5 Nano}
& Base & 63.00\phantom{$^{\dagger}$} & \cirange{61.67}{64.33} & \multicolumn{2}{c}{--} & \multicolumn{2}{c}{--} \\
& CoT & 64.04\phantom{$^{\dagger}$} & \cirange{62.71}{65.37} & 18.81 & \cirange{17.09}{20.66} & 9.40 & \cirange{8.41}{10.45} \\
& Self-refine & 63.20\phantom{$^{\dagger}$} & \cirange{61.87}{64.53} & 11.89 & \cirange{10.47}{13.40} & 6.67 & \cirange{5.84}{7.56} \\
& External judge & \uline{68.64}\phantom{$^{\dagger}$} & \cirange{67.31}{69.90} & \uline{21.62} & \cirange{19.77}{23.56} & \textbf{3.75} & \cirange{3.14}{4.46} \\
& \ours{} (Ours) & \textbf{68.68}\phantom{$^{\dagger}$} & \cirange{67.42}{69.94} & \textbf{24.22} & \cirange{22.28}{26.18} & \uline{5.21} & \cirange{4.44}{6.00} \\
\midrule
\multicolumn{8}{@{}l}{\emph{RealWorldQA ($n=628$)}} \\
\midrule
\multirow{5}{*}{\makecell[l]{Claude 4.5\\Haiku}}
& Base & 62.74\phantom{$^{\dagger}$} & \cirange{58.92}{66.56} & \multicolumn{2}{c}{--} & \multicolumn{2}{c}{--} \\
& CoT & 60.67\phantom{$^{\dagger}$} & \cirange{56.69}{64.49} & 10.26 & \cirange{6.56}{14.29} & \uline{9.39} & \cirange{6.61}{12.34} \\
& Self-refine & \uline{63.85}\phantom{$^{\dagger}$} & \cirange{60.03}{67.51} & 27.35 & \cirange{21.67}{33.07} & 14.47 & \cirange{11.11}{18.01} \\
& External judge & 60.83\phantom{$^{\dagger}$} & \cirange{57.01}{64.65} & \textbf{49.15} & \cirange{42.68}{55.56} & 32.23 & \cirange{27.65}{36.85} \\
& \ours{} (Ours) & \textbf{69.11}$^{\dagger}$ & \cirange{65.45}{72.77} & \uline{31.20} & \cirange{25.33}{37.22} & \textbf{8.38} & \cirange{5.74}{11.20} \\
\midrule
\multirow{5}{*}{\makecell[l]{Gemini 2.5\\Flash Lite}}
& Base & 65.13\phantom{$^{\dagger}$} & \cirange{61.47}{68.79} & \multicolumn{2}{c}{--} & \multicolumn{2}{c}{--} \\
& CoT & 66.72\phantom{$^{\dagger}$} & \cirange{63.06}{70.38} & 13.24 & \cirange{8.91}{17.89} & 4.65 & \cirange{2.72}{6.78} \\
& Self-refine & 68.15\phantom{$^{\dagger}$} & \cirange{64.49}{71.65} & 14.16 & \cirange{9.70}{18.94} & \uline{2.93} & \cirange{1.46}{4.69} \\
& External judge & \uline{71.34}\phantom{$^{\dagger}$} & \cirange{67.68}{74.84} & \uline{34.25} & \cirange{28.02}{40.61} & 8.80 & \cirange{6.15}{11.64} \\
& \ours{} (Ours) & \textbf{76.59}$^{\dagger}$ & \cirange{73.25}{79.93} & \textbf{34.70} & \cirange{28.44}{41.09} & \textbf{0.98} & \cirange{0.24}{2.01} \\
\midrule
\multirow{5}{*}{GPT-5 Nano}
& Base & 70.06\phantom{$^{\dagger}$} & \cirange{66.40}{73.56} & \multicolumn{2}{c}{--} & \multicolumn{2}{c}{--} \\
& CoT & 71.50\phantom{$^{\dagger}$} & \cirange{67.84}{75.00} & \uline{27.13} & \cirange{20.88}{33.69} & 9.55 & \cirange{6.88}{12.35} \\
& Self-refine & 71.97\phantom{$^{\dagger}$} & \cirange{68.47}{75.47} & 26.60 & \cirange{20.35}{33.16} & 8.64 & \cirange{6.11}{11.32} \\
& External judge & \textbf{80.25}\phantom{$^{\dagger}$} & \cirange{77.07}{83.28} & \textbf{49.47} & \cirange{42.31}{56.60} & \uline{6.59} & \cirange{4.37}{9.01} \\
& \ours{} (Ours) & \uline{75.64}\phantom{$^{\dagger}$} & \cirange{72.14}{78.98} & 26.60 & \cirange{20.40}{33.00} & \textbf{3.41} & \cirange{1.82}{5.20} \\
\midrule
\multicolumn{8}{@{}l}{\emph{Visual7W ($n=5{,}000$)}} \\
\midrule
\multirow{5}{*}{\makecell[l]{Claude 4.5\\Haiku}}
& Base & 71.60\phantom{$^{\dagger}$} & \cirange{70.34}{72.86} & \multicolumn{2}{c}{--} & \multicolumn{2}{c}{--} \\
& CoT & 74.68\phantom{$^{\dagger}$} & \cirange{73.48}{75.88} & 21.13 & \cirange{19.06}{23.31} & \uline{4.08} & \cirange{3.46}{4.78} \\
& Self-refine & 73.86\phantom{$^{\dagger}$} & \cirange{72.66}{75.06} & 25.35 & \cirange{23.12}{27.65} & 6.90 & \cirange{6.10}{7.77} \\
& External judge & \uline{78.20}\phantom{$^{\dagger}$} & \cirange{77.06}{79.34} & \textbf{47.89} & \cirange{45.30}{50.47} & 9.78 & \cirange{8.83}{10.79} \\
& \ours{} (Ours) & \textbf{81.26}$^{\dagger}$ & \cirange{80.18}{82.34} & \uline{37.04} & \cirange{34.54}{39.56} & \textbf{1.20} & \cirange{0.87}{1.56} \\
\midrule
\multirow{5}{*}{\makecell[l]{Gemini 2.5\\Flash Lite}}
& Base & 75.44\phantom{$^{\dagger}$} & \cirange{74.24}{76.64} & \multicolumn{2}{c}{--} & \multicolumn{2}{c}{--} \\
& CoT & 78.48\phantom{$^{\dagger}$} & \cirange{77.34}{79.56} & 20.36 & \cirange{18.15}{22.67} & 2.60 & \cirange{2.10}{3.14} \\
& Self-refine & 81.84\phantom{$^{\dagger}$} & \cirange{80.76}{82.92} & 31.68 & \cirange{29.14}{34.38} & \uline{1.83} & \cirange{1.42}{2.30} \\
& External judge & \uline{82.66}\phantom{$^{\dagger}$} & \cirange{81.58}{83.67} & \textbf{45.52} & \cirange{42.78}{48.34} & 5.25 & \cirange{4.58}{5.99} \\
& \ours{} (Ours) & \textbf{85.22}$^{\dagger}$ & \cirange{84.21}{86.23} & \uline{45.28} & \cirange{42.50}{48.07} & \textbf{1.78} & \cirange{1.37}{2.21} \\
\midrule
\multirow{5}{*}{GPT-5 Nano}
& Base & 80.20\phantom{$^{\dagger}$} & \cirange{79.06}{81.28} & \multicolumn{2}{c}{--} & \multicolumn{2}{c}{--} \\
& CoT & 81.86\phantom{$^{\dagger}$} & \cirange{80.78}{82.94} & 25.15 & \cirange{22.47}{27.88} & 4.14 & \cirange{3.53}{4.77} \\
& Self-refine & 83.82\phantom{$^{\dagger}$} & \cirange{82.81}{84.83} & 26.26 & \cirange{23.55}{29.12} & \textbf{1.97} & \cirange{1.58}{2.44} \\
& External judge & \textbf{85.08}\phantom{$^{\dagger}$} & \cirange{84.07}{86.03} & \textbf{41.41} & \cirange{38.38}{44.48} & \uline{4.14} & \cirange{3.56}{4.80} \\
& \ours{} (Ours) & \uline{84.44}\phantom{$^{\dagger}$} & \cirange{83.36}{85.45} & \uline{38.79} & \cirange{35.76}{41.83} & 4.29 & \cirange{3.67}{4.93} \\
\bottomrule
\end{tabular*}
\end{table}

\subsection{Unseen Targets on RealWorldQA}
\label{sec:supp:rwqa_unseen}

The unseen targets of \Cref{tab:unseen} answer 84 to 92\% of the A-OKVQA test questions correctly at Base, which leaves few errors to repair. We therefore also evaluate the three same-family unseen targets on RealWorldQA, which is unseen in training as well and where their base accuracy is about 77\% (\Cref{tab:supp:rwqa_unseen}).

\begin{table}[H]
\caption{\textbf{Zero-shot transfer to unseen target models on RealWorldQA, with 95\% confidence intervals} (\Cref{tab:unseen}). Markers as in \Cref{tab:main}.}
\label{tab:supp:rwqa_unseen}
\centering
\footnotesize
\setlength{\tabcolsep}{4pt}
\renewcommand{\arraystretch}{1.05}
\begin{tabular*}{\textwidth}{@{\extracolsep{\fill}} ll r@{\hspace{3pt}}l r@{\hspace{3pt}}l r@{\hspace{3pt}}l @{}}
\toprule
Target Model & Method
& \multicolumn{2}{c}{\makecell{Overall Accuracy\\(\%, $\uparrow$)}}
& \multicolumn{2}{c}{\makecell{Repair Rate\\(\%, $\uparrow$)}}
& \multicolumn{2}{c}{\makecell{Harm Rate\\(\%, $\downarrow$)}} \\
\midrule
\multirow{5}{*}{\makecell[l]{Claude 4.5\\Opus}}
& Base & 76.91\phantom{$^{\dagger}$} & \cirange{73.57}{80.10} & \multicolumn{2}{c}{--} & \multicolumn{2}{c}{--} \\
& CoT & 76.11\phantom{$^{\dagger}$} & \cirange{72.77}{79.46} & 8.28 & \cirange{4.11}{13.04} & 3.52 & \cirange{2.02}{5.26} \\
& Self-refine & 77.07\phantom{$^{\dagger}$} & \cirange{73.73}{80.25} & 3.45 & \cirange{0.71}{6.75} & \textbf{0.83} & \cirange{0.20}{1.70} \\
& External judge & \textbf{78.18}\phantom{$^{\dagger}$} & \cirange{74.84}{81.37} & \uline{14.48} & \cirange{8.96}{20.42} & \uline{2.69} & \cirange{1.41}{4.23} \\
& \ours{} (Ours) & \uline{78.03}\phantom{$^{\dagger}$} & \cirange{74.84}{81.21} & \textbf{33.79} & \cirange{26.19}{41.73} & 8.70 & \cirange{6.28}{11.27} \\
\midrule
\multirow{5}{*}{\makecell[l]{Gemini 2.5\\Flash}}
& Base & 76.75\phantom{$^{\dagger}$} & \cirange{73.41}{79.94} & \multicolumn{2}{c}{--} & \multicolumn{2}{c}{--} \\
& CoT & \uline{77.71}\phantom{$^{\dagger}$} & \cirange{74.36}{80.89} & \uline{15.75} & \cirange{10.00}{21.97} & 3.53 & \cirange{1.94}{5.29} \\
& Self-refine & 77.55\phantom{$^{\dagger}$} & \cirange{74.20}{80.73} & 4.79 & \cirange{1.50}{8.55} & \uline{0.41} & \cirange{0.00}{1.05} \\
& External judge & \uline{77.71}\phantom{$^{\dagger}$} & \cirange{74.36}{80.89} & 4.11 & \cirange{1.32}{7.69} & \textbf{0.00} & \cirange{0.00}{0.00} \\
& \ours{} (Ours) & \textbf{81.85}$^{\dagger}$ & \cirange{78.82}{84.87} & \textbf{33.56} & \cirange{26.00}{41.35} & 3.53 & \cirange{1.99}{5.26} \\
\midrule
\multirow{5}{*}{GPT-5.2}
& Base & 77.23\phantom{$^{\dagger}$} & \cirange{73.89}{80.41} & \multicolumn{2}{c}{--} & \multicolumn{2}{c}{--} \\
& CoT & 77.23\phantom{$^{\dagger}$} & \cirange{74.04}{80.41} & 16.08 & \cirange{10.29}{22.39} & \uline{4.74} & \cirange{2.94}{6.71} \\
& Self-refine & \uline{78.18}\phantom{$^{\dagger}$} & \cirange{75.00}{81.37} & \uline{20.28} & \cirange{13.79}{27.14} & \uline{4.74} & \cirange{2.95}{6.73} \\
& External judge & 77.07\phantom{$^{\dagger}$} & \cirange{73.73}{80.25} & 0.70 & \cirange{0.00}{2.29} & \textbf{0.41} & \cirange{0.00}{1.05} \\
& \ours{} (Ours) & \textbf{83.44}$^{\dagger}$ & \cirange{80.57}{86.31} & \textbf{49.65} & \cirange{41.50}{57.69} & 6.60 & \cirange{4.47}{8.88} \\
\bottomrule
\end{tabular*}
\end{table}

\ours{} significantly outperforms Base and every baseline for Gemini 2.5 Flash and GPT-5.2. It gains 5.10 and 6.21 points over Base, compared with 0.17 and 0.87 points for the same targets on A-OKVQA, and its margin over the strongest baseline grows from 0.17 and 0.26 points on A-OKVQA to 4.14 and 5.26 points. For Claude 4.5 Opus, \ours{} more than doubles the repair rate of the external judge (33.79\% vs.\ 14.48\%) and is on par with it in accuracy. This establishes joint transfer to an unseen benchmark and unseen targets, where the repair advantage of \ours{} turns into a significant practical benefit for Gemini 2.5 Flash and GPT-5.2.

\subsection{Open-Weight Target}
\label{sec:supp:openweight}

\ours{} needs only black-box access, so it also applies to open-weight models. We evaluate it on the open-weight Inkling-Small \citep{tml2026inkling} on all four benchmarks, with Inkling as the external judge. $\mathcal{G}_\theta$ is used without retraining, with the model's name as the target identifier, so both the target and its identifier are unseen in training. We also include two further methods. TTAug \citep{kaya2026efficient} augments each question into 16 inputs and averages their next-token probabilities at every decoding step, following its default setup, and VISER \citep{izadi2025viser} is the training-free visual prompting method of \Cref{sec:supp:controls_ci}.

\begin{table}[H]
\caption{\textbf{Open-weight target (Inkling-Small).} Overall accuracy (\%) of \Cref{tab:openweight} with 95\% confidence intervals. $^{\dagger}$ marks accuracy that is significantly above Base and every other method.}
\label{tab:supp:openweight}
\centering
\footnotesize
\setlength{\tabcolsep}{4pt}
\renewcommand{\arraystretch}{1.05}
\begin{tabular*}{\textwidth}{@{\extracolsep{\fill}} l r@{\hspace{3pt}}l r@{\hspace{3pt}}l r@{\hspace{3pt}}l r@{\hspace{3pt}}l @{}}
\toprule
Method & \multicolumn{2}{c}{A-OKVQA} & \multicolumn{2}{c}{VCR} & \multicolumn{2}{c}{RealWorldQA} & \multicolumn{2}{c}{Visual7W} \\
\midrule
Base & 70.39\phantom{$^{\dagger}$} & \cirange{67.77}{73.01} & 50.22\phantom{$^{\dagger}$} & \cirange{48.84}{51.60} & 67.04\phantom{$^{\dagger}$} & \cirange{63.38}{70.70} & 67.38\phantom{$^{\dagger}$} & \cirange{66.08}{68.68} \\
CoT & 69.34\phantom{$^{\dagger}$} & \cirange{66.64}{71.97} & 51.28\phantom{$^{\dagger}$} & \cirange{49.90}{52.66} & 65.45\phantom{$^{\dagger}$} & \cirange{61.78}{69.11} & 68.46\phantom{$^{\dagger}$} & \cirange{67.16}{69.74} \\
Self-refine & 75.11\phantom{$^{\dagger}$} & \cirange{72.58}{77.55} & 51.44\phantom{$^{\dagger}$} & \cirange{50.06}{52.82} & 67.99\phantom{$^{\dagger}$} & \cirange{64.33}{71.66} & 71.26\phantom{$^{\dagger}$} & \cirange{70.00}{72.52} \\
External judge & 75.72\phantom{$^{\dagger}$} & \cirange{73.19}{78.17} & 53.86\phantom{$^{\dagger}$} & \cirange{52.48}{55.24} & 69.11\phantom{$^{\dagger}$} & \cirange{65.45}{72.77} & 74.96\phantom{$^{\dagger}$} & \cirange{73.76}{76.16} \\
\midrule
TTAug & 76.33\phantom{$^{\dagger}$} & \cirange{73.89}{78.78} & 54.30\phantom{$^{\dagger}$} & \cirange{52.92}{55.68} & 68.47\phantom{$^{\dagger}$} & \cirange{64.81}{72.13} & 72.56\phantom{$^{\dagger}$} & \cirange{71.32}{73.80} \\
VISER & 67.34\phantom{$^{\dagger}$} & \cirange{64.63}{70.04} & 51.40\phantom{$^{\dagger}$} & \cirange{50.02}{52.78} & 66.72\phantom{$^{\dagger}$} & \cirange{63.06}{70.38} & 68.92\phantom{$^{\dagger}$} & \cirange{67.64}{70.20} \\
\midrule
\ours{} (Ours) & \textbf{85.85}$^{\dagger}$ & \cirange{83.84}{87.86} & \textbf{57.74}$^{\dagger}$ & \cirange{56.36}{59.10} & \textbf{73.41}$^{\dagger}$ & \cirange{69.90}{76.75} & \textbf{79.18}$^{\dagger}$ & \cirange{78.04}{80.30} \\
\bottomrule
\end{tabular*}
\end{table}

\ours{} is significantly more accurate than Base and every other method on all four benchmarks. TTAug improves over Base by a few points, in line with the gains reported in its paper, while \ours{} queries the target once instead of decoding 16 inputs in parallel. Without retraining, \ours{} therefore extends to an open-weight target and outperforms every baseline there.

\FloatBarrier
\section{Controls and Audits}
\label{sec:supp:controls}

This appendix gives the setups and the full results for the ablations, controls, and audits of \Cref{sec:exp-ablation}. They test whether the gains of \ours{} come from the learned, input-specific content of its hints, or instead from a small model passing on its own answer, from generic prompting, or from answer leakage. The appendix follows the four explanations tested in \Cref{sec:exp-ablation}, which are static prompts (\Cref{sec:supp:static}), visual grounding (\Cref{sec:supp:controls_ci}), answer heuristics (\Cref{sec:supp:heuristics}), and generator training (\Cref{sec:supp:ablation}). It then gives the answer-leakage audit at its three levels (\Cref{sec:supp:leakage}) and the unhinted resampling control (\Cref{sec:supp:resample}). \Cref{tab:supp:controls} reports 95\% confidence intervals for the controls and baselines of \Cref{tab:ablation}, and significance refers to the paired bootstrap of \Cref{sec:supp:eval}.

\begin{table}[H]
\caption{\textbf{Controls and baselines on the A-OKVQA test set, with 95\% confidence intervals} (\Cref{tab:ablation}).}
\label{tab:supp:controls}
\centering
\footnotesize
\setlength{\tabcolsep}{4pt}
\renewcommand{\arraystretch}{1.05}
\resizebox{\textwidth}{!}{%
\begin{tabular}{@{}ll r@{\hspace{3pt}}l r@{\hspace{3pt}}l r@{\hspace{3pt}}l @{}}
\toprule
Target Model & Method
& \multicolumn{2}{c}{\makecell{Overall Accuracy\\(\%, $\uparrow$)}}
& \multicolumn{2}{c}{\makecell{Repair Rate\\(\%, $\uparrow$)}}
& \multicolumn{2}{c}{\makecell{Harm Rate\\(\%, $\downarrow$)}} \\
\midrule
\multirow{9}{*}{\makecell[l]{Claude 4.5\\Haiku}}
& Base & 78.25 & \cirange{75.89}{80.61} & \multicolumn{2}{c}{--} & \multicolumn{2}{c}{--} \\
& GEPA~\citep{agrawal2026gepa} & 75.37 & \cirange{72.84}{77.82} & 14.06 & \cirange{9.92}{18.44} & 7.59 & \cirange{5.89}{9.36} \\
& Grounding instruction & 76.24 & \cirange{73.80}{78.69} & 18.88 & \cirange{14.23}{23.94} & 7.81 & \cirange{6.09}{9.63} \\
& VISER~\citep{izadi2025viser} & 75.20 & \cirange{72.75}{77.73} & 14.06 & \cirange{9.88}{18.47} & 7.81 & \cirange{6.12}{9.62} \\
& Mismatched-image hint & 74.85 & \cirange{72.31}{77.38} & 20.88 & \cirange{15.94}{26.02} & 10.16 & \cirange{8.19}{12.16} \\
& Prophet~\citep{yu2025prophet} & 75.02 & \cirange{72.49}{77.47} & 24.90 & \cirange{19.61}{30.38} & 11.05 & \cirange{9.04}{13.14} \\
& RICES~\citep{yang2022pica,alayrac2022flamingo} ($k{=}8$) & 77.55 & \cirange{75.11}{80.00} & 15.66 & \cirange{11.28}{20.32} & 5.25 & \cirange{3.83}{6.75} \\
& LM-Guided CoT~\citep{lee2024lmguidedcot} & 76.33 & \cirange{73.89}{78.78} & 19.28 & \cirange{14.49}{24.30} & 7.81 & \cirange{6.10}{9.61} \\
& \ours{} (Ours) & 84.28 & \cirange{82.09}{86.45} & 49.40 & \cirange{43.09}{55.56} & 6.03 & \cirange{4.52}{7.63} \\
\midrule
\multirow{9}{*}{\makecell[l]{Gemini 2.5\\Flash Lite}}
& Base & 82.88 & \cirange{80.70}{85.06} & \multicolumn{2}{c}{--} & \multicolumn{2}{c}{--} \\
& GEPA~\citep{agrawal2026gepa} & 84.19 & \cirange{82.01}{86.29} & 33.16 & \cirange{26.70}{39.77} & 5.27 & \cirange{3.90}{6.71} \\
& Grounding instruction & 85.50 & \cirange{83.41}{87.51} & 32.14 & \cirange{25.69}{38.91} & 3.48 & \cirange{2.34}{4.69} \\
& VISER~\citep{izadi2025viser} & 79.56 & \cirange{77.21}{81.83} & 22.96 & \cirange{17.13}{29.13} & 8.75 & \cirange{6.98}{10.61} \\
& Mismatched-image hint & 81.40 & \cirange{79.13}{83.58} & 12.24 & \cirange{7.83}{17.02} & 4.32 & \cirange{3.07}{5.66} \\
& Prophet~\citep{yu2025prophet} & 86.11 & \cirange{84.10}{88.12} & 41.84 & \cirange{34.95}{48.78} & 4.74 & \cirange{3.44}{6.14} \\
& RICES~\citep{yang2022pica,alayrac2022flamingo} ($k{=}8$) & 84.89 & \cirange{82.79}{86.90} & 21.94 & \cirange{16.27}{27.88} & 2.11 & \cirange{1.26}{3.07} \\
& LM-Guided CoT~\citep{lee2024lmguidedcot} & 85.33 & \cirange{83.23}{87.34} & 32.14 & \cirange{25.67}{38.78} & 3.69 & \cirange{2.53}{4.93} \\
& \ours{} (Ours) & 90.57 & \cirange{88.73}{92.31} & 61.73 & \cirange{54.83}{68.56} & 3.48 & \cirange{2.29}{4.80} \\
\midrule
\multirow{9}{*}{GPT-5 Nano}
& Base & 84.37 & \cirange{82.27}{86.47} & \multicolumn{2}{c}{--} & \multicolumn{2}{c}{--} \\
& GEPA~\citep{agrawal2026gepa} & 85.68 & \cirange{83.67}{87.69} & 29.05 & \cirange{22.54}{35.94} & 3.83 & \cirange{2.66}{5.06} \\
& Grounding instruction & 84.63 & \cirange{82.53}{86.64} & 31.84 & \cirange{25.26}{38.81} & 5.59 & \cirange{4.18}{7.09} \\
& VISER~\citep{izadi2025viser} & 83.76 & \cirange{81.57}{85.85} & 29.05 & \cirange{22.58}{35.80} & 6.11 & \cirange{4.65}{7.68} \\
& Mismatched-image hint & 84.02 & \cirange{81.83}{86.11} & 29.61 & \cirange{23.03}{36.42} & 5.90 & \cirange{4.46}{7.44} \\
& Prophet~\citep{yu2025prophet} & 85.07 & \cirange{82.97}{87.07} & 27.37 & \cirange{20.95}{34.05} & 4.24 & \cirange{3.02}{5.56} \\
& RICES~\citep{yang2022pica,alayrac2022flamingo} ($k{=}8$) & 84.89 & \cirange{82.79}{86.90} & 17.88 & \cirange{12.43}{23.67} & 2.69 & \cirange{1.73}{3.76} \\
& LM-Guided CoT~\citep{lee2024lmguidedcot} & 85.76 & \cirange{83.67}{87.77} & 39.66 & \cirange{32.54}{46.91} & 5.69 & \cirange{4.28}{7.20} \\
& \ours{} (Ours) & 88.12 & \cirange{86.11}{90.04} & 45.81 & \cirange{38.46}{53.05} & 4.04 & \cirange{2.73}{5.42} \\
\bottomrule
\end{tabular}%
}
\end{table}

\subsection{Static Prompts}
\label{sec:supp:static}

These controls test whether a better prompt for the whole task is enough, with one prompt shared by all inputs. The \emph{failure-mode checklist} prepends the 12-mode taxonomy (\Cref{sec:supp:taxonomy}) to every question. For \emph{GEPA} \citep{agrawal2026gepa}, we optimize one A-OKVQA task prompt in DSPy \citep{khattab2023dspy}, with the \texttt{auto=medium} budget and GPT-5 as the reflection model, using only the training pool of \ours{} with a deterministic split grouped by image. The single optimized prompt is applied to every test question under the same decoding as Base and is reproduced at the end of this subsection.

The checklist does not improve accuracy reliably (\Cref{tab:ablation}). \ours{} is significantly more accurate than GEPA on all three targets and repairs more errors on each (\Cref{tab:supp:controls}). One prompt for the whole task cannot say what matters in a particular image.

{\bf The optimized prompt.} The system prompt and the user template found by GEPA are reproduced below.

\promptbox{%
{\bfseries System prompt}\par
You will be given a single multiple-choice visual question: an image plus four answer options labeled A, B, C, D. For each item, select exactly one option letter and produce a very concise (one or two short sentences) justification grounded in visible evidence and basic commonsense/factual knowledge. Follow these rules every time: 1) Quick scan: first scan the entire image to find the relevant region(s) to answer the question (e.g., sky area for weather, near the ground for objects, close-ups for faces). 2) Visible grounding: base your choice on 1\textendash{}2 concrete, specific visual cues \textemdash{} mention attributes such as color, shape, position, texture, and context (examples: "uniform gray sky with no blue patches," "metal pole with coin slot and two dials at curb," "green grass and painted yard lines," "wide deck with four trucks/wheels visible beneath"). Avoid vague impressions like "looks like" or "seems." 3) Object- and scene-specific cues: use these domain rules when relevant: - Sky/weather: overcast = largely uniform gray sky, diffused light, no visible sun or blue; mostly cloudy = large cloud coverage but some visible blue/brightness; mostly sunny = mostly blue sky with some clouds and distinct shadows; clear = entirely blue sky, bright sun, sharp shadows. - Scooters vs skateboards: handlebars \& narrow deck + hands on bar = scooter; wider deck with 4 trucks/wheels and no handlebar = skateboard. - Birds and other animals: long bill + throat pouch = pelican; long arched neck = swan; hooked beak + talons = bird of prey. - Boats/harbors: presence of docks, water, and moored small boats indicates harbor/water context. - Trash/containers: dumpsters have large rectangular shape, lids or signage, and are often near buildings/curbs. Mention visible features that identify them (wheels, lid, sticker). 4) Process of elimination: when helpful, briefly state one clear mismatch for at least one eliminated option (e.g., "not A because no handlebars are visible"), but keep justification concise. 5) Ambiguity: even when ambiguous, choose the single best-supported option \textemdash{} do not hedge (avoid "maybe," "might be," etc.). 6) Concision and format: output exactly one letter (A, B, C, or D) immediately followed by one or two short sentences of justification. No extra text, no lists, no headings. 7) Image-safety / ethics: never identify or name real people in images, and do not make sensitive inferences about a person's identity (race, religion, health, criminality, sexual life, etc.). 8) Grayscale / color note: if the image is grayscale, describe relative tone/contrast (light, dark, uniform) rather than color names. 9) Do not memorize or reveal dataset answer keys or prior example labels \textemdash{} treat each item independently. Use concise, direct, image-evidence-based reasoning grounded in the cues above.\par
You answer multiple-choice questions from user-provided inputs.\par
Return exactly one JSON object with keys "answer" and "reasoning".\par
- "answer": MUST be one of the option letters shown in the question (e.g., "A","B","C", etc.).\par
- "reasoning": 1-3 concise sentences.\par
No extra text, no markdown, no additional keys, no leading/trailing whitespace outside the JSON.\par
Required output format:\par
\{"answer":"<LETTER>","reasoning":"<1-3 concise sentences>"\}\par
\medskip
{\bfseries User template}\par
Use the user-provided inputs to answer.\par
Question:\par
\{QUESTION\}
}

\subsection{Visual-Grounding Controls}
\label{sec:supp:controls_ci}

These controls test whether the gains come from the visual evidence in each hint or from a general push to look more carefully. The \emph{grounding instruction} is a static instruction, ``Inspect the relevant regions and ground your answer in visual evidence.'' \emph{VISER} \citep{izadi2025viser} is a training-free visual prompting method. Following its default setup, we draw three equally spaced 1-pixel black lines that split the image into four bands numbered 1 to 4 on the left, and add the prompt ``Scan the image sequentially based on horizontal lines''. For the \emph{mismatched-image hint}, $\mathcal{G}_\theta$ writes the hint from a mismatched image, and the target answers with this hint and the correct image. This tests whether the hints work through the visual evidence of each image or through their form alone.

\ours{} is significantly more accurate than all three on all three targets. The grounding instruction significantly helps Gemini 2.5 Flash Lite but significantly hurts Claude 4.5 Haiku, and VISER significantly lowers accuracy on both, with no significant change on GPT-5 Nano for either. We suspect that VISER's lines, designed for binding problems such as counting and visual search, distract the target on our questions, which mostly require general reasoning about scenes and knowledge. When the hint is written from a mismatched image, the gain of \ours{} disappears and accuracy falls to or below Base on all three targets.

\subsection{Answer Heuristics}
\label{sec:supp:heuristics}

These baselines test whether a small model's answer or retrieved examples can take the place of the hint. \emph{Prophet} \citep{yu2025prophet} feeds an LLM the candidate answers and confidence scores of a small VQA model. Following it, we give the target the choice of the LoRA-tuned answerer of \Cref{sec:supp:direct} and its softmax confidence over the option letters, through the same prompt as \ours{}, so only the hint content differs. A typical injected hint reads ``A smaller vision-language model (Qwen3-VL-4B) answered B (bike) with 87\% confidence.'' \emph{RICES}, retrieval-based in-context example selection \citep{yang2022pica,alayrac2022flamingo}, prepends the training examples most similar to a test input as few-shot demonstrations. We use the 8 training questions nearest to each test question by CLIP ViT-L/14 \citep{radford2021clip} image and question similarity, drawn from the A-OKVQA training split.

\ours{} is significantly more accurate than both on all three targets. Prophet lowers the accuracy of Claude 4.5 Haiku below Base even though the injected answers are more accurate than its own. It repairs 24.90\% of the errors of Claude 4.5 Haiku but harms 11.05\% of its correct answers, the highest harm rate in \Cref{tab:ablation}. This target also has the highest harm rate among our targets in most settings (\Cref{tab:main}) and appears sensitive to injected content. We suspect that it sometimes treats another model's answer as a distractor and second-guesses a correct answer.

\subsubsection{Direct Accuracy of the Hint Generator and of a LoRA-Tuned Answerer}
\label{sec:supp:direct}

\Cref{tab:supp:direct} reports how accurately $\mathcal{G}_\theta$ answers the questions itself, next to a LoRA-tuned 4B answerer. The answerer is Qwen3-VL-4B-Instruct fine-tuned with LoRA to answer directly, using the same SFT recipe and the same training budget as $\mathcal{G}_\theta$ on the A-OKVQA training split.

\begin{table}[H]
\caption{\textbf{Direct answering accuracy (\%) of the hint generator and of a LoRA-tuned 4B answerer,} with 95\% confidence intervals.}
\label{tab:supp:direct}
\centering
\small
\begin{tabular}{l r@{\hspace{3pt}}l r@{\hspace{3pt}}l}
\toprule
Benchmark & \multicolumn{2}{c}{$\mathcal{G}_\theta$ (hint generator)} & \multicolumn{2}{c}{LoRA-tuned 4B answerer} \\
\midrule
A-OKVQA & 77.55 & \cirange{75.11}{80.00} & 85.76 & \cirange{83.67}{87.77} \\
VCR & 66.00 & \cirange{64.68}{67.30} & 62.12 & \cirange{60.78}{63.46} \\
Visual7W & 78.24 & \cirange{77.10}{79.38} & 81.74 & \cirange{80.66}{82.80} \\
RealWorldQA & 66.72 & \cirange{63.06}{70.22} & 69.27 & \cirange{65.61}{72.77} \\
\bottomrule
\end{tabular}
\end{table}

$\mathcal{G}_\theta$ is optimized to steer the target rather than to answer, and it is a worse answerer than the LoRA-tuned answerer on three of the four benchmarks. The \ours{}-hinted target is nevertheless more accurate than this answerer in 9 of the 12 target-benchmark cells of \Cref{tab:main}, by 2.36 to 7.32 points. In the other 3, all for Claude 4.5 Haiku, it is within 1.5 points of the answerer. The gains of \ours{} are therefore not explained simply by a 4B model having been trained on this dataset, and are more plausibly due to the hints themselves. On RealWorldQA, the three unseen targets of \Cref{sec:supp:rwqa_unseen} already answer better than this answerer at Base (about 77\% vs.\ 69.27\%), and \ours{} still lifts them to 78.03 to 83.44\%, with significant gains on two of them.

\subsubsection{Cascade with the LoRA-Tuned Answerer}
\label{sec:supp:cascade}

An LLM cascade in the style of FrugalGPT \citep{chen2023frugalgpt} returns the answer of the LoRA-tuned answerer of \Cref{sec:supp:direct} when its confidence is at least 95\%, a threshold calibrated on a held-out development set, and queries the unhinted target otherwise. The answerer keeps 779 of the 1{,}145 questions on A-OKVQA and 209 of the 628 on RealWorldQA. Repair and harm are measured against the Base answers of the target.

\begin{table}[H]
\caption{\textbf{Cascade with the LoRA-tuned answerer, with 95\% confidence intervals.}}
\label{tab:supp:cascade}
\centering
\footnotesize
\setlength{\tabcolsep}{4pt}
\renewcommand{\arraystretch}{1.05}
\begin{tabular*}{\textwidth}{@{\extracolsep{\fill}} lll r@{\hspace{3pt}}l r@{\hspace{3pt}}l r@{\hspace{3pt}}l @{}}
\toprule
Benchmark & Target Model & Method
& \multicolumn{2}{c}{\makecell{Overall Accuracy\\(\%, $\uparrow$)}}
& \multicolumn{2}{c}{\makecell{Repair Rate\\(\%, $\uparrow$)}}
& \multicolumn{2}{c}{\makecell{Harm Rate\\(\%, $\downarrow$)}} \\
\midrule
\multirow{6}{*}{A-OKVQA}
& \multirow{2}{*}{\makecell[l]{Claude 4.5\\Haiku}} & Cascade & 85.07 & \cirange{82.97}{87.07} & 38.96 & \cirange{32.93}{45.06} & 2.12 & \cirange{1.23}{3.11} \\
& & \ours{} (Ours) & 84.28 & \cirange{82.09}{86.45} & 49.40 & \cirange{43.09}{55.56} & 6.03 & \cirange{4.52}{7.63} \\
& \multirow{2}{*}{\makecell[l]{Gemini 2.5\\Flash Lite}} & Cascade & 87.77 & \cirange{85.85}{89.61} & 39.29 & \cirange{32.37}{46.15} & 2.21 & \cirange{1.35}{3.19} \\
& & \ours{} (Ours) & 90.57 & \cirange{88.73}{92.31} & 61.73 & \cirange{54.83}{68.56} & 3.48 & \cirange{2.29}{4.80} \\
& \multirow{2}{*}{GPT-5 Nano} & Cascade & 86.03 & \cirange{84.02}{88.03} & 30.17 & \cirange{23.53}{37.02} & 3.62 & \cirange{2.49}{4.85} \\
& & \ours{} (Ours) & 88.12 & \cirange{86.11}{90.04} & 45.81 & \cirange{38.46}{53.05} & 4.04 & \cirange{2.73}{5.42} \\
\midrule
\multirow{6}{*}{RealWorldQA}
& \multirow{2}{*}{\makecell[l]{Claude 4.5\\Haiku}} & Cascade & 66.08 & \cirange{62.42}{69.75} & 15.81 & \cirange{11.26}{20.63} & 4.06 & \cirange{2.23}{6.11} \\
& & \ours{} (Ours) & 69.11 & \cirange{65.45}{72.77} & 31.20 & \cirange{25.33}{37.22} & 8.38 & \cirange{5.74}{11.20} \\
& \multirow{2}{*}{\makecell[l]{Gemini 2.5\\Flash Lite}} & Cascade & 70.22 & \cirange{66.72}{73.73} & 21.46 & \cirange{16.10}{26.96} & 3.67 & \cirange{1.95}{5.60} \\
& & \ours{} (Ours) & 76.59 & \cirange{73.25}{79.93} & 34.70 & \cirange{28.44}{41.09} & 0.98 & \cirange{0.24}{2.01} \\
& \multirow{2}{*}{GPT-5 Nano} & Cascade & 71.97 & \cirange{68.47}{75.48} & 24.47 & \cirange{18.44}{30.77} & 7.73 & \cirange{5.32}{10.29} \\
& & \ours{} (Ours) & 75.64 & \cirange{72.14}{78.98} & 26.60 & \cirange{20.40}{33.00} & 3.41 & \cirange{1.82}{5.20} \\
\bottomrule
\end{tabular*}
\end{table}

\ours{} is significantly more accurate than the cascade in five of the six cases (\Cref{tab:supp:cascade}). In the sixth, Claude 4.5 Haiku on A-OKVQA, the two are on par. On A-OKVQA, where the answerer was fine-tuned, its deterministic answers decide 68\% of the questions and avoid the sampling noise of a target call, which keeps harm low. On the less saturated RealWorldQA they decide 33\%, and \ours{} is significantly more accurate for all three targets.

\subsection{Generator Training}
\label{sec:supp:ablation}

These conditions isolate what each training stage adds and test whether what the generator is trained to write matters.

{\bf Untrained hint writer.} To test whether the gains of \ours{} come from its training pipeline rather than from the presence of any contrastive hint, we use the off-the-shelf Qwen3-VL-4B-Instruct, i.e., the initialization of $\mathcal{G}_\theta$ before Step~3, as the hint writer. It is prompted to write a contrastive hint for each test question, without access to the ground-truth answer or any training. \ours{} is more accurate than the untrained hint writer on all three targets and roughly doubles its repair rate at comparable harm (\Cref{tab:ablation}). For 72 of the 1{,}145 test questions (6.3\%), the untrained model's hint showed a clear inclination toward one of the answer options despite instructions not to reveal an answer. This option reflects the untrained model's own guess and is not necessarily correct. We keep these questions in its reported accuracy.

{\bf LM-Guided CoT.} Following LM-Guided CoT \citep{lee2024lmguidedcot}, we train the same Qwen3-VL-4B-Instruct with the same LoRA recipe and compute budget as $\mathcal{G}_\theta$ to write a rationale for each question, supervised by the ground-truth rationales of A-OKVQA. At test time the target receives this rationale in place of the hint. Unlike a hint of \ours{}, a rationale is free to state the answer, and it often does, because a rationale is meant to lead to the answer. \ours{} is nevertheless significantly more accurate than LM-Guided CoT on all three targets and repairs more errors on each (\Cref{tab:supp:controls}). A rationale from the same 4B model and the same data helps less than a hint trained on verified repairs.

{\bf Training stages.} SFT accounts for most of the gain over Base, and GRPO mainly keeps the high repair rate while reducing harm. Under the paired bootstrap, the harm reduction is significant for Claude 4.5 Haiku (8.04\% to 6.03\%).

{\bf Generator size.} We chose Qwen3-VL-4B-Instruct because it can run on a personal computer and fits a modest training budget. This choice shows that anticipatory hinting can be distilled into a compact local model that may be deployed in compute-limited settings.

\subsection{Answer-Leakage Audits}
\label{sec:supp:leakage}

Because the Step~2 proposer sees the ground-truth answer, we audit the hints for answer leakage. The editor screens the training hints during data construction, and the hints that $\mathcal{G}_\theta$ writes at test time are audited at the three levels of \Cref{sec:exp-ablation}.

{\bf Editor screening.} The editor enforces no answer leakage together with the other hint constraints (contrastive form, no certainty language, and no extraneous reasoning). It required revisions for 8.4\% of first-round proposals. After the proposer--editor loop, 2.6\% of all proposed hints still failed its checks and were dropped, so no rejected hint was used to train $\mathcal{G}_\theta$.

\subsubsection{The Hint Text Does Not Reveal the Answer}
\label{sec:supp:hinttext}

{\bf Manual audit.} We randomly sampled 100 hints generated by $\mathcal{G}_\theta$ for the A-OKVQA test set and inspected each alongside its question and answer options. None revealed or semantically entailed the correct answer.

{\bf Text-only answer-recovery probes.} If a hint encoded the identity of the correct option, a learner that never sees the image or the question could recover the answer from the hint and the answer options more accurately than from the options alone. We therefore train probes on the official A-OKVQA training split and evaluate them on the held-out test split. The control probe receives only the four answer options, and the treatment probe additionally receives the target-conditioned hint generated by $\mathcal{G}_\theta$. We test three probe capacities. The \emph{frozen linear} probe uses a frozen \texttt{all-MiniLM-L6-v2} sentence encoder with an L2-regularized multinomial logistic-regression head. The \emph{frozen MLP} probe uses the same frozen encoder with a trainable $384{\rightarrow}256{\rightarrow}4$ GELU MLP. In the \emph{end-to-end} probe, MiniLM and the MLP head are fine-tuned jointly so that the representation itself can adapt to any answer-bearing pattern. Results are means over three optimization seeds (3407, 42, and 1234), with 95\% confidence intervals from 20{,}000 hierarchical paired-bootstrap replicates over seeds and test questions. Across all three hint sets and probe capacities, adding the hint never significantly improves recovery over the options-only control or over the 25\% chance level (\Cref{tab:supp:probe}), and increasing probe capacity through end-to-end fine-tuning does not change this conclusion. \Cref{sec:supp:textonly} complements these probes with a stronger reader. There, each target model answers from the question, the options, and the hint without the image.

\begin{table}[H]
\caption{\textbf{Text-only answer-recovery probes.}}
\label{tab:supp:probe}
\centering
\footnotesize
\setlength{\tabcolsep}{4pt}
\renewcommand{\arraystretch}{1.05}
\begin{tabular*}{\textwidth}{@{\extracolsep{\fill}} ll r@{\hspace{3pt}}l r@{\hspace{3pt}}l @{}}
\toprule
Target Model & Probe & \multicolumn{2}{c}{\makecell{Options only\\accuracy (\%)}} & \multicolumn{2}{c}{\makecell{Hint $+$ options\\accuracy (\%)}} \\
\midrule
\multirow{3}{*}{\makecell[l]{Claude 4.5\\Haiku}}
& Frozen linear & 24.56 & \cirange{22.03}{27.10} & 22.38 & \cirange{20.02}{24.83} \\
& Frozen MLP & 25.67 & \cirange{23.25}{28.09} & 23.60 & \cirange{20.86}{26.34} \\
& End-to-end & 27.36 & \cirange{24.56}{30.19} & 28.55 & \cirange{22.09}{37.76} \\
\midrule
\multirow{3}{*}{\makecell[l]{Gemini 2.5\\Flash Lite}}
& Frozen linear & 24.56 & \cirange{22.03}{27.10} & 23.69 & \cirange{21.24}{26.14} \\
& Frozen MLP & 25.67 & \cirange{23.25}{28.09} & 23.43 & \cirange{20.89}{25.99} \\
& End-to-end & 27.36 & \cirange{24.56}{30.19} & 23.86 & \cirange{21.53}{26.22} \\
\midrule
\multirow{3}{*}{GPT-5 Nano}
& Frozen linear & 24.56 & \cirange{22.03}{27.10} & 26.14 & \cirange{23.60}{28.67} \\
& Frozen MLP & 25.67 & \cirange{23.25}{28.09} & 23.51 & \cirange{20.98}{26.08} \\
& End-to-end & 27.36 & \cirange{24.56}{30.19} & 30.48 & \cirange{22.90}{37.33} \\
\bottomrule
\end{tabular*}
\end{table}

\subsubsection{The Hint Does Not Work Without the Image}
\label{sec:supp:textonly}

This test asks whether the hint itself leaks the answer to the target. We give each target the question and options without the image, with and without the hint (\Cref{tab:supp:textonly}). These are capable LLMs and the same models that read the hints in \ours{}.

\begin{table}[H]
\caption{\textbf{Leakage test with a text-only LLM.} Accuracy (\%) with 95\% confidence intervals when the target sees the question and options but not the image. The chance level (\%) is given in parentheses under each benchmark.}
\label{tab:supp:textonly}
\centering
\footnotesize
\setlength{\tabcolsep}{4pt}
\renewcommand{\arraystretch}{1.05}
\begin{tabular*}{\textwidth}{@{\extracolsep{\fill}} ll r@{\hspace{3pt}}l r@{\hspace{3pt}}l c @{}}
\toprule
Benchmark & Target Model & \multicolumn{2}{c}{Text only} & \multicolumn{2}{c}{Text $+$ hint} & \makecell{\ours{}\\with image} \\
\midrule
\multirow{3}{*}{\makecell[l]{A-OKVQA\\(25)}} & Claude 4.5 Haiku & 50.83 & \cirange{47.95}{53.71} & 53.19 & \cirange{50.31}{56.07} & 84.28 \\
 & Gemini 2.5 Flash Lite & 49.00 & \cirange{46.11}{51.88} & 53.54 & \cirange{50.66}{56.42} & 90.57 \\
 & GPT-5 Nano & 50.66 & \cirange{47.77}{53.54} & 54.41 & \cirange{51.53}{57.29} & 88.12 \\
\midrule
\multirow{3}{*}{\makecell[l]{VCR\\(25)}} & Claude 4.5 Haiku & 39.06 & \cirange{37.70}{40.42} & 40.22 & \cirange{38.86}{41.58} & 64.98 \\
 & Gemini 2.5 Flash Lite & 39.80 & \cirange{38.44}{41.16} & 40.74 & \cirange{39.38}{42.10} & 67.04 \\
 & GPT-5 Nano & 42.68 & \cirange{41.32}{44.06} & 41.74 & \cirange{40.38}{43.10} & 68.68 \\
\midrule
\multirow{3}{*}{\makecell[l]{Visual7W\\(25)}} & Claude 4.5 Haiku & 39.54 & \cirange{38.18}{40.90} & 43.00 & \cirange{41.62}{44.38} & 81.26 \\
 & Gemini 2.5 Flash Lite & 40.08 & \cirange{38.72}{41.44} & 46.36 & \cirange{44.98}{47.74} & 85.22 \\
 & GPT-5 Nano & 38.94 & \cirange{37.60}{40.30} & 40.02 & \cirange{38.66}{41.38} & 84.44 \\
\midrule
\multirow{3}{*}{\makecell[l]{RealWorldQA\\(38.2)}} & Claude 4.5 Haiku & 40.92 & \cirange{37.10}{44.75} & 45.06 & \cirange{41.24}{48.89} & 69.11 \\
 & Gemini 2.5 Flash Lite & 38.22 & \cirange{34.39}{42.04} & 41.24 & \cirange{37.42}{45.06} & 76.59 \\
 & GPT-5 Nano & 43.63 & \cirange{39.81}{47.45} & 42.52 & \cirange{38.69}{46.34} & 75.64 \\
\bottomrule
\end{tabular*}
\end{table}

Without the image, the models already score well above chance on A-OKVQA, VCR, and Visual7W, so the questions themselves carry some bias toward certain answers. The hint adds a few points to this in most settings, so it can carry a mild cue that is usable without the image. If the hints leaked $\mathcal{G}_\theta$'s answer, however, a capable reader could recover it from the text and approach $\mathcal{G}_\theta$'s own accuracy (\Cref{tab:supp:direct}). Instead, accuracy with the hint but without the image stays 22 to 38 points below it on every benchmark, and 24 to 44 points below \ours{} with the image. This cue therefore does not account for most of the gains of \ours{}.

\subsubsection{The Hint Does Not Hand Over the Generator's Own Answer}
\label{sec:supp:follow}

If the hints carried $\mathcal{G}_\theta$'s own answer, the hinted target would follow $\mathcal{G}_\theta$ when it is wrong. \Cref{tab:supp:follow} restricts to the questions that $\mathcal{G}_\theta$ answers incorrectly and compares each target with and without the hint.

\begin{table}[H]
\caption{\textbf{How targets answer when $\mathcal{G}_\theta$ is wrong.} Values are percentages of the questions that $\mathcal{G}_\theta$ answers incorrectly.}
\label{tab:supp:follow}
\centering
\small
\begin{tabular}{l l c c c c}
\toprule
& & \multicolumn{2}{c}{Target correct (\%)} & \multicolumn{2}{c}{\makecell{Target picks $\mathcal{G}_\theta$'s\\wrong option (\%)}} \\
\cmidrule(lr){3-4}\cmidrule(lr){5-6}
Benchmark & Target Model & \ours{} & Base & \ours{} & Base \\
\midrule
\multirow{3}{*}{A-OKVQA} & Claude 4.5 Haiku & 59 & 57 & 26 & 28 \\
 & Gemini 2.5 Flash Lite & 70 & 60 & 20 & 24 \\
 & GPT-5 Nano & 62 & 63 & 23 & 23 \\
\midrule
\multirow{3}{*}{VCR} & Claude 4.5 Haiku & 42 & 31 & 44 & 44 \\
 & Gemini 2.5 Flash Lite & 41 & 28 & 45 & 49 \\
 & GPT-5 Nano & 36 & 35 & 45 & 48 \\
\midrule
\multirow{3}{*}{Visual7W} & Claude 4.5 Haiku & 52 & 42 & 35 & 40 \\
 & Gemini 2.5 Flash Lite & 58 & 46 & 34 & 36 \\
 & GPT-5 Nano & 55 & 54 & 31 & 31 \\
\midrule
\multirow{3}{*}{RealWorldQA} & Claude 4.5 Haiku & 50 & 40 & 34 & 42 \\
 & Gemini 2.5 Flash Lite & 56 & 41 & 34 & 44 \\
 & GPT-5 Nano & 51 & 41 & 38 & 44 \\
\bottomrule
\end{tabular}
\end{table}

When $\mathcal{G}_\theta$ is wrong, the hinted target is generally right more often than Base and chooses the wrong option of $\mathcal{G}_\theta$ no more often than without the hint, so the hints do not simply pass or imply $\mathcal{G}_\theta$'s answer on to the target.

\subsection{Unhinted Resampling Control}
\label{sec:supp:resample}

Base at test time is a single sample under each API's default settings, and API sampling changes some answers between runs. To measure this resampling noise floor, we drew a second unhinted sample for the three primary targets on A-OKVQA at the same settings and scored it against the first (\Cref{tab:supp:resample}). The same values appear in \Cref{tab:ablation} as the second Base sample.

\begin{table}[H]
\caption{\textbf{Unhinted resampling control on A-OKVQA.} A second unhinted sample is scored against the first, with 95\% confidence intervals. The last two columns repeat the rates of \ours{} from \Cref{tab:main}.}
\label{tab:supp:resample}
\centering
\footnotesize
\setlength{\tabcolsep}{4pt}
\renewcommand{\arraystretch}{1.05}
\begin{tabular*}{\textwidth}{@{\extracolsep{\fill}} l r@{\hspace{3pt}}l r@{\hspace{3pt}}l r@{\hspace{3pt}}l c c @{}}
\toprule
Target Model & \multicolumn{2}{c}{\makecell{Overall Accuracy\\(\%)}} & \multicolumn{2}{c}{\makecell{Repair Rate\\(\%)}} & \multicolumn{2}{c}{\makecell{Harm Rate\\(\%)}} & \makecell{\ours{}\\repair} & \makecell{\ours{}\\harm} \\
\midrule
Claude 4.5 Haiku & 77.21 & \cirange{74.76}{79.65} & 6.02 & \cirange{3.24}{9.15} & 3.01 & \cirange{1.94}{4.18} & 49.40 & 6.03 \\
Gemini 2.5 Flash Lite & 82.53 & \cirange{80.35}{84.72} & 8.16 & \cirange{4.52}{12.20} & 2.11 & \cirange{1.26}{3.07} & 61.73 & 3.48 \\
GPT-5 Nano & 84.98 & \cirange{82.88}{86.99} & 15.64 & \cirange{10.50}{21.15} & 2.17 & \cirange{1.32}{3.14} & 45.81 & 4.04 \\
\bottomrule
\end{tabular*}
\end{table}

This control separates hint-induced repair and harm from ordinary response variability. Resampling repairs only 6 to 16\% of Base errors, far below the 46 to 62\% that \ours{} repairs. The repair rate of \ours{} is 30 to 54 points above the floor, while its harm rate is only 1.4 to 3.0 points above it. By contrast, the repair rates of CoT and self-refine on GPT-5 Nano in \Cref{tab:main} (18.44\% and 17.88\%) fall within the resampling interval. Beyond ordinary response variability, the hints therefore add far more repairs than harms.
\FloatBarrier
\section{Gains and Base Accuracy}
\label{sec:supp:headroom}

We examine how the gains of \ours{} relate to base accuracy across the 17 target--benchmark combinations in \Cref{tab:main,tab:unseen}: the eight targets on A-OKVQA and the three primary targets on each of VCR, RealWorldQA, and Visual7W. For each combination, let $B$ denote base accuracy and $G$ the accuracy of \ours{} minus base accuracy, both in percentage points. We report Pearson's $r$ for the linear association, Spearman's $\rho$ as a rank-based check, and two-sided permutation $p$-values over the 17 combinations. The analysis asks whether stronger base models show smaller absolute gains and, if so, whether this reflects shrinking error headroom or weaker correction of the errors that remain.

{\bf Absolute gains shrink near ceiling.} $B$ and $G$ are moderately negatively associated (Pearson $r=-0.60$ with $p=0.012$, and Spearman $\rho=-0.61$ with $p=0.012$), and a least-squares fit associates each 10-point increase in base accuracy with about 1.9 points less gain. As an intuitive summary, the four combinations with base accuracy of at least 85\% gain 0.98 points on average, compared with 6.17 points for the other 13.

{\bf The reduction reflects shrinking headroom rather than weaker correction.} With repair rate $R$ and harm rate $H$ (both in percent), the gain decomposes as $G=((100-B)R-BH)/100$, so the repairable portion $100-B$ necessarily shrinks as $B$ grows. To separate this mechanical ceiling from correction quality, we normalize the gain by the remaining error headroom, $G/(100-B)$, whose association with base accuracy is weak and not significant ($r=-0.19$, $p=0.46$). Direct checks agree: the repair rate does not decline with base accuracy (a nonsignificant positive correlation, $r=+0.41$, $p=0.10$), and the harm rate shows a nonsignificant negative correlation ($r=-0.42$, $p=0.09$).

Higher base accuracy therefore places a practical ceiling on the absolute improvement because fewer errors remain to be repaired, while the ability of \ours{} to correct the remaining errors does not diminish. The net benefit of \ours{} is accordingly largest where base performance is less saturated, such as in domain- or knowledge-specific applications.

{\bf Why large repair advantages give small margins near ceiling.} The same decomposition explains the small margins on the unseen targets of \Cref{tab:unseen}. Because repair is weighted by $100-B$ and harm by $B$, one point of harm offsets 5 to 11 points of repair at their base accuracies. This is why a repair rate 1.5 to 3.8 times that of the strongest baseline yields margins of only 0.17 to 0.70 points. At a base accuracy of about 77\%, as for the same-family targets on RealWorldQA (\Cref{sec:supp:rwqa_unseen}), one point of harm offsets only about 3 points of repair, and the same repair advantage yields much larger gains.

\FloatBarrier
\section{Analysis of Harmful Hints}
\label{sec:supp:harm}

To understand how hints can hurt, we manually examined 100 cases in which \ours{} changed a base-correct response to an incorrect one. Three patterns emerged, listed in decreasing order of frequency:
\begin{enumerate}[leftmargin=*,topsep=2pt,itemsep=2pt]
  \item \textbf{Unhelpful foreground redirection.} The hint asks the target to inspect a salient foreground element that is not useful for answering the question, distracting it from the relevant evidence.
  \item \textbf{Failure despite useful guidance.} The hint correctly directs attention to relevant evidence, but the target nevertheless flips to an incorrect response. These cases are consistent with target-model capability limits or response instability rather than a factual defect in the hint.
  \item \textbf{Fabricated, unnecessary details.} The hint introduces unverifiable features that are not needed to answer the question, causing the target to search for nonexistent evidence or become distracted.
\end{enumerate}
Harmful cases are thus most often associated with attention misallocation or target-side instability, while outright fabrication in the hint is less frequent. Fabricated guidance appears in only 6 of the 100 cases, so fabrication is a rare source of harm.

\FloatBarrier
\section{Inference and Training Cost}
\label{sec:supp:cost}

{\bf Inference.} Per test query, \ours{} runs one forward pass of the 4B hint generator followed by one call to the black-box target, with no iterative target feedback. Base and CoT make one target call, self-refine makes two sequential target calls, and the external-judge baseline makes two target calls and one call to a more capable judge model, all in sequence. Because the hint generator is much smaller and faster than the frontier targets and runs before a single target call, \ours{} avoids the sequential rounds of post-response critique. The compute of the hint generator must still be counted, but it amounts to a single pass through a compact model that can run on local hardware. \Cref{sec:exp-cost} compares the cost of an accuracy point gained with \ours{} and by switching to a stronger model of the same family.

{\bf One-time training cost.} Constructing the training data for the 17{,}056 A-OKVQA training questions cost approximately \$260 in API calls (\Cref{tab:supp:cost}), including repeated target calls and extended revision rounds in the agentic loop. This cost is driven mainly by the choice of target model rather than by the method: at the prices used for this estimate, Claude 4.5 Haiku costs roughly 10--20$\times$ more than the other targets, and running the identical pipeline over the same questions with only Gemini 2.5 Flash Lite and GPT-5 Nano as targets brings the total to roughly \$150. The rest of the budget is dominated by the Step~2 proposer and editor, which are independent of the target and could be reduced further with a cheaper proposer or fewer revision rounds. The cost is incurred once, offline, when the training data for the hint generator is constructed (about 1.5 cents per training question), and it does not recur at deployment. It also scales with the size of the training set, so adapting \ours{} to a smaller domain-specific dataset would cost correspondingly less. Training the hint generator itself took about 11 hours on a single H200 GPU (2 hours for Stage~1 and 9 hours for Stage~2).

\begin{table}[H]
\caption{\textbf{Approximate one-time API cost (USD) of constructing the training data.}}
\label{tab:supp:cost}
\centering
\small
\begin{tabular}{l r l}
\toprule
Step & Cost & Main expense \\
\midrule
Step~1: base-response sampling & \$70 & \$54 in Claude 4.5 Haiku target calls \\
Step~2: agentic hint generation and optimization & \$150 & \$100 in GPT-5 proposer and editor calls \\
Step~3: hint generator training & \$40 & \$30 in Claude 4.5 Haiku target calls \\
\midrule
Total & \$260 & \\
\bottomrule
\end{tabular}
\end{table}

\FloatBarrier
\section{Qualitative Examples of Hint-Driven Error Repair}
\label{sec:supp:qualitative}

This appendix presents one representative repair example for each of the three primary target models, drawn from the A-OKVQA test set. Each example shows a question the base model answered incorrectly, the hint produced by $\mathcal{G}_\theta$, and the corrected response obtained after prepending that hint to the prompt. In every case the hint directs attention toward discriminative visual evidence without revealing the correct answer, which illustrates the contrastive design discussed in \Cref{sec:phase2}. The code and the prompts are available at \url{https://github.com/AlaaLab/test-time-hinting}.

\begingroup\raggedright

\subsection{Claude 4.5 Haiku}
\label{sec:supp:qualitative:claude}

\begin{figure}[h]
  \centering
  \includegraphics[width=0.45\textwidth]{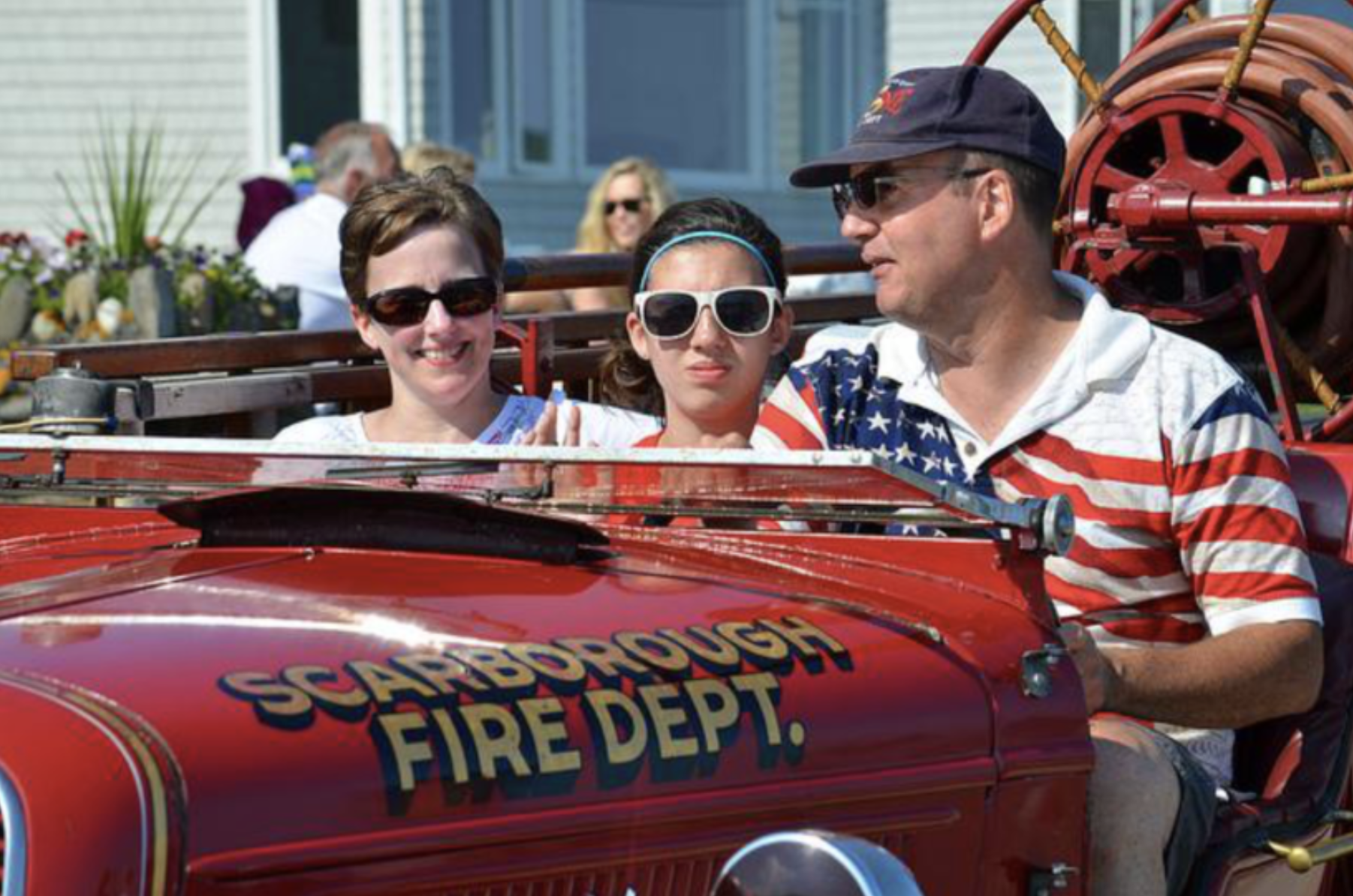}
  \caption{Qualitative example for Claude~4.5~Haiku.}
  \label{fig:supp:qual:claude}
\end{figure}

\paragraph{Question.}
\texttt{Who are these three people? A. visitors B. customers C. firefighters D. passengers}

\paragraph{Ground-truth answer.}
\texttt{A.}

\paragraph{Base response (incorrect).}
\begin{itemize}[leftmargin=*, itemsep=2pt]
  \item \textbf{Answer:} \texttt{C.}
  \item \textbf{Reasoning:} \texttt{The image shows three people sitting in a red vehicle with "SCARBOROUGH FIRE DEPT" written on it. The vehicle appears to be a fire truck, suggesting they are likely firefighters or participating in an event related to the fire department. The presence of the fire department's name on the vehicle strongly indicates their affiliation.}
\end{itemize}

\paragraph{Generated hint.}\mbox{}
\promptbox{%
\{"hint": [\newline
~~"Firefighters vs visitors: check whether they wear uniforms/helmets or just casual summer clothes and sunglasses.",\newline
~~"Working crew vs ride-along passengers: see whether they are operating gear or simply seated for a ride/photo.",\newline
~~"Focus on the people, not the truck branding; read the text but verify what their clothing and activity suggest."\newline
]\}
}

\paragraph{Hinted response (correct).}
\begin{itemize}[leftmargin=*, itemsep=2pt]
  \item \textbf{Answer:} \texttt{A.}
  \item \textbf{Reasoning:} \texttt{The image shows three people sitting in a red vehicle with "SCARBOROUGH FIRE DEPT." written on it. They are dressed in casual clothing and sunglasses, suggesting they are not actively working as firefighters. Their relaxed demeanor and the context of being in a fire department vehicle imply they are likely visitors or participants in an event.}
\end{itemize}

\subsection{Gemini 2.5 Flash Lite}
\label{sec:supp:qualitative:gemini}

\begin{figure}[h]
  \centering
  \includegraphics[width=0.45\textwidth]{figures/gemini_example.png}
  \caption{Qualitative example for Gemini~2.5~Flash~Lite.}
  \label{fig:supp:qual:gemini}
\end{figure}

\paragraph{Question.}
\texttt{Why is the guy standing in the median? A. awaiting greenlight B. washing windshields C. panhandling D. selling flowers}

\paragraph{Ground-truth answer.}
\texttt{A.}

\paragraph{Base response (incorrect).}
\begin{itemize}[leftmargin=*, itemsep=2pt]
  \item \textbf{Answer:} \texttt{C.}
  \item \textbf{Reasoning:} \texttt{The person is standing in the median of a busy street with traffic. They are not actively engaged in any activity like washing windshields or selling flowers. Their posture and location suggest they are soliciting money from passing drivers.}
\end{itemize}

\paragraph{Generated hint.}\mbox{}
\promptbox{%
\{"hint": [\newline
~~"Waiting to cross vs panhandling: check whether he's at the curb by the crosswalk under the signal, facing traffic, or positioned in the middle of the lane with a raised hand and any sign of begging.",\newline
~~"Verify the traffic light above him: is it red/stop or green/advance?",\newline
~~"Scan his hand and the ground for any flowers or items he might be selling."\newline
]\}
}

\paragraph{Hinted response (correct).}
\begin{itemize}[leftmargin=*, itemsep=2pt]
  \item \textbf{Answer:} \texttt{A.}
  \item \textbf{Reasoning:} \texttt{The image shows a person standing in the median of a street at a crosswalk. The pedestrian signal displays as a red hand, indicating that it is not safe to cross. This suggests the person is waiting for the light to change to a walk signal.}
\end{itemize}

\subsection{GPT-5 Nano}
\label{sec:supp:qualitative:gpt}

\begin{figure}[h]
  \centering
  \includegraphics[width=0.45\textwidth]{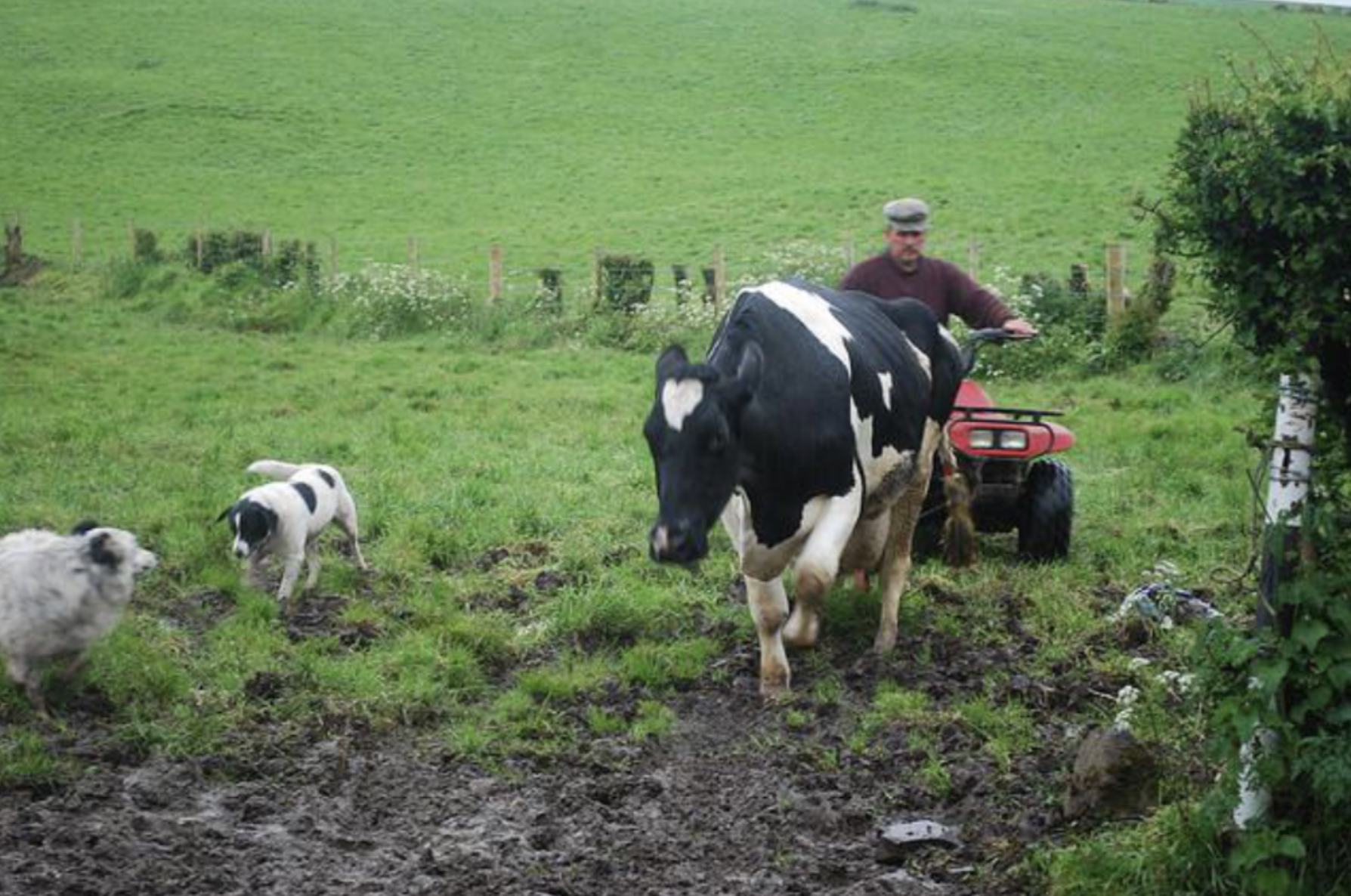}
  \caption{Qualitative example for GPT-5~Nano.}
  \label{fig:supp:qual:gpt}
\end{figure}

\paragraph{Question.}
\texttt{What vehicle is the man riding? A. tractor B. bike C. plane D. car}

\paragraph{Ground-truth answer.}
\texttt{A.}

\paragraph{Base response (incorrect).}
\begin{itemize}[leftmargin=*, itemsep=2pt]
  \item \textbf{Answer:} \texttt{B.}
  \item \textbf{Reasoning:} \texttt{The man is seated on a four-wheeled vehicle (quad bike/ATV) behind him, which aligns with the "bike" option.}
\end{itemize}

\paragraph{Generated hint.}\mbox{}
\promptbox{%
\{"hint": [\newline
~~"Tractor vs bike: check whether the vehicle has a seat, handlebars, and two wheels, or a large engine with a front loader and wide rear tires.",\newline
~~"Focus on the wheels and the presence of handlebars.",\newline
~~"Plane/car vs ground vehicle: confirm it's on the grassy ground rather than in the air or on a road."\newline
]\}
}

\paragraph{Hinted response (correct).}
\begin{itemize}[leftmargin=*, itemsep=2pt]
  \item \textbf{Answer:} \texttt{A.}
  \item \textbf{Reasoning:} \texttt{The man is sitting on a red farm vehicle behind the cow, which resembles a small tractor rather than a bicycle, car, or plane. The handlebars and farm-oriented build indicate a tractor.}
\end{itemize}

\endgroup

\FloatBarrier
\section{Licenses and Prompts}
\label{sec:supp:licenses}

This work uses publicly available datasets under their respective licenses: A-OKVQA (Apache License 2.0), VCR (AI2 custom research-only license), Visual7W (MIT License), and RealWorldQA (CC BY-ND 4.0). The hint generator is initialized from Qwen3-VL-4B-Instruct (Apache License 2.0). All proprietary models were accessed through their commercial APIs in accordance with the providers' terms of use. The study involves no human subjects, and the manual leakage audit (\Cref{sec:supp:leakage}) was conducted by the authors. Code and the prompts for all agents, baselines, and target models are available at \url{https://github.com/AlaaLab/test-time-hinting}.

\end{document}